\documentclass[10pt,twocolumn,letterpaper]{article}

\usepackage{iccv}

\usepackage{times}
\usepackage{epsfig}
\usepackage{graphicx}
\usepackage{amsmath}
\usepackage{amssymb}
\usepackage{caption}
\usepackage{subcaption}
\usepackage[dvipsnames,svgnames,x11names]{xcolor}
\usepackage{ifthen}
\usepackage[export]{adjustbox}
\usepackage{booktabs}%
\usepackage{array}
\usepackage{bm}
\usepackage{paralist}

\usepackage{enumitem}
\setlist[itemize]{align=parleft,left=0pt..1em}

\usepackage[accsupp]{axessibility}
\usepackage{multirow}

\makeatletter
\@namedef{ver@everyshi.sty}{}
\makeatother
\usepackage{tikz}
\usetikzlibrary{positioning}
\usetikzlibrary{shapes}
\usetikzlibrary{calc}
\usetikzlibrary{backgrounds}
\usetikzlibrary{fit}
\usetikzlibrary{arrows}

\usepackage[pagebackref=true,breaklinks=true,colorlinks,bookmarks=false]{hyperref}

\newboolean{addsupplements} 
\setboolean{addsupplements}{true}

\iccvfinalcopy
\def\iccvPaperID{2350} 

\ificcvfinal\fi

\usepackage{ifthen}
\usepackage{array}
\usepackage{colortbl}

\newcommand{\setmode}[1]{\def\mode{#1}}
\setmode{draft} 
\def\authornote#1#2#3{{\textcolor{#2}{\textsl{\small#1:[*#3*]}}}}

\ifthenelse{\equal{\mode}{draft}} { 
    \newcommand{\mbnote}[1]{\authornote{MB}{Red}{#1}} 
    \newcommand{\rbnote}[1]{\authornote{RB}{Blue}{#1}} 
    \newcommand{\vjnote}[1]{\authornote{VJ}{Magenta}{#1}} 
    \newcommand{\jbnote}[1]{\authornote{JB}{cyan}{#1}} 
    \newcommand{\clnote}[1]{\authornote{CL}{Fuchsia}{#1}} 
    \newcommand{\hlnote}[1]{\authornote{HL}{Orange}{#1}} 
    } {}

\ifthenelse{\equal{\mode}{final}} {
    \newcommand{\mbnote}[1]{}
    \newcommand{\rbnote}[1]{}
    \newcommand{\vjnote}[1]{}
    \newcommand{\jbnote}[1]{}
    \newcommand{\clnote}[1]{}
    \newcommand{\hlnote}[1]{}
    \typeout{************* MODE: Final}
    } {}

\newcolumntype{L}{>{$}l<{$}}
\newcolumntype{C}{>{$}c<{$}}
\newcolumntype{R}{>{$}r<{$}}

\newcommand{\fig}[1]{Fig.~\ref{#1}}
\newcommand{\tbl}[1]{Table~\ref{#1}}

\newcommand{\ignore}[1]{}

\newcommand{\vect}[1]{\bm{#1}}
\newcommand{\matt}[1]{\bm{\MakeUppercase{#1}}}

\newcommand{\expnumber}[2]{{#1}\mathrm{e}{#2}}

\makeatletter
\DeclareRobustCommand\onedot{\futurelet\@let@token\@onedot}
\def\@onedot{\ifx\@let@token.\else.\null\fi\xspace}

\def\eg{\emph{e.g}\onedot} 
\def\ie{\emph{i.e}\onedot} 
 
\def\etc{\emph{etc}\onedot} 
\def\vs{\emph{vs}\onedot}
\def\wrt{w.r.t\onedot} 
\def\etal{\emph{et al}\onedot}

\makeatother

\newcommand\newsubcap[1]{\phantomcaption%
       \caption*{\figurename~\thefigure\thesubfigure: #1}}
       
\newcommand{\inlinesection}[1]{\vspace{0.05cm} \noindent {\bf #1}}
\newcommand{\titlecaption}[2]{\caption{\textbf{#1.}\xspace#2}}
\newcommand{\titlesubcaption}[2]{\newsubcap{\textbf{#1.}\xspace#2}}

\makeatletter
\newcommand{\Distance}[3]{
\tikz@scan@one@point\pgfutil@firstofone($#1-#2$)\relax  
\pgfmathsetmacro{#3}{round(0.99626*veclen(\the\pgf@x,\the\pgf@y)/0.0283465)/1000}
}
\makeatother

\definecolor{mpurple}{RGB}{106,27,154}
\definecolor{mpurplelight}{RGB}{206,147,216}

\definecolor{mblue}{RGB}{40,53,147}
\definecolor{mbluelight}{RGB}{159,168,218}

\definecolor{mteal}{RGB}{0,105,92}
\definecolor{mteallight}{RGB}{128,203,196}

\definecolor{morange}{RGB}{216,67,21}
\definecolor{morangelight}{RGB}{255,171,145}

\definecolor{mgrayblue}{RGB}{120,144,156}

\definecolor{mamber}{RGB}{255,143,0}

\definecolor{mgreen}{RGB}{85,139,47}
\definecolor{mgreenlight}{RGB}{174,213,129}

\definecolor{bestcol}{RGB}{217,95,14}

\definecolor{secondbestcol}{RGB}{254,196,79}
\newcommand{\secondbest}{\cellcolor{secondbestcol}}
\definecolor{thirdbestcol}{RGB}{255,247,188}
\newcommand{\thirdbest}{\cellcolor{thirdbestcol}}

\title{NeRD: Neural Reflectance Decomposition from Image Collections}
\author{\vspace{-1.25em}%
Mark Boss${}^{1}$,\quad 
Raphael Braun${}^{1}$,\quad 
Varun Jampani${}^2$,\quad %
Jonathan T. Barron${}^2$,\\ \quad \\ %
Ce Liu${}^2$,\quad %
Hendrik P.A. Lensch${}^1$%
\\ \quad \\ 
${}^1$University of T{\"{u}}bingen,\quad%
${}^2$Google Research
} 

\renewcommand{\thefootnote}{\fnsymbol{footnote}}

\begin{document}
\captionsetup{style=base}


\twocolumn[{
\maketitle
\vspace{-1.75em}
\captionsetup{type=figure}
\centerline{
\begin{tikzpicture}[node distance=0.3cm]%

\node[black, inner sep=0pt] (input1) {\includegraphics[width=0.075\textwidth,frame]{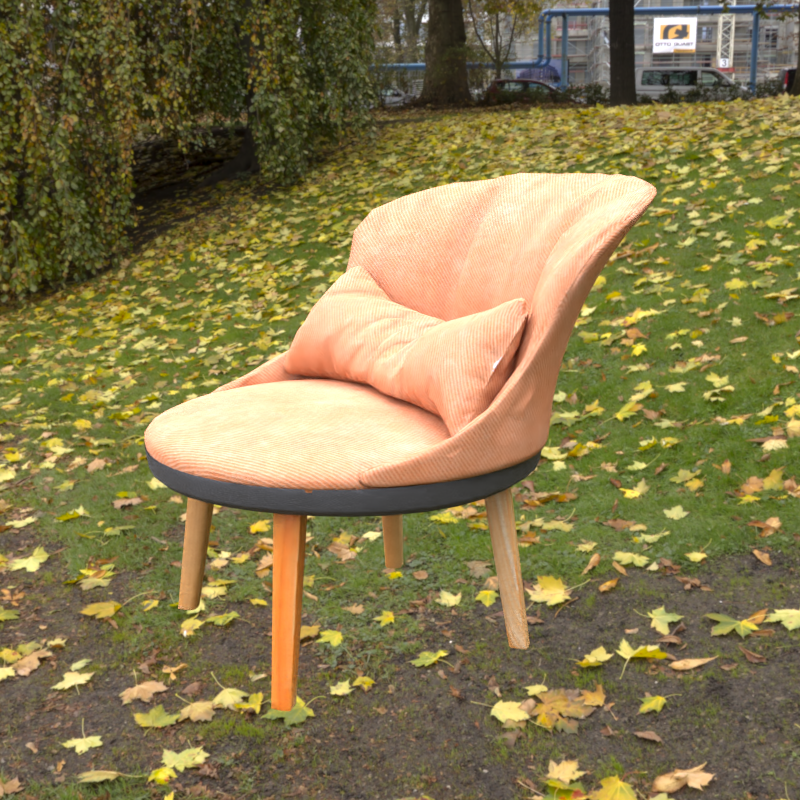}};
\node[black, inner sep=0pt, below=0cm of input1] (input2) {\includegraphics[width=0.075\textwidth,frame]{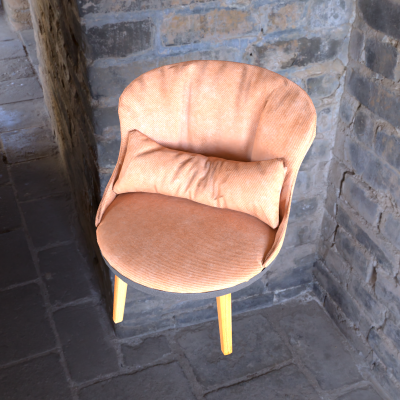}};
\node[black, inner sep=0pt, below=0cm of input2] (input3)  {\includegraphics[width=0.075\textwidth,frame]{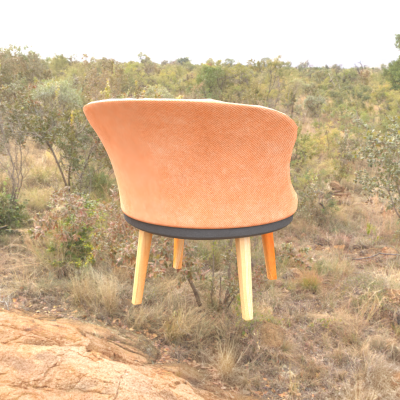}};

\node[left=of input1] (dummy) {};

\node[right=0.1cm of input2] (volumeImg) {\includegraphics[width=0.325\textwidth, trim=300 0 50 0,clip]{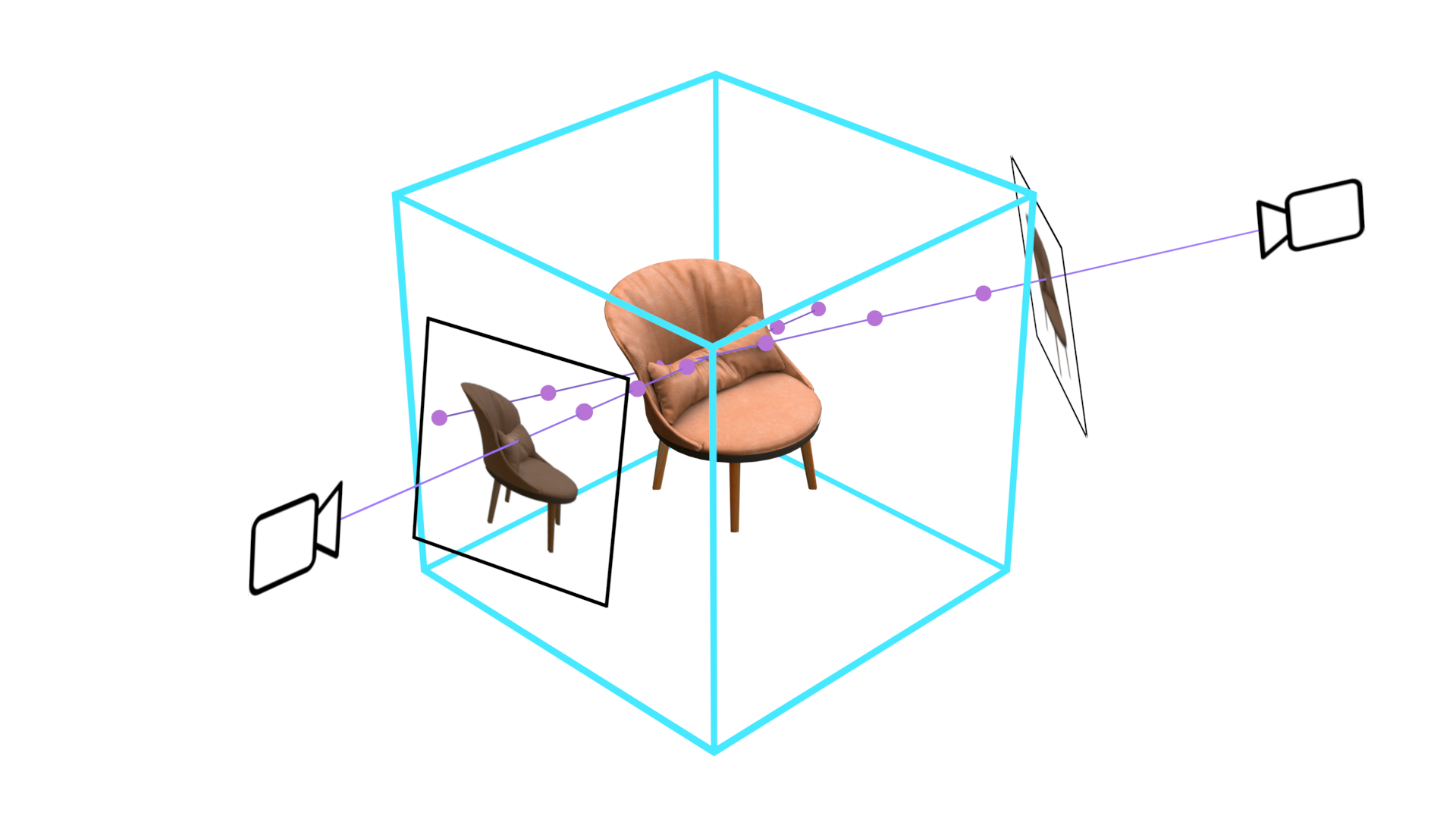}};

\coordinate[right=0.8cm of volumeImg] (brdfLeftMid);

\node[above=0.0125cm of brdfLeftMid] (basecolor) {\includegraphics[width=0.1\textwidth]{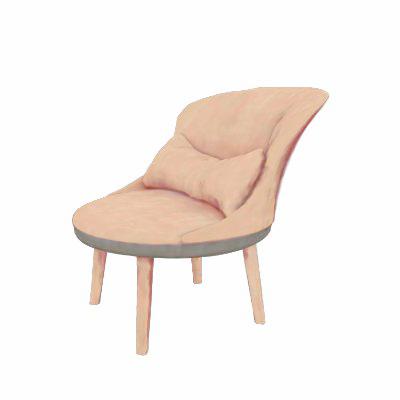}};
\node[below=0.0125cm of brdfLeftMid] (roughness) {\includegraphics[width=0.1\textwidth]{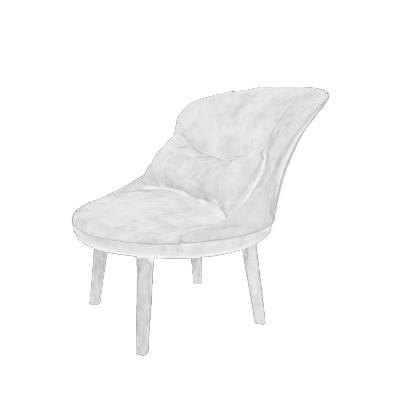}};
\node[right=0.025cm of basecolor] (metallic) {\includegraphics[width=0.1\textwidth]{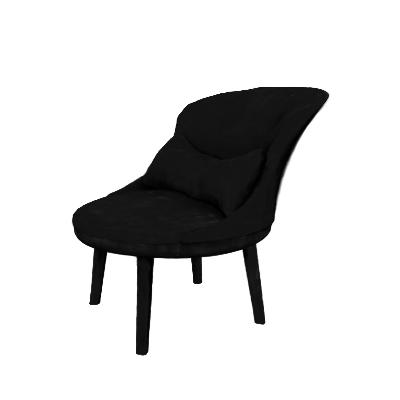}};
\node[right=0.025cm of roughness] (normal) {\includegraphics[width=0.1\textwidth]{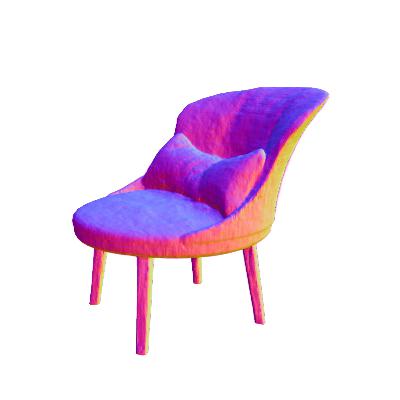}};
\node[draw=black!60!white, fit=(basecolor) (roughness) (metallic) (roughness), inner sep=0pt] (parameterBox) {};
\coordinate (brdfMid) at ($(roughness)!0.5!(normal)$);
\coordinate (brdfRightMid) at (brdfLeftMid -| normal.east);

\node[above=-4mm of basecolor] {\scriptsize  Basecolor};
\node[above=-4mm of metallic] {\scriptsize Metallic};
\node[below=-4mm of roughness] {\scriptsize Roughness};
\node[below=-4mm of normal] {\scriptsize Normal};

\coordinate[right=1.65cm of brdfRightMid] (rerenderMid);
\node[above=0.0125cm of rerenderMid, inner sep=0pt] (rerender1) {\includegraphics[width=0.115\textwidth]{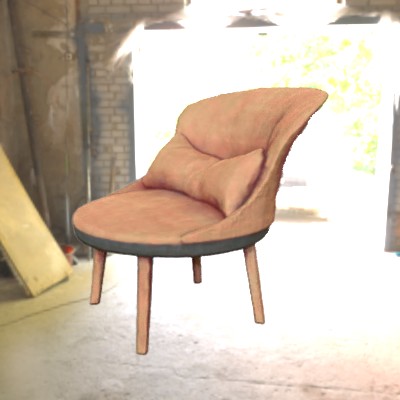}};
\node[below=0.0125cm of rerenderMid, inner sep=0pt] (rerender2) {\includegraphics[width=0.115\textwidth]{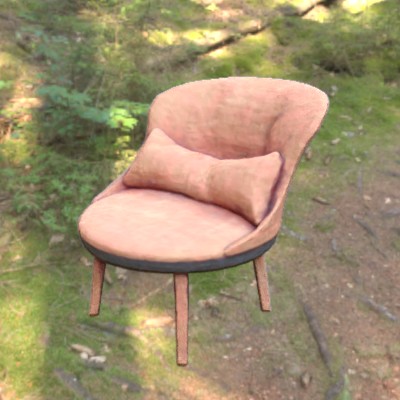}};

\coordinate[right=2.2cm of rerenderMid] (meshMid);
\node[above=0.0125cm of meshMid, inner sep=0pt] (mesh1) {\includegraphics[width=0.115\textwidth]{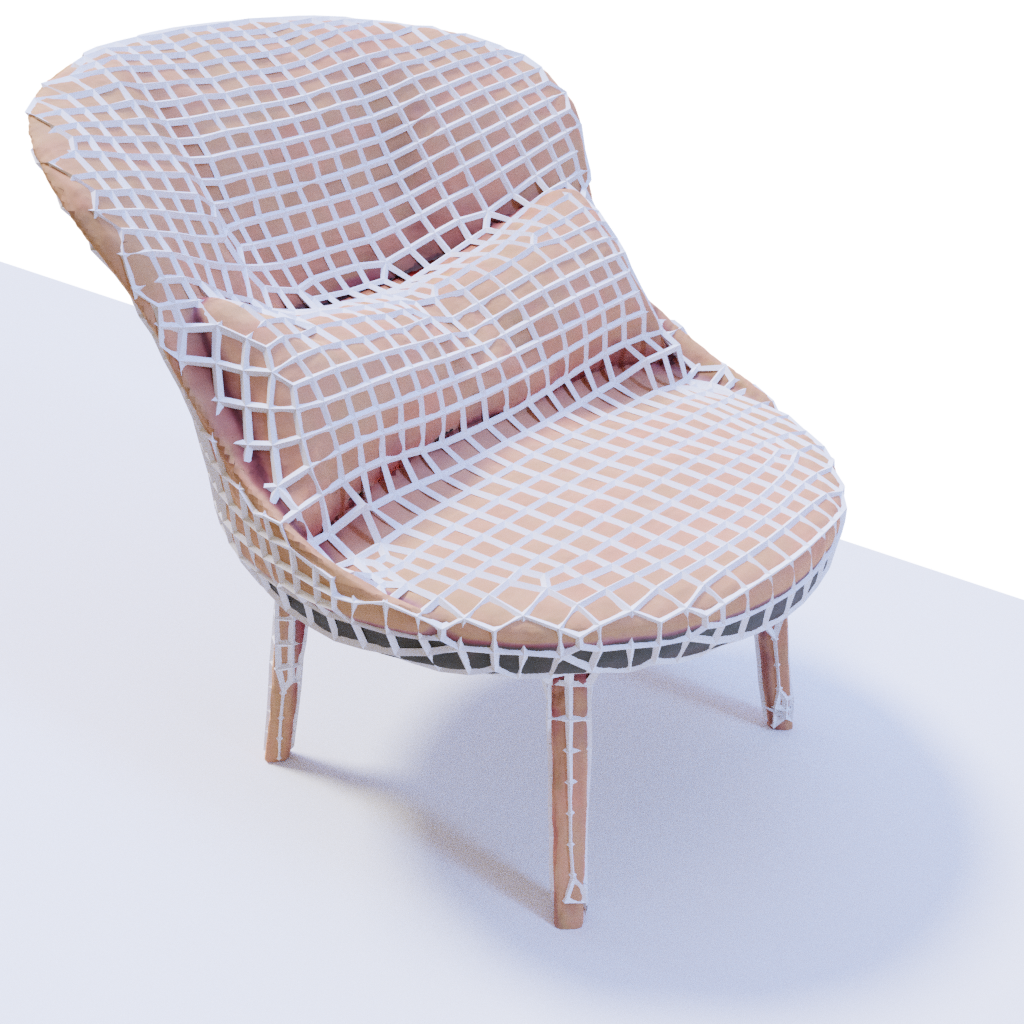}};
\node[below=0.0125cm of meshMid, inner sep=0pt] (mesh2) {\includegraphics[width=0.115\textwidth]{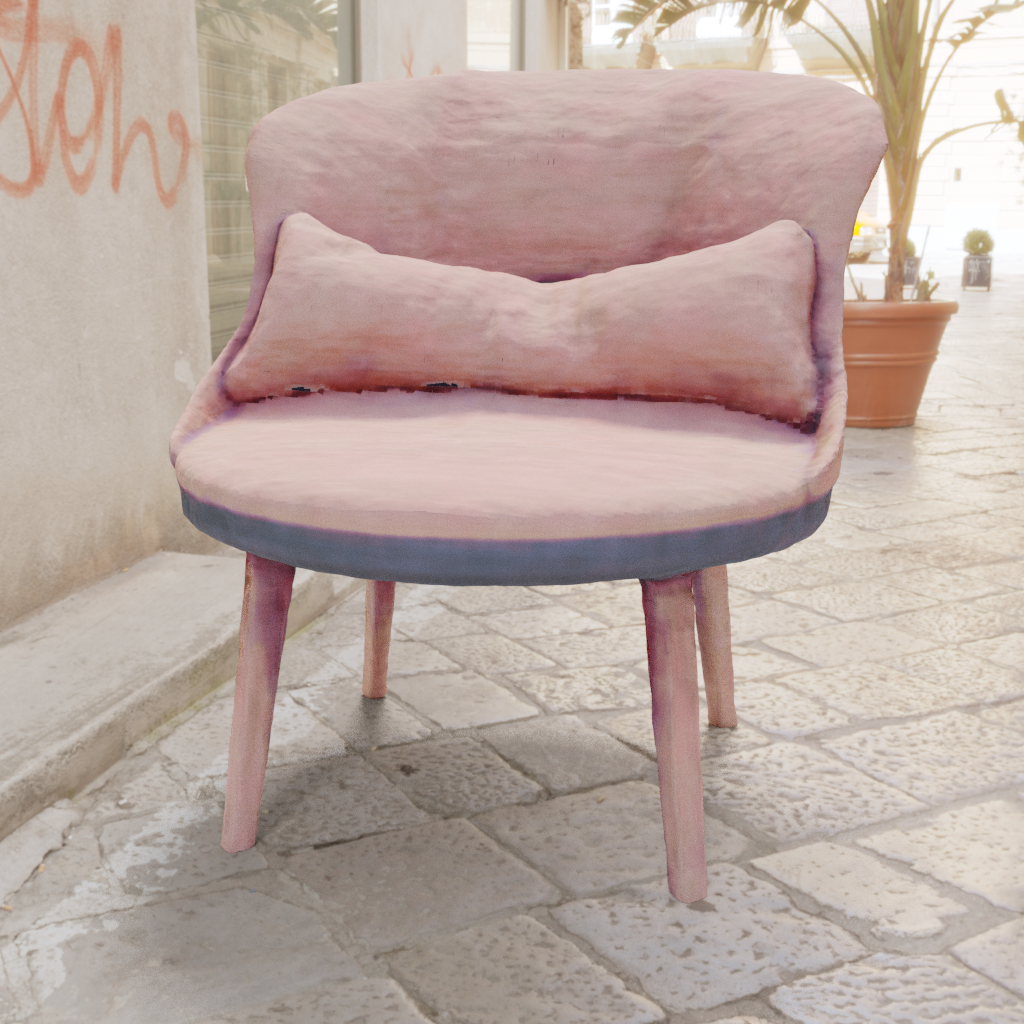}};

\node[draw=black!60!white, fit=(rerender1) (rerender2) (mesh1) (mesh2), inner sep=1pt] (resultBox) {};

\node[below=0.15cm of input3.south -| input2.south,font=\scriptsize] (labelInput) {Multi-View Images};
\node[font=\scriptsize] (labelVolume) at (labelInput -| volumeImg) { NeRD Volume};
\node[font=\scriptsize] (labelBrdf) at (labelInput -| brdfMid) { Decomposed BRDF};
\node[text width=0.115\textwidth, align=center, font=\scriptsize] (labelRerender) at (labelInput -| rerender2)  {Relighting \& View synthesis};
\node[font=\scriptsize] (labelRerender) at (labelInput -| mesh1) {Textured  Mesh};

\draw[-latex, line width=0.75pt] (input2.east) -- ([xshift=0.7cm]volumeImg.west);
\draw[-latex, line width=0.75pt] ([xshift=-1.5cm]volumeImg.east) -- ([xshift=-0.05525\textwidth]brdfLeftMid);
\draw[-latex, line width=0.75pt] (brdfRightMid) -- (resultBox);

\end{tikzpicture}

}
\vspace{-1em}
\caption
{
\textbf{Neural Reflectance Decomposition for Relighting.} \xspace {We encode multiple views of an object under varying or fixed illumination into the NeRD volume. 
We decompose each given image into geometry, spatially-varying BRDF parameters and a rough approximation of the incident illumination in a globally consistent manner. 
We then extract a relightable textured mesh that can be re-rendered under novel illumination conditions in real-time.
}
}
\label{fig:teaser}
\vspace{1em}
}]


\renewcommand*{\thefootnote}{\arabic{footnote}}

\begin{abstract}
\vspace{-4mm}

Decomposing a scene into its shape, reflectance, and illumination is a challenging but important problem in computer vision and graphics. 
This problem is inherently more challenging when the illumination is not a single light source under laboratory conditions but is instead an unconstrained environmental illumination. 
Though recent work has shown that implicit representations can be used to model the radiance field of an object, most of these techniques only enable view synthesis and not relighting.
Additionally, evaluating these radiance fields is resource and time-intensive.  
We propose a neural reflectance decomposition (NeRD) technique that uses physically-based rendering to decompose the scene into spatially varying BRDF material
properties. In contrast to existing techniques, our input images can be captured under
different illumination conditions. In addition, we also propose techniques to convert
the learned reflectance volume into a relightable textured mesh enabling fast real-time
rendering with novel illuminations.
We demonstrate the potential of the proposed approach with experiments on both synthetic and real datasets, where we are able to obtain high-quality relightable 3D assets from image
collections.
\ificcvfinal The datasets and code are available at the project page: \url{https://markboss.me/publication/2021-nerd/}.
\fi
\end{abstract}
\vspace{-4mm}

\section{Introduction}
\vspace{-3mm}

Capturing the geometry and material properties of an object is essential for several computer vision and graphics applications such as view synthesis~\cite{Boss2020, xu2019},
relighting~\cite{Barron2015, Boss2020, Goldman2009, haber2009, li2020inverse, xu2018}, object insertion~\cite{bi2020a, Gardner2017, li2020inverse} \etc.
This problem is often referred to as \emph{inverse rendering}~\cite{Kimiccv17, Ramamoorthi2001},
where shape and material properties are estimated from a set of images, \eg, representing the surface properties as spatially-varying Bidirectional Reflectance Distribution functions (SVBRDF)~\cite{Nicodemus1965}.

Modeled according to physics, the reflected color observed by a viewer is the integral of the product of SVBRDF and the incoming illumination over the hemisphere around that surface's normal~\cite{Kajiya1986}.
Disentangling this integral and estimating shape, illumination, and SVBRDF from images is a highly ill-posed and underconstrained inverse problem.
For instance, an image region may appear dark either due to a dark surface color (material), the absence of incident light at that surface (illumination), or due to the normal of that surface facing away from the incident light (shape).

Traditional SVBRDF estimation techniques involve capturing images using a light-stage setup where the light direction and camera view settings are controlled~\cite{Asselin2020, Boss2018, lawrence2004, Lensch2001, Lensch2003}.
More recent approaches for SVBRDF estimation employ more practical capture setups~\cite{bi2020b, bi2020a, bi2020c, Boss2020, Gao2019, Nam2018}, but limit the illumination to a single dominant source (\eg, a flash attached to a camera).
Assuming known illumination or constraining its complexity significantly reduces the ambiguity of shape and material estimation and limits the practical utility to laboratory settings or to flash photography in dark environments.

In contrast to standard SVBRDF and shape estimation techniques, 
recently introduced coordinate-based scene representation networks such as Neural Radiance Fields (NeRF)~\cite{martinbrualla2020nerfw, mildenhall2020, zhang2020} can directly perform high-quality view synthesis without explicitly estimating shape or SVBRDF.
They represent the radiance field of the scene using a neural network trained specifically for a single scene, using as input multiple images of that scene.
These neural networks directly encode the geometry and appearance as volumetric density and color functions parameterized by 3D coordinates of query points in the scene.
Realistic novel views can be generated by raymarching through the volume.
Though these approaches are capable of reproducing view-dependent appearance effects, the radiance of a point in a direction is ``baked in'' to these networks, making them unusable for relighting.
Even if such techniques could be extended to relighting, the rendering speed of these methods limits their practicality --- the time required by NeRF to generate a single view is about 30 seconds~\cite{mildenhall2020}.

This work presents a shape and SVBRDF estimation technique that allows for a more flexible capture setting  while enabling relighting under novel illuminations.
Our key technique is an explicit decomposition model for shape, reflectance, and illumination 
within a NeRF-like coordinate-based neural representation framework~\cite{mildenhall2020}.
Compared to NeRF, our volumetric geometry representation stores SVBRDF parameters at each 3D point instead of radiance. Each image is then differentiably rendered with a jointly optimized spherical Gaussian illumination model (see Figure~\ref{fig:teaser}). 
Shape, BRDF parameters, and illumination are all optimized simultaneously to minimize the photometric rendering loss \wrt each input image. 
We call our approach ``Neural Reflectance Decomposition'' (NeRD)

NeRD not only enables simultaneous relighting and view synthesis but also allows for a more flexible range of image acquisition settings:
Input images of the object need not be captured under the same illumination conditions.
NeRD supports both camera motion around an object as well as captures of rotating objects.
All NeRD requires as input is a set of images of an object with known camera pose (computed for \eg using COLMAP~\cite{schoenberger2016sfm,schoenberger2016mvs}), where each image is accompanied by a foreground segmentation mask. 
Besides the SVBRDF and shape parameters, we also explicitly optimize the illumination corresponding to each image for varying illuminations or globally for static illumination.

As a post-processing step, we propose a way to extract a 3D surface mesh along with SVBRDF parameters as textures from the learned coordinate-based representation network.
This allows for a highly flexible representation for downstream tasks such as real-time rendering of novel views, relighting, 3D asset generation, \etc.

\section{Related Work}
\vspace{-2mm}


\inlinesection{Neural scene representations.}
Recently, neural scene representations have attracted considerable attention~\cite{liu2020, martinbrualla2020nerfw, mescheder2019occupancy, mildenhall2020, sitzmann2020siren,Sitzmann2019DeepVoxelsLP, qiangeng2019disn, zhang2020, yariv2020multiview}.
These methods surpassed previous state-of-the-art in novel view interpolation and achieved photo-realistic results in most cases. 
The primary innovation of these methods is to model a scene using a volumetric, voxel or implicit representation, and then train a neural network per object to represent it. 
Because these neural representations are inherently 3D, they enable novel view synthesis.
Our approach follows a similar representation but decomposes the appearance into shape, BRDF and illumination.
One significant concern with these approaches is their long training and inference time~\cite{mildenhall2020}.
We address the latter issue by explicitly extracting a surface mesh and BRDF parameters to make use of the learned 3D model in common game engines or path tracers.
Some concurrent works~\cite{srinivasan2020,physg2020} also try to estimate BRDFs in
neural volumes. In NeRV~\cite{srinivasan2020}, the illumination is assumed to be known.
Another work, PhySG~\cite{physg2020} also leverages spherical Gaussians to model the illumination, but constraints itself to scenes under a fixed illumination, compared to our setup which handles both fixed and varying illumination.

\inlinesection{BRDF estimation.}
Though highly accurate BRDF measurements can be achieved under laboratory conditions with known view and light positions~\cite{Asselin2020, Boss2018, lawrence2004, Lensch2001, Lensch2003}, the complicated setup of these methods often renders on-site material capture infeasible. 
Methods aiming at ``casual'' capture frequently rely on neural networks to learn a prior on the relationship between images and their underlying BRDFs.
Often, planar surfaces under camera flash illumination are considered for single-shot~\cite{Aittala2018, Deschaintre2018, Li2018, sang2020single}, few-shot~\cite{Aittala2018} or multi-shot~\cite{Albert2018, Boss2018, Deschaintre2019, Deschaintre2020,Gao2019} estimation.
This casual setup can be extended to estimating the BRDF and shape of objects~\cite{bi2020a,bi2020c,Boss2020,Nam2018,Zhang2020InverseRendering} or scenes~\cite{Sengupta2019}.
Recently, Bi \etal~\cite{bi2020b} leveraged a NeRF-style framework to decompose a scene into shape and BRDF parameters with a single co-located light source. %
Uncontrolled natural illumination adds additional ambiguities, which are partially addressed by self augmented networks~\cite{Li2017, Ye2018} that work on single input images. However, their SVBRDF model assumes homogeneous specularities. Full SVBRDF estimations in natural light setups have been proposed by Dong \etal~\cite{dong2014} by explicitly optimizing for illumination and reflectance from temporal appearance traces of rotating objects. Later, geometry reconstruction was added to the process~\cite{Xia2016}.  %
Our method supports more flexible and practical capture settings.

\begin{figure*}[!htb]
    \centering
    \begin{subfigure}[t]{0.37\textwidth}
        \centering
        \input{figures/architecture/coarse}
        \titlecaption{Sampling Network}{%
            The main task of the coarse sampling network is to generate a finer distribution for sampling in the decomposition network. %
            To match the input during training the color prediction needs to account for the illumination. 
            We combine a compacted $\matt{\Gamma}^j$ from $\text{N}_{\theta_2}$ with the latent color output of $\text{N}_{\theta_1}$ to generate the illumination-dependent color in $\text{N}_{\theta_3}$.
            }
        \label{fig:coarse_architecture}
    \end{subfigure}
    \hfill
    \begin{subfigure}[t]{0.61\textwidth}
        \centering
        \input{figures/architecture/fine}
        \titlecaption{Decomposition Network}{%
            With the sampling pattern generated from the coarse network, we perform SVBRDF decomposition at each point in neural volume. The density, $\sigma$, and direct RGB color $\vect{d}$ is queried from the $\text{N}_{\phi_1}$. Additionally, a vector is passed to $\text{N}_{\phi_2}$, which decodes it to the point's BRDF parameters $\vect{b}$. 
            By compressing the BRDF to a low-dimensional latent space, all surface points contribute to training a joint space of plausible BRDFs for the scene. Each point still interpolates its parameters in this space. 
            The gradient from the density forms the normal $\vect{n}$ and is passed with the BRDF and spherical Gaussians $\Gamma^j$ to the differentiable renderer.
        }
        \label{fig:fine_architecture}
    \end{subfigure}
    \titlecaption{NeRD Architecture}{The architecture consists of two networks. Here, $\text{N}_{\theta_1}$/$\text{N}_{\phi_1}$ denote instances of the main network which encodes the reflectance volume. $r(t)$ defines a ray with sampling positions $\vect{x_i}$, $\gamma(\vect{x})$ is the Fourier Embedding~\cite{mildenhall2020}, and $\vect{\Gamma}^j$ denotes the SG parameters per image $j$. $\vect{c}$ is the output color and $\sigma$ is the density in the volume. The individual samples along the ray need to be alpha composed based on the density $\sigma$ along the ray. This is denoted as ``Comp.''.
    }
    \label{fig:architecture}
\end{figure*}

\section{Method}
\vspace{-2mm}

Our method jointly optimizes a model for shape, BRDF, and illumination by minimizing the photometric error to input image collection of an object that are captured under fixed or different illuminations.

\inlinesection{Problem setup.}
Our input consists of a set of $q$ images with $s$ pixels each, $I_j \in \mathbb{R}^{s \times 3}; j \in {1,...,q}$ potentially captured under different illumination conditions. We aim to learn a 3D volume $\mathcal{V}$, where at each point $\vect{x}=(x,y,z) \in \mathbb{R}^{3}$ in 3D space, we estimate BRDF parameters $\vect{b} \in \mathbb{R}^{5}$, surface normal $\vect{n} \in \mathbb{R}^{3}$ and density $\sigma \in \mathbb{R}$. The environment maps are represented by spherical Gaussian mixtures (SG) with parameters $\matt{\Gamma} \in \mathbb{R}^{24 \times 7}$ (24 lobes).

\inlinesection{Preliminaries.}
We follow the general architecture of NeRF~\cite{mildenhall2020}. 
NeRF creates a neural volume for novel view synthesis using two Multi-Layer-Perceptrons (MLP). 
NeRF model encodes view-dependent color and object density information at each point in 3D space using MLPs.
NeRF consists of two MLPs in which a course \textit{sampling network} samples the volume in a fixed grid and learns the rough shape of an object \ie estimating density $\sigma$ at a given input 3D location $(x,y,z)$.
The second finer network uses this course density information to generate a more 
dense sampling pattern along the viewing ray where higher density gradients are located.
Formally, rays can be defined as $r(t) = \vect{o} + t \vect{d}$ with ray origin $\vect{o}$ and the ray direction $\vect{d}$. Each ray is cast through the image plane and samples a different pixel location with corresponding color $\vect{\hat{c}}^j_r$.
Marching along the ray through the volume at each sample coordinate $\vect{x} = (x, y, z)$, the networks $\text{N}_{(.)}$ are queried for the volume parameters $\vect{p}(t)$.
Here, we use $\vect{p}(t)$ as a stand in for the color $\vect{c}(t)$, density $\sigma(t)$ or in our case BRDF parameters $\vect{b}(t)$.
Following Tancik \etal~\cite{tancik2020fourfeat} and Mildenhall \etal~\cite{mildenhall2020}, which showed that coordinate-based approaches struggle with learning details based on high frequency $\vect{x}$ inputs,
we also use their proposed Fourier embedding $\gamma(\vect{x})$ representation of a 3D point.
The sampled volume parameters are combined along the ray via alpha composition (Comp.) using the density at each point $\sigma(t)$:
$P(\vect{r})=\int_{t_n}^{t_f} T(t)\sigma(t)\vect{p}(t)\, dt$ with $T(t) = \exp \left(-\int_{t_n}^{t} \sigma(s)\, ds \right)$~\cite{mildenhall2020}, based on the 
near and far bounds of the ray $t_n$ and $t_f$ respectively.

\inlinesection{NeRD overview.}
In comparison to NeRF, NeRD architecture mainly differs in the second finer network.
NeRD uses \emph{decomposition network} as a finer network which stores the lighting independent reflectance parameters instead of the direct view-dependent color. Also, the sampling network in NeRD
differs from NeRF as we learn illumination dependent colors as NeRD can work with differently illuminated
input images.
An overview of both networks is shown in \fig{fig:architecture}.
The parameters of the networks and the SGs are optimized by backpropagation informed by comparing the output of a differentiable rendering step to each input image $I_j$ for individual rays across the 3D volume.

\begin{figure*}
    \centering
    \begin{subfigure}[t]{0.33\linewidth}
        \centering
        \begin{tikzpicture}[node distance=0.1cm]%
        \node (input) {\includegraphics[width=0.36\linewidth]{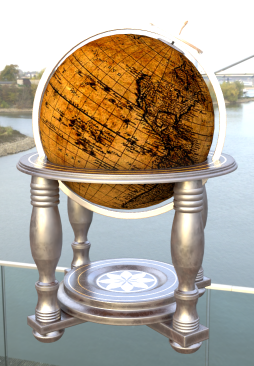}};
        \node[right=0mm of input] (segmentation) {\includegraphics[width=0.36\linewidth]{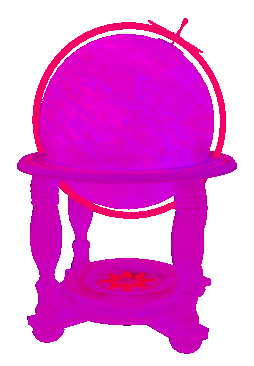}};
        \node[below=0.05mm of input] {\scriptsize \textbf{Input}};
        \node[below=0.05mm of segmentation] {\scriptsize \textbf{Material Map}};
        \end{tikzpicture}
        \titlesubcaption{Compressed BRDF Space}{Instead of directly estimating the BRDF, we learn a 2D embedding per scene which clusters similar materials. As several points jointly estimate BRDFs, this stabilizes the decomposition and improves quality. Notice how similar materials are identified across the surface in the resulting material map.
        }
        \label{fig:brdf_segmentation}
    \end{subfigure}
    \hfill
    \begin{subfigure}[t]{0.38\linewidth}
        \centering
        \begin{tikzpicture}[node distance=0.1cm]%
            \node (no_sgs) {\includegraphics[width=0.44\linewidth]{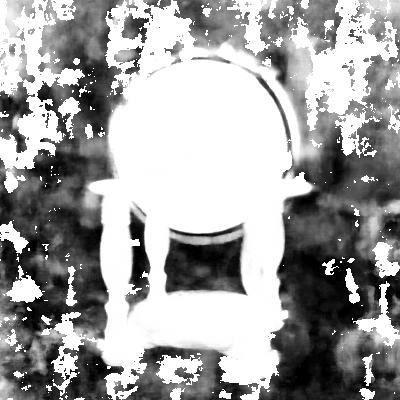}};
            \node[right=0mm of no_sgs] (sgs) {\includegraphics[width=0.44\linewidth]{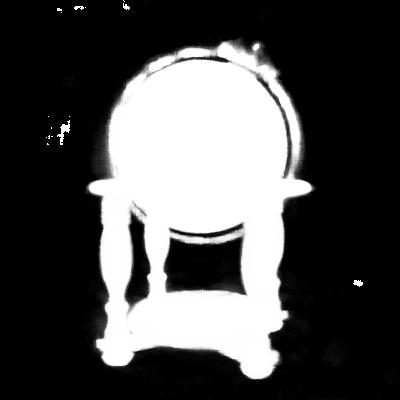}};
            \node[below=0.05mm of no_sgs] {\scriptsize \textbf{Direct color prediction}};
            \node[below=0.05mm of sgs] {\scriptsize \textbf{Considering illumination}};
        \end{tikzpicture}
        \titlesubcaption{SGs dependent Sampling Network}{$\text{N}_{\theta_1}$ can to some extent model view-dependence by composition along the ray. 
        This, however, is too weak to deal with varying illumination. Apparent highlights introduce spurious geometry that mimics the effect for individual views. 
        We can obtain better shape by estimating the illumination dependent radiance with Spherical Gaussians (SG).}
        \label{fig:sgs_addition}
    \end{subfigure}
    \hfill
    \begin{subfigure}[t]{0.25\linewidth}
        \centering
        \begin{tikzpicture}[node distance=0.1cm]%
            \node (learned) {\includegraphics[width=0.44\linewidth,trim=72 27 75 10,clip]{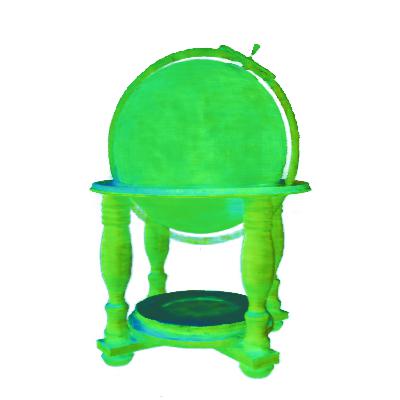}};
            \node[right=0mm of learned] (gradient) {\includegraphics[width=0.44\linewidth,trim=72 27 75 10,clip]{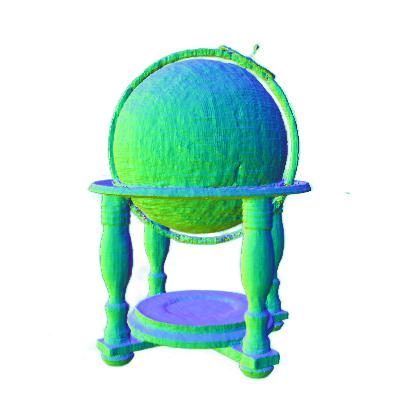}};
            \node[below=0.05mm of learned] {\scriptsize \textbf{Directly Predicted}};
            \node[below=0.05mm of gradient] {\scriptsize \textbf{$\nabla \sigma$-based}};
        \end{tikzpicture}
        \titlesubcaption{Surface Normal Estimation}{Instead of directly predicting the normal as another output of the $\text{N}_{\theta_1}$, the normal in our approach is calculated from the gradient of the density $\nabla \sigma$. Photometric information thus influences both $\vect{n}$ and $\sigma$ during training.}
        \label{fig:gradient_normals}
    \end{subfigure}
    \label{fig:techniques}
\end{figure*}

\inlinesection{Sampling network.}
The \emph{sampling network} directly estimates a view-independent but illumination dependent color $\vect{c}^j$ at each point, which is optimized by a MSE: $\frac{1}{s}\sum^s (\vect{\hat{c}}^j_r - \vect{c}^j_r)^2$.
The sampling network's main goal is to establish a useful sampling pattern for the \emph{decomposition network}. The sample network structure is visualised in \fig{fig:coarse_architecture}.
Compared to NeRF, our training images can have varying illuminations. 
Therefore, the network needs to consider the illumination $\matt{\Gamma}^j$ to create the illumination dependent color $\vect{c}^j$ that should match image $I_j$.
The density $\sigma$ is not dependent on the illumination, which is why we extract it directly as the side-output of $\text{N}_{\theta_1}$.
Here, we follow a concept from NeRF-w~\cite{martinbrualla2020nerfw} that combines an embedding of the estimated illumination with the latent color vector produced by $\text{N}_{\theta_1}$.
As the dimensionality of the SGs can be large, we add a compaction network ($\text{N}_{\theta_2}$), which encodes the $24 \times 7$ dimensional SGs to 16 dimensions.
The compacted SG's embedding is then appended to the output of the last layer of $\text{N}_{\theta_1}$ and jointly passed to the final estimation network $\text{N}_{\theta_3}$ that outputs color values.
Without the illumination dependent color prediction, several floaters would appear in the volume estimate, introducing wrong semi-transparent geometry to paint in highlights for individual views (see \fig{fig:sgs_addition}).
By introducing the illumination-dependent branch, the resulting 3D volume is sparser and more consistent.

\inlinesection{Decomposition network.}
After a ray has sampled the \emph{sampling network}, additional $m$ samples are placed based on the density $\sigma$. This is visualized in \fig{fig:fine_architecture} as the additional green points on the ray.
The decomposition network is trained with the same loss as the \emph{sampling network}. However, we introduce an explicit decomposition step and a rendering step in-between.
Our decomposition step estimates view and illumination independent BRDF parameters $\vect{b}$ and a surface normal $\vect{n}$ at each point. 
The popular Cook-Torrance analytical BRDF model~\cite{Cook1982} is used for rendering.
Here, we choose the Disney BRDF Basecolor-Metallic parametrization~\cite{Burley2012} instead of independently predicting the diffuse and specular color, as it enforces physical correctness.
The illumination $\matt{\Gamma}^j$, in the form of spherical Gaussians (SG), is also jointly optimized.
After rendering the decomposed parameters, the final output is a view and illumination dependent color $\vect{c}^j_{{\omega_o}_r}$.

By keeping the rendering differentiable, the loss from the input color $\vect{\hat{c}}^j_r$ can be backpropagated to the BRDF $\vect{b}$, the normal $\vect{n}$, and the illumination $\matt{\Gamma}^j$.
Our rendering step approximates the general rendering equation $L_o(\vect{x},\vect{\omega}_o) = \int_\Omega f_r(\vect{x},\vect{\omega_i},\vect{\omega_o})L_i(\vect{x},\vect{\omega_i})(\vect{\omega_i} \cdot \vect{n}) d\vect{\omega_i}$ using a sum of 24 SG evaluations. The $\vect{\omega_i}$ and $\vect{\omega_o}$ defines the incoming and outgoing ray direction, respectively.
The reflectance due to diffuse and specular lobes is separately evaluated by functions $\rho_d$ and $\rho_s$, respectively~\cite{Wang2009}.
Overall, our image formation is defined as:
$L_o(\vect{x},\vect{\omega}_o) \approx \sum^{24}_{m=1} \rho_d(\vect{\omega}_o,\matt{\Gamma}_m,\vect{n},\vect{b}) + \rho_s(\vect{\omega}_o,\matt{\Gamma}_m,\vect{n},\vect{b})$. Our differentiable rendering implementation follows Boss \etal~\cite{Boss2020}.

The overall network architecture is shown in \fig{fig:fine_architecture}. 
Especially in the early stages of the estimation, joint optimization of BRDF and shape proved difficult.
Therefore, we estimate the density $\sigma$ and, in the beginning, a view-independent color $\vect{d}$ for each point in $\text{N}_{\phi_1}$. The direct color prediction $\vect{d}$ is compared with the input image, and the loss is faded out over time when the rough shape is established.

To compute the shading, the surface normal is required.
One approach could be to simply learn the normal as another output~\cite{bi2020b}. 
However, this typically leads to inconsistent normals that do not necessarily fit the object's shape (\fig{fig:gradient_normals}). 
Specific reflections can be created by shifting the normal instead of altering the BRDF.
Coupling the surface normal to the actual shape can resolve some of this ambiguity~\cite{Boss2020}.
In coordinate-based volume representations like NeRF~\cite{mildenhall2020}, we can establish this link by defining the normal as the normalized negative gradient of the density field: $\vect{n} = - \frac{\nabla_{\vect{x}} \sigma}{\| \nabla_{\vect{x}} \sigma \|}$. 
While the density field defines the surface implicitly, the density in the 3D volume changes drastically at the boundary between non-opaque air to the opaque object. 
Thus, the gradient at a surface will be perpendicular to the implicitly represented surface. This is a similar to the normal reconstruction from SDFs of Yariv \etal~\cite{yariv2020multiview}.

By calculating the gradient inside the optimization and allowing the photometric loss from the differentiable rendering to optimize the normal, we optimize the $\sigma$ parameter in the second order. 
Therefore, the neighborhood of surrounding points in the volume is smoothed and made more coherent with the photometric observations. 
As a more densely defined implicit volume allows for a smoother normal, we additionally jitter the ray samples during training.
Each ray is now cast in a subpixel direction, and the target color is obtained by bilinear interpolation.

For the BRDF estimation, we use the property that often real-world objects consist of a few highly similar BRDFs which might be spatially separated.
To account for this we introduce an additional network $\text{N}_{\phi_2}$ which receives the latent vector output of $\text{N}_{\phi_1}$.
This autoencoder creates a severe bottleneck, a two-dimensional latent space, which encodes all possible BRDFs in this scene.
As the embedding space enforces a compression, similar BRDFs will share the same embedding.
This step couples the BRDF estimation of multiple surface points, increasing the robustness.
The assignment to various BRDFs is visualized \fig{fig:brdf_segmentation}, which can be utilized for material-based segmentation.

The approach will converge to a globally consistent state, as the underlying shape and BRDF is assumed to be the same for all input images. The SGs are estimated for each input image, but we can force them to be the same or a rotated version of a single SG in case of static illumination.

\inlinesection{Dynamic range, tonemapping and whitebalancing.}
As most online image collections consist of Low Dynamic Range (LDR) images with at least an sRGB curve and white balancing applied, we have to ensure that our rendering setup's linear output recreates these mapping steps before computing a loss.
However, rendering can produce a large value range depending on the incident light and the object's specularity. 
Real-world cameras also face this problem and tackle it by changes in aperture, shutter speed, and ISO. 
Based on the meta-data information encoded in JPEG files, we can reconstruct the input image's exposure value and apply this to our re-rendering.
NeRD is then forced to always work with physically plausible ranges. 
For synthetic examples, we calculate these exposure values based on Saturation Based Sensitivity auto exposure calculation~\cite{ISO12232} and also apply an sRGB curve.

Cameras also apply a white balancing based on the illumination, or it is set by hand afterward. 
This can reduce some ambiguity between illumination and material color and, in particular, fixes the overall intensity of the illumination. 
For synthetic data, we evaluate a small spot of material with 80\% gray value in the environment. 
We assume a perfect white balancing and exposure on real-world data, at least for one of the input images. 
The RGB color ($\vect{w}$) of the white point is stored.
After each training step a single-pixel 
with a rough 80\% gray material is rendered in the estimated illumination and a factor $\vect{f} = \frac{\vect{w}}{\vect{b}}$ is calculated. This factor is then applied to the corresponding SG. 
As the training will adopt the BRDF to the normalized SG, a single white-balanced input can implicitly update and correct all other views. 
In practice, the calculated factor $\vect{f}$ could change the SGs abruptly in one step causing unstable training. 
Therefore, we clip the range of $\vect{f}$ to $[0.99;1.01]$ to spread the update over multiple training iterations. 

\inlinesection{Mesh extraction.}
The ability to extract a consistent textured mesh from NeRD after training is one of the key advantages of the decomposition approach and enables real-time rendering and relighting.
This is not possible with NeRF-based approaches where the view-dependent appearance is directly baked into the volume.
The basic process generates a point cloud, computes a mesh, including a texture atlas, and then fills the texture atlas with BRDF parameters. More details are given in the supplementary material.

\inlinesection{Training and losses.}
The estimation is driven by a Mean Squared Error (MSE) loss between the input image and the results of evaluating randomly generated rays.
For the \emph{sampling network} this loss is applied to the RGB prediction and for the \emph{decomposition network} to the re-rendered result $\vect{c}^j$ and the direct color prediction $\vect{d}$.
The loss for the color prediction based on $\vect{d}$ is exponentially faded out.
Additionally, we leverage the foreground/background mask as a supervision signal, where all values along the ray in background regions are forced to 0. 
This loss is exponentially faded in throughout the training to reduce optimization
instabilities.
By gradually increasing this loss, the network is forced to provide a more accurate silhouette, which prevents the smearing of information at the end of the training. 
The networks are trained for 300K steps with the Adam optimizer~\cite{kingma2014adam} with a learning rate of $\expnumber{5}{-4}$.
On 4 NVIDIA 2080 Ti, the training takes about 1.5 days.
The final mesh extraction takes approximately 90 minutes.

\section{Results}
\vspace{-2mm}

\begin{figure*}[!htb]
    \centering
    \normalsize%
\renewcommand{\arraystretch}{0.6}%
\setlength{\tabcolsep}{0pt}%
\begin{tabular}{@{}l>{\centering\arraybackslash}m{42pt}>{\centering\arraybackslash}m{42pt}>{\centering\arraybackslash}m{42pt}>{\centering\arraybackslash}m{42pt}>{\centering\arraybackslash}m{84pt}>{\centering\arraybackslash}m{42pt}>{\centering\arraybackslash}m{42pt}>{\centering\arraybackslash}m{42pt}>{\centering\arraybackslash}m{42pt}>{\centering\arraybackslash}m{42pt}@{}}%
&\scriptsize Base Color&\scriptsize Metalness&\scriptsize Roughness&\scriptsize Normal&\scriptsize Environment&\scriptsize Image&\scriptsize Relight 1& \scriptsize Relight 2 & \scriptsize Point Light 1 & \scriptsize Point Light 2\\\midrule%
\rotatebox[origin=c]{90}{\footnotesize Ours}&\includegraphics[width=42pt]{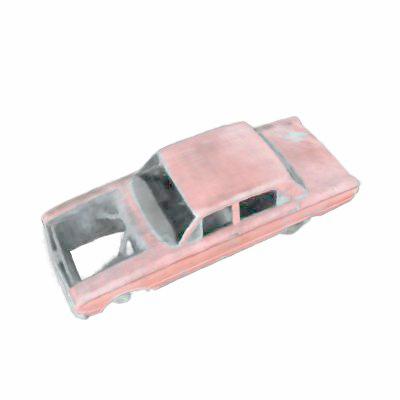}&\includegraphics[width=42pt]{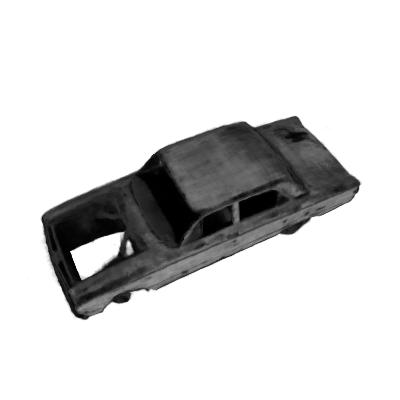}&\includegraphics[width=42pt]{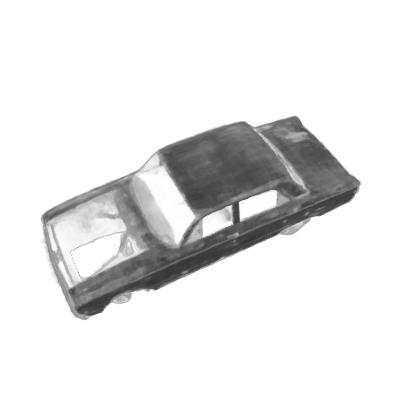}&\includegraphics[width=42pt]{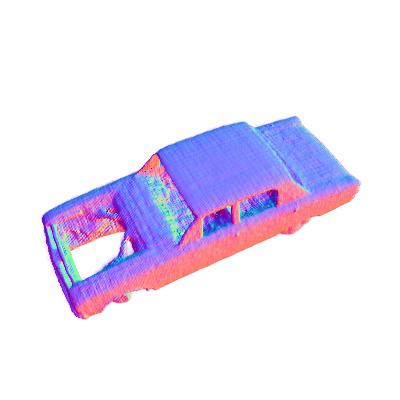}&\includegraphics[width=84pt]{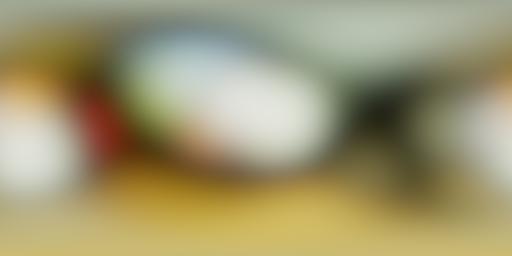}&\includegraphics[width=42pt]{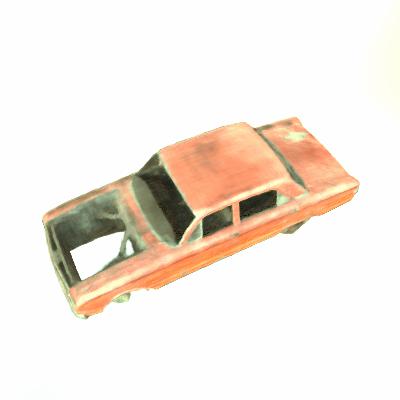}&\includegraphics[width=42pt]{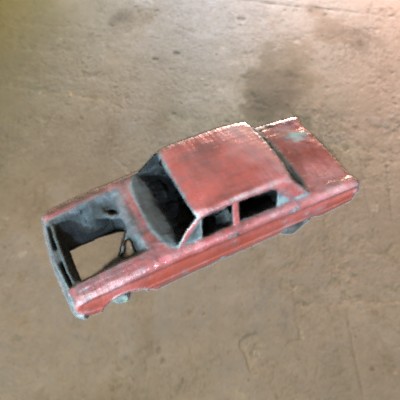}&\includegraphics[width=42pt]{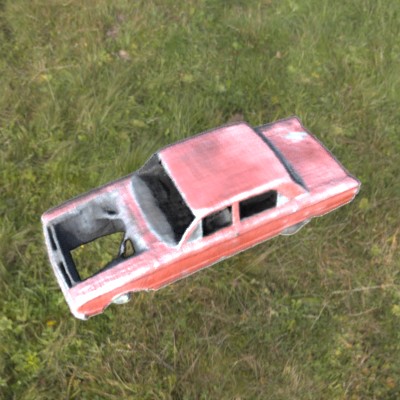}&\includegraphics[width=42pt]{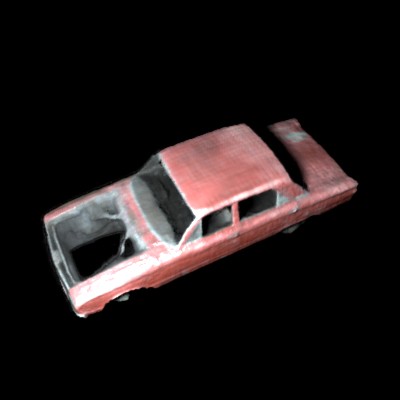}&\includegraphics[width=42pt]{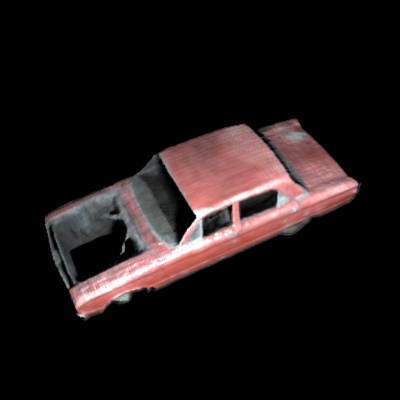}\\%
%
\rotatebox[origin=c]{90}{\footnotesize GT}&\includegraphics[width=42pt]{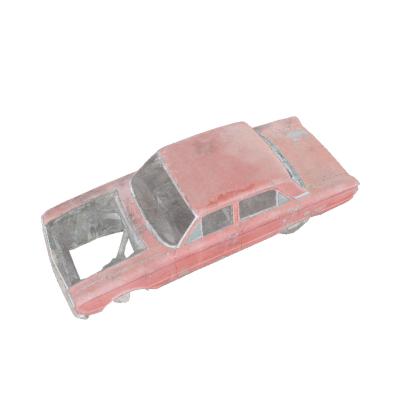}&\includegraphics[width=42pt]{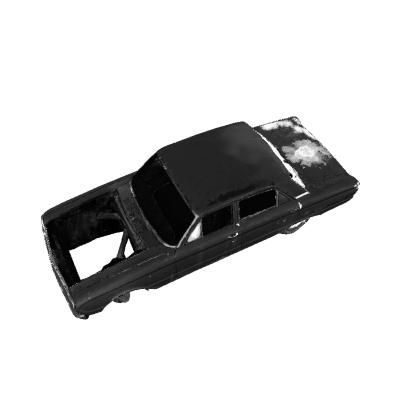}&\includegraphics[width=42pt]{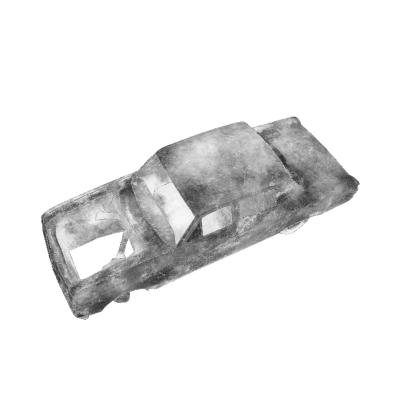}&\includegraphics[width=42pt]{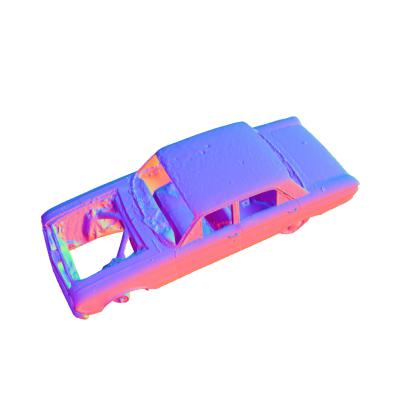}&\includegraphics[width=84pt]{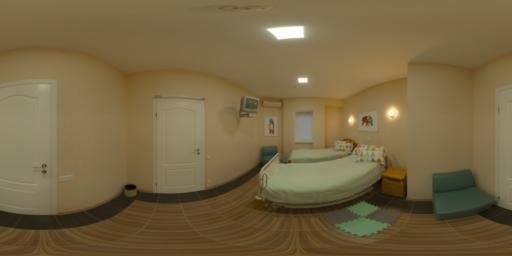}&\includegraphics[width=42pt]{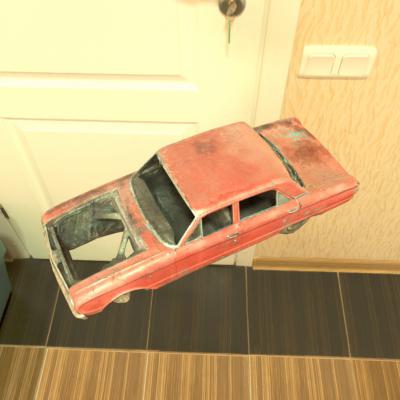}&\includegraphics[width=42pt]{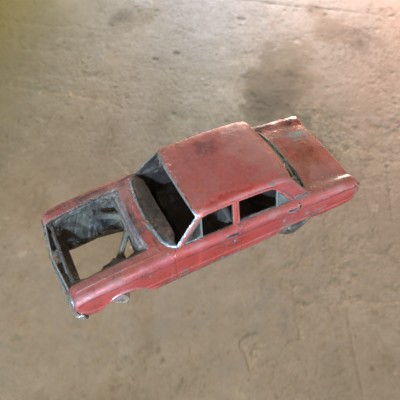}&\includegraphics[width=42pt]{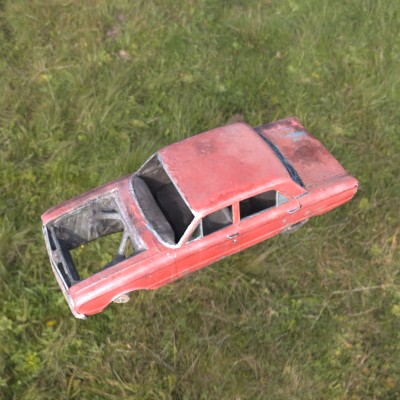}&\includegraphics[width=42pt]{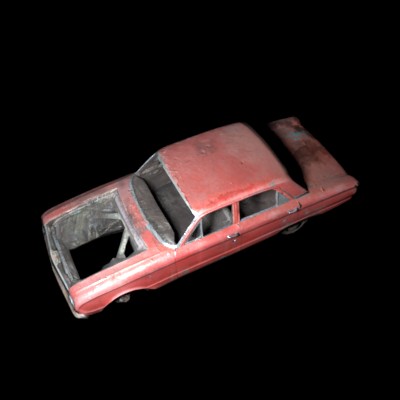}&\includegraphics[width=42pt]{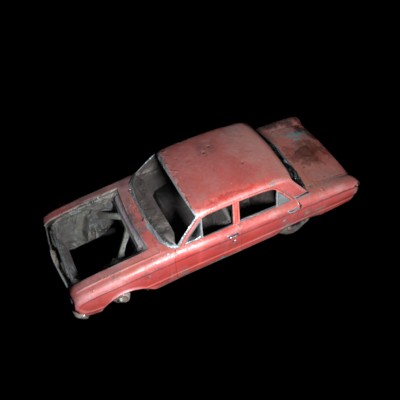}\\\midrule %
\rotatebox[origin=c]{90}{\footnotesize Ours}&\includegraphics[width=42pt]{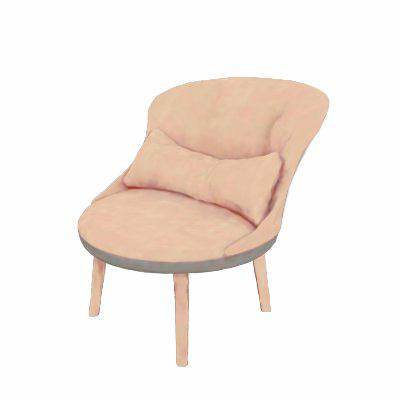}&\includegraphics[width=42pt]{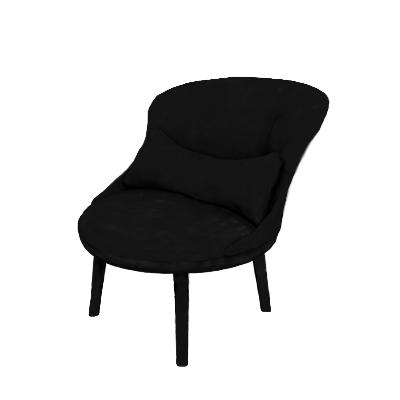}&\includegraphics[width=42pt]{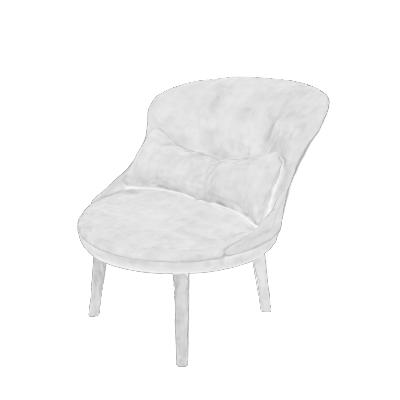}&\includegraphics[width=42pt]{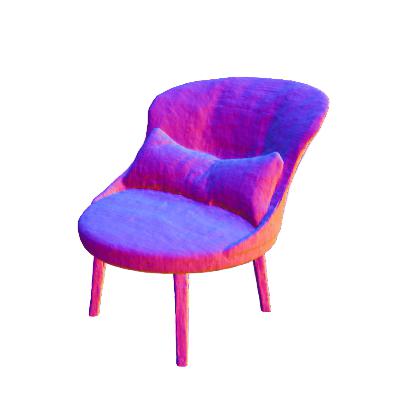}&\includegraphics[width=84pt]{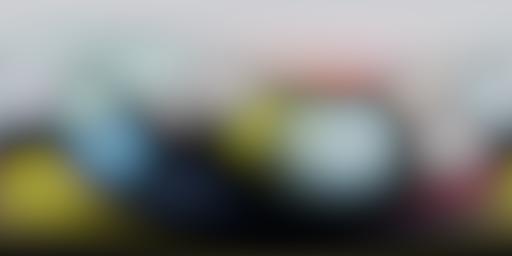}&\includegraphics[width=42pt]{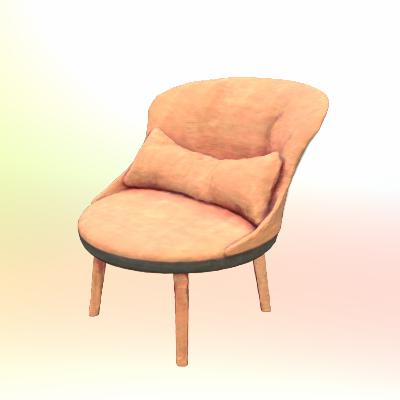}&\includegraphics[width=42pt]{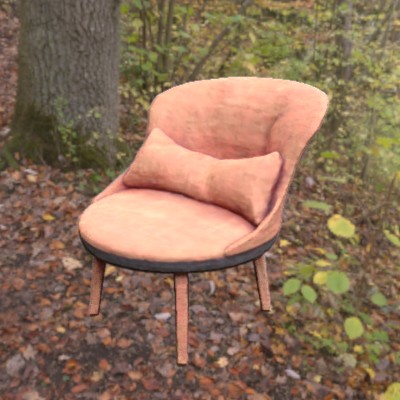}&\includegraphics[width=42pt]{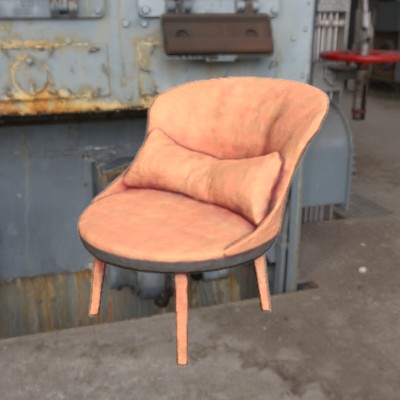}&\includegraphics[width=42pt]{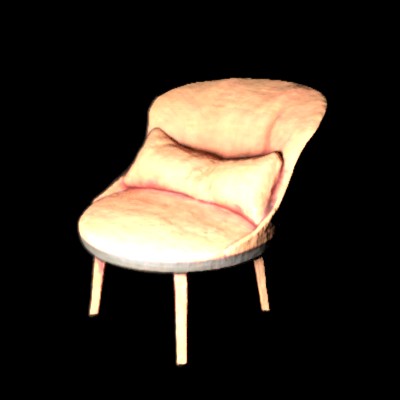}&\includegraphics[width=42pt]{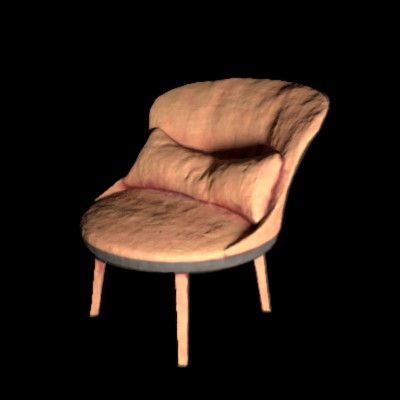}\\%
%
\rotatebox[origin=c]{90}{\footnotesize GT}&\includegraphics[width=42pt]{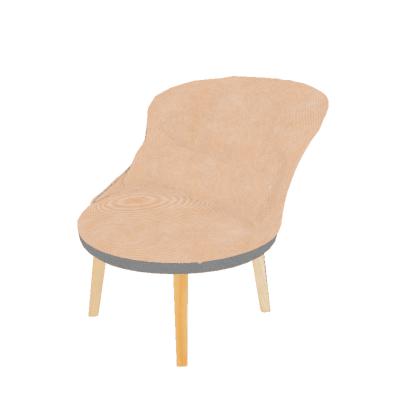}&\includegraphics[width=42pt]{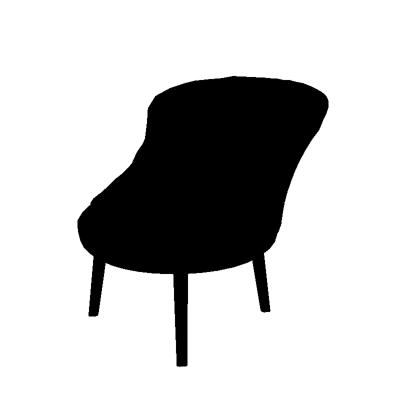}&\includegraphics[width=42pt]{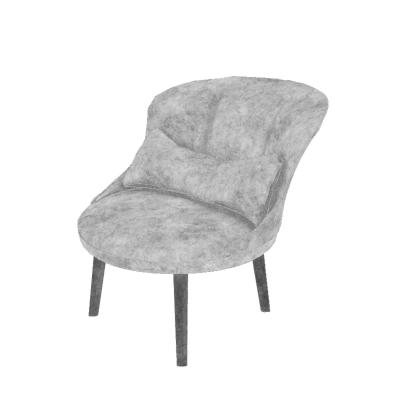}&\includegraphics[width=42pt]{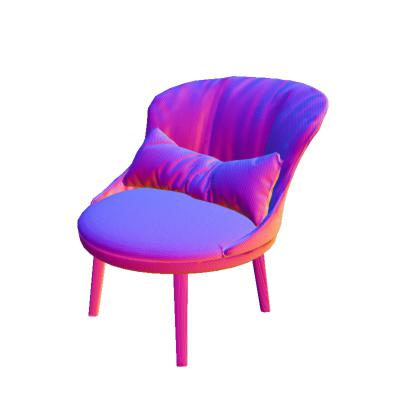}&\includegraphics[width=84pt]{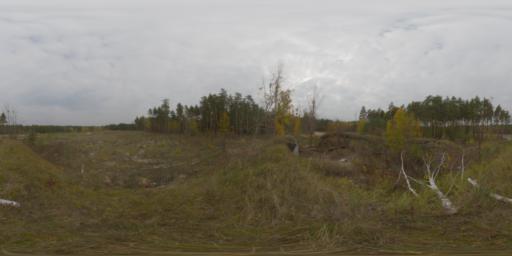}&\includegraphics[width=42pt]{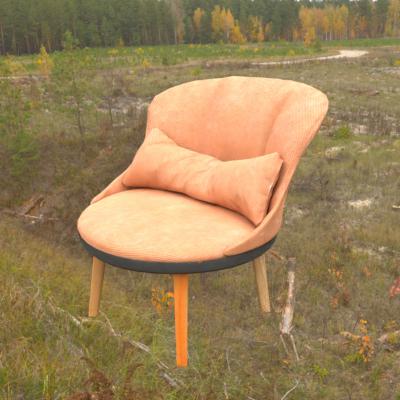}&\includegraphics[width=42pt]{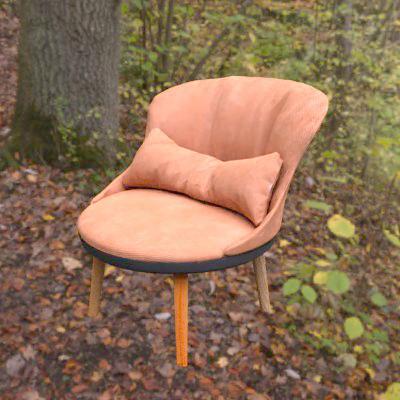}&\includegraphics[width=42pt]{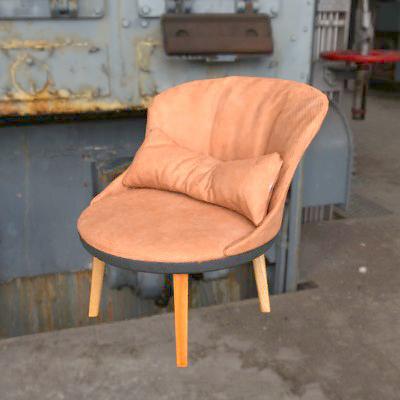}&\includegraphics[width=42pt]{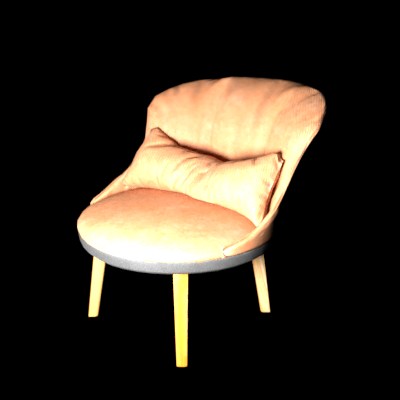}&\includegraphics[width=42pt]{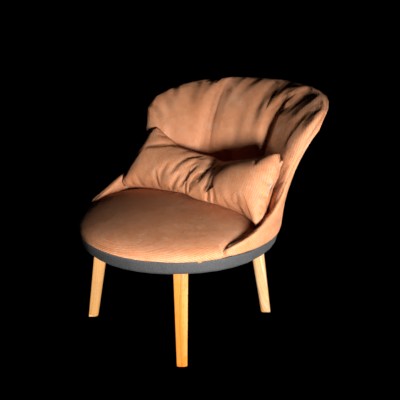}\\
\end{tabular}

    \titlecaption{Decomposition on Synthetic Examples}{Two scenes are highlighted to show the decomposition performance of our method. Notice the accurate performance in relighting with unseen illuminations.}
    \label{fig:syn_brdf}
\end{figure*}

\begin{figure*}[!htb]
    \centering
    \normalsize%
\renewcommand{\arraystretch}{0.6}%
\setlength{\tabcolsep}{0pt}%
\begin{tabular}{@{}>{\centering\arraybackslash}m{40pt}>{\centering\arraybackslash}m{40pt}>{\centering\arraybackslash}m{40pt}>{\centering\arraybackslash}m{40pt}>{\centering\arraybackslash}m{40pt}>{\centering\arraybackslash}m{40pt}>{\centering\arraybackslash}m{40pt}>{\centering\arraybackslash}m{40pt}>{\centering\arraybackslash}m{40pt}>{\centering\arraybackslash}m{40pt}@{}}%
\tiny Base Color&\tiny Metalness&\tiny Roughness&\tiny Normal&\tiny GT&\tiny Re{-}render - Illum 0&\tiny Illum 1&\tiny View + Illum 1&\tiny View + Illum 2&\tiny View + Illum 0\\%
\midrule%
\includegraphics[width=40pt,trim=115 50 115 50, clip]{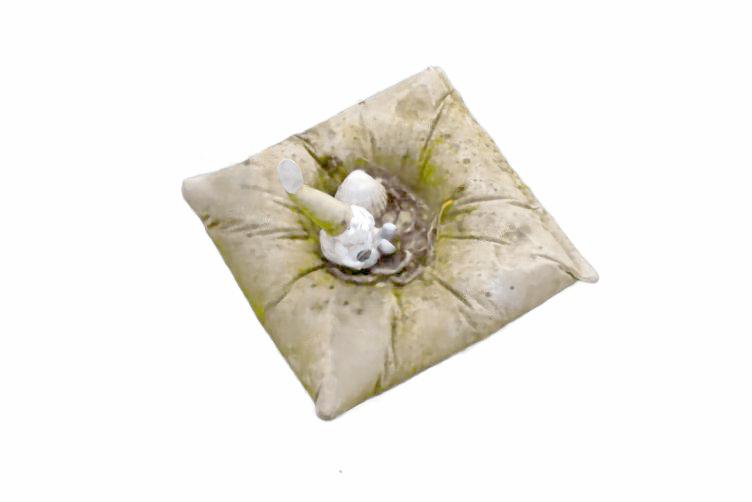}&\includegraphics[width=40pt,trim=145 60 145 60, clip]{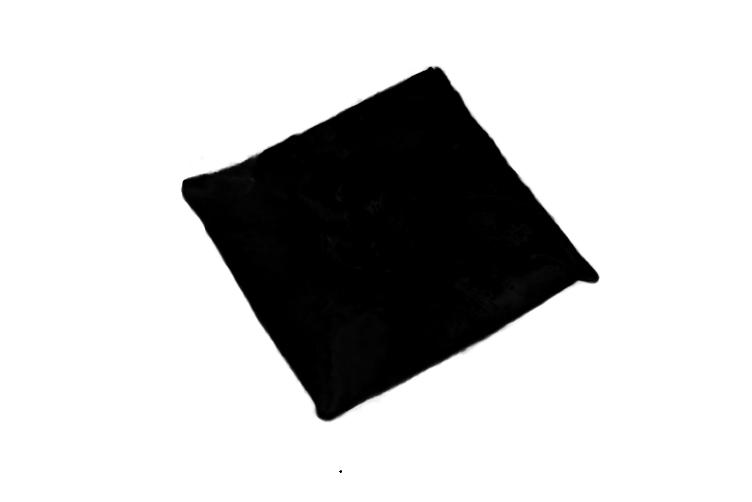}&\includegraphics[width=40pt,trim=145 60 145 60, clip]{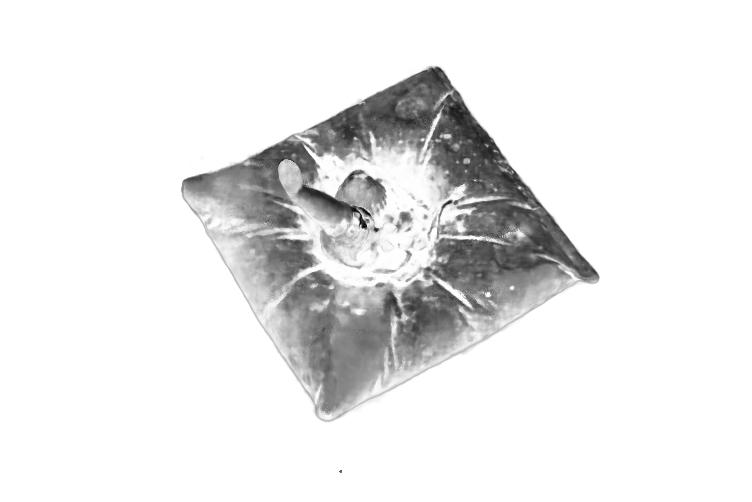}&\includegraphics[width=40pt,trim=145 60 145 60,clip]{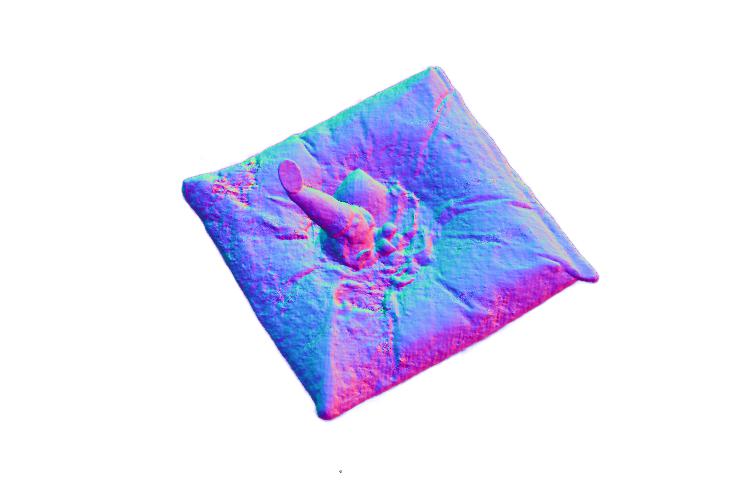}&\includegraphics[width=40pt,trim=5 15 5 15,clip]{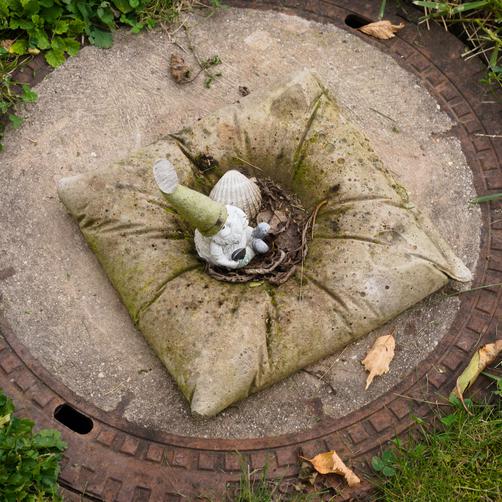}&\includegraphics[width=40pt,trim=145 60 145 60,clip]{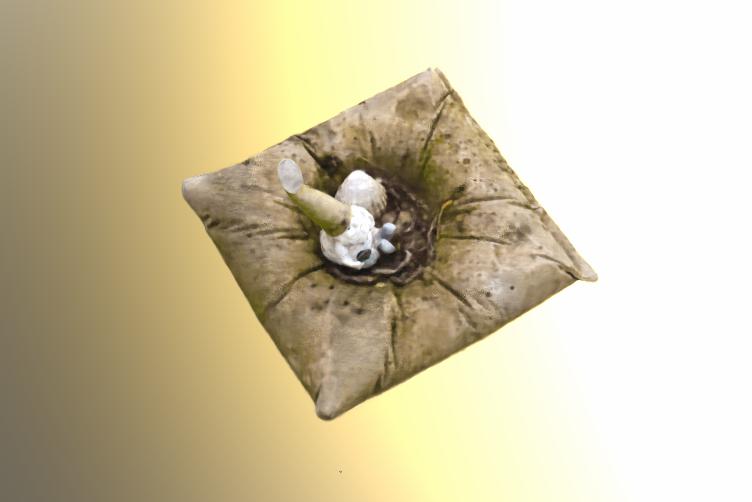}&\includegraphics[width=40pt,trim=145 60 145 60,clip]{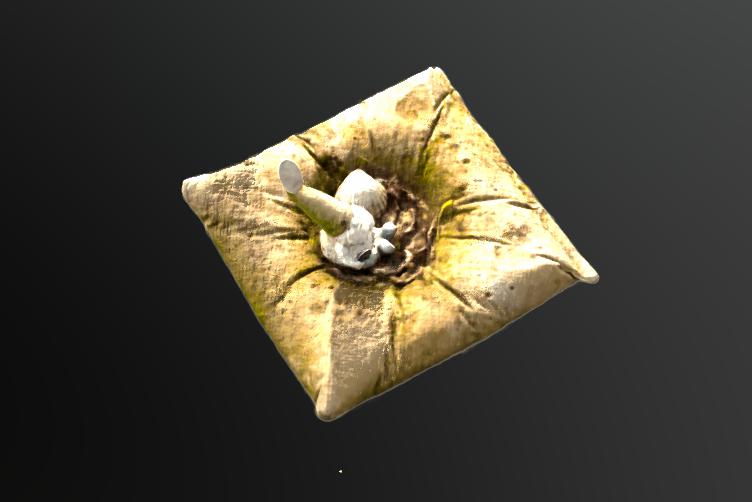}&\includegraphics[width=40pt,trim=145 60 145 60,clip]{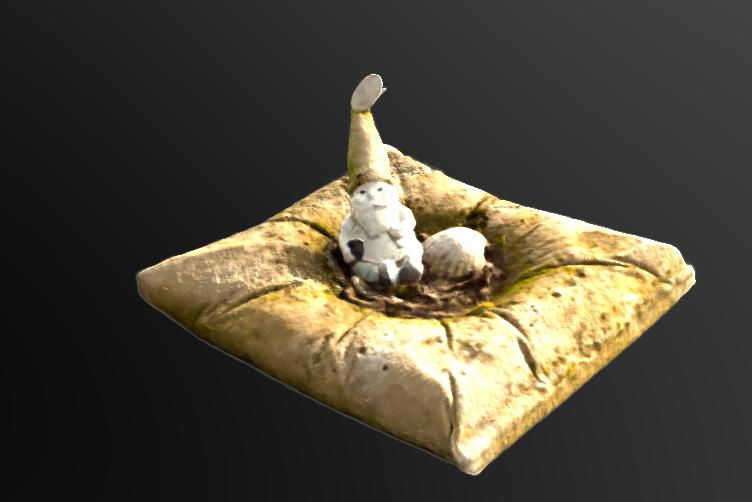}&\includegraphics[width=40pt,trim=145 60 145 60,clip]{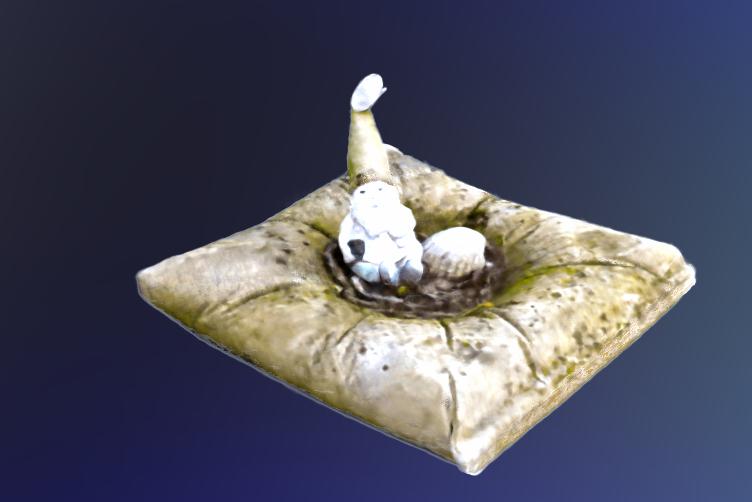}&\includegraphics[width=40pt,trim=145 60 145 60,clip]{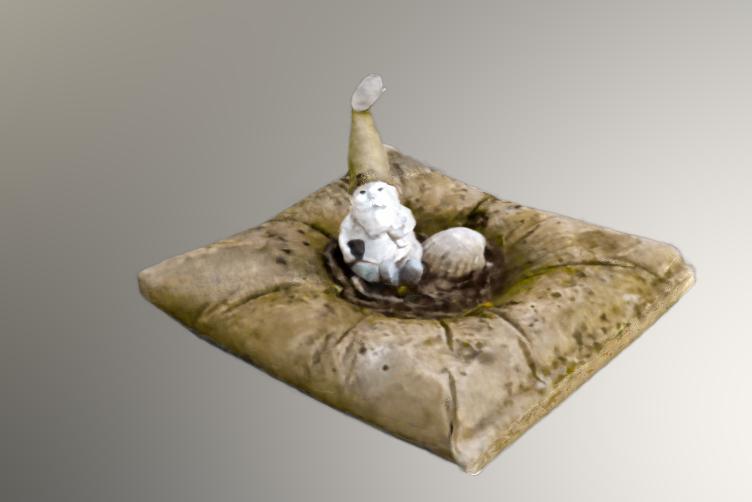}\\\midrule %
\tiny GT&\tiny Re{-}render - Illum 0&\tiny Illum 1&\tiny View + Illum 1&\tiny View + Illum 0& \tiny GT&\tiny Re{-}render - Illum 0&\tiny Illum 1&\tiny View + Illum 1& \tiny View + Illum 0\\\midrule%
\includegraphics[width=40pt]{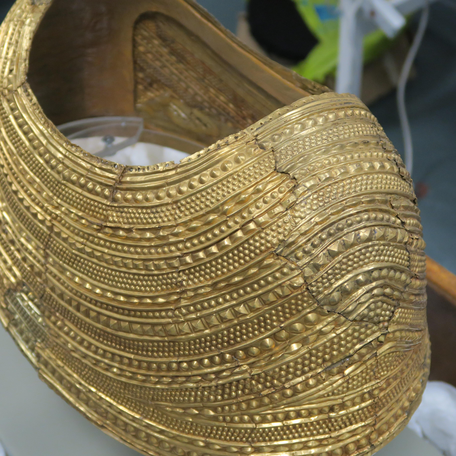}&\includegraphics[width=40pt]{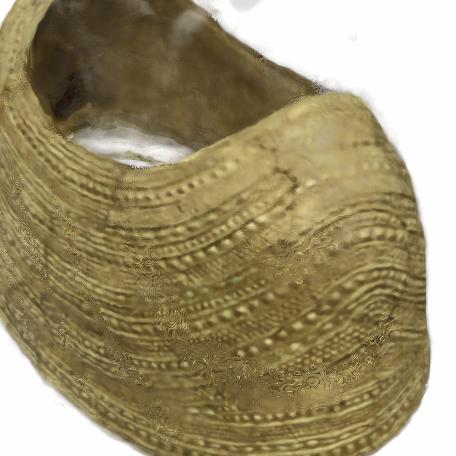}&\includegraphics[width=40pt]{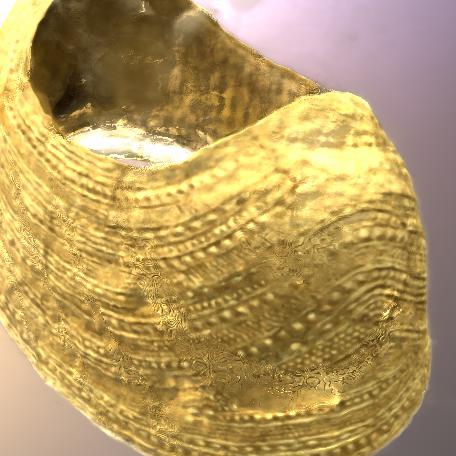}&\includegraphics[width=40pt]{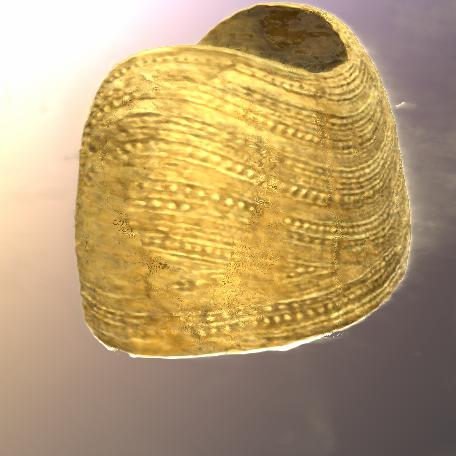}&\includegraphics[width=40pt]{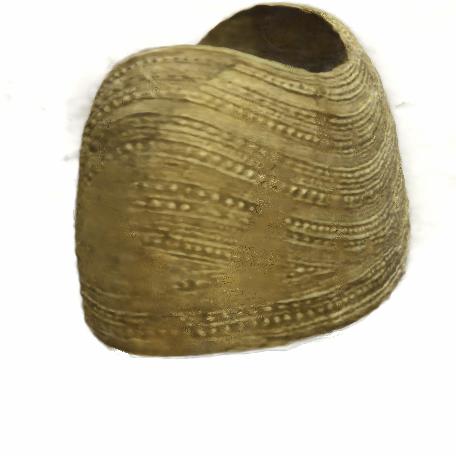}& %
\includegraphics[width=40pt,trim=55 15 63 0,clip]{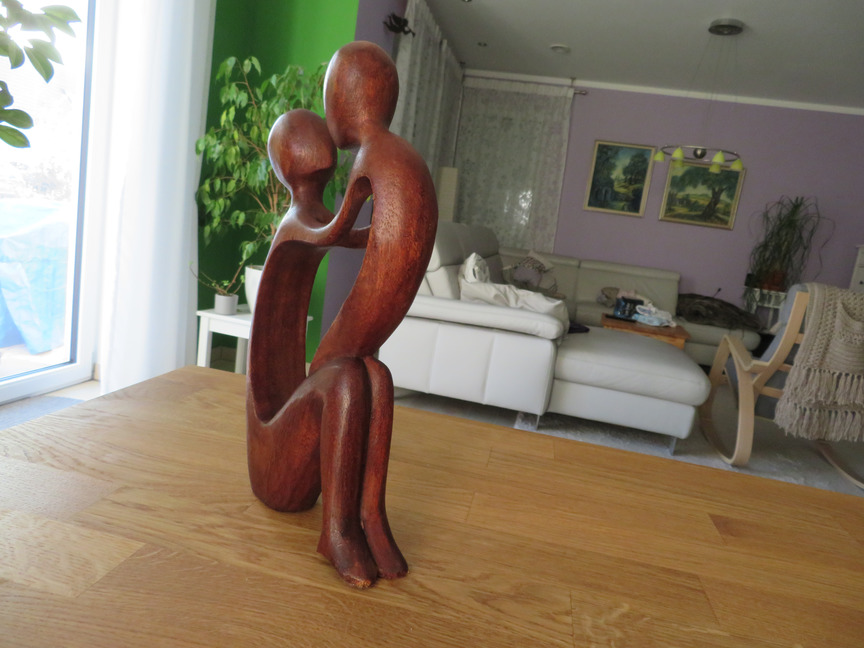}&\includegraphics[width=40pt,trim=55 15 63 0,clip]{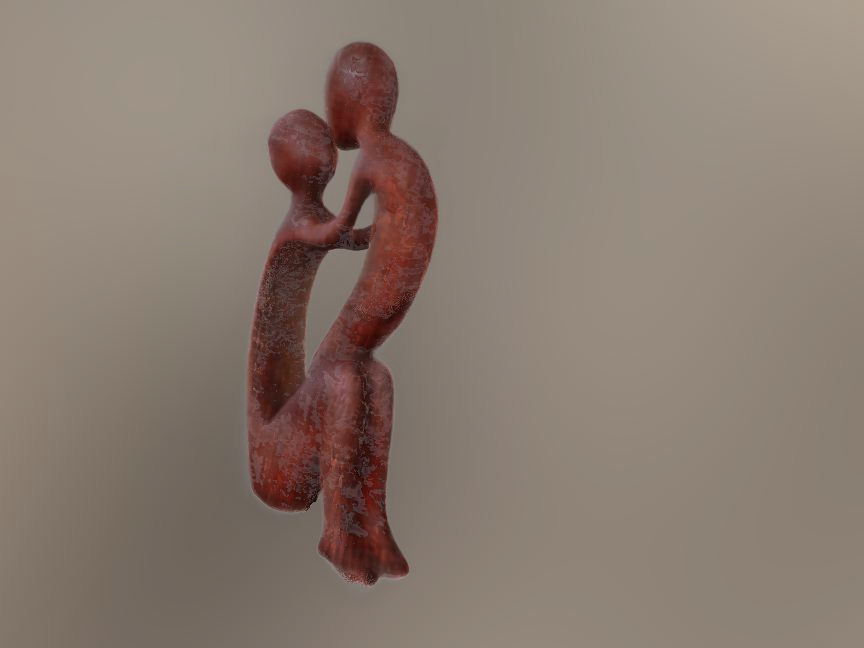}&\includegraphics[width=40pt,trim=55 15 63 0,clip]{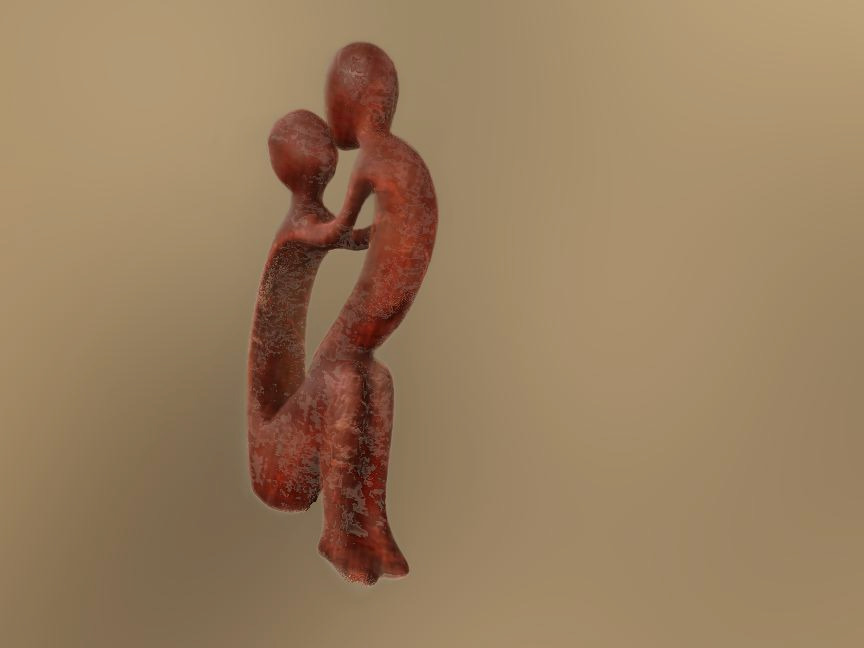}&\includegraphics[width=40pt,trim=55 15 63 0,clip]{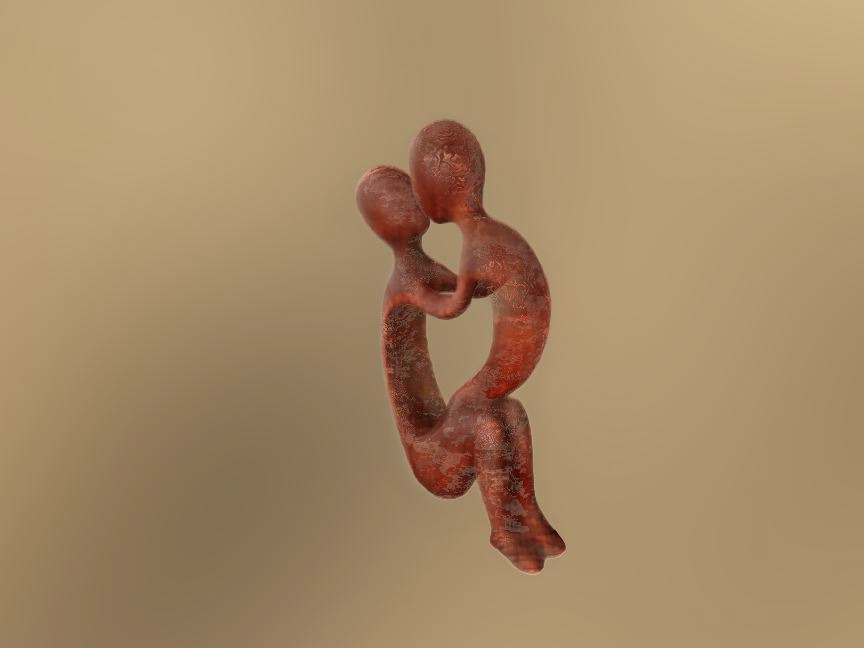}&\includegraphics[width=40pt,trim=55 15 63 0,clip]{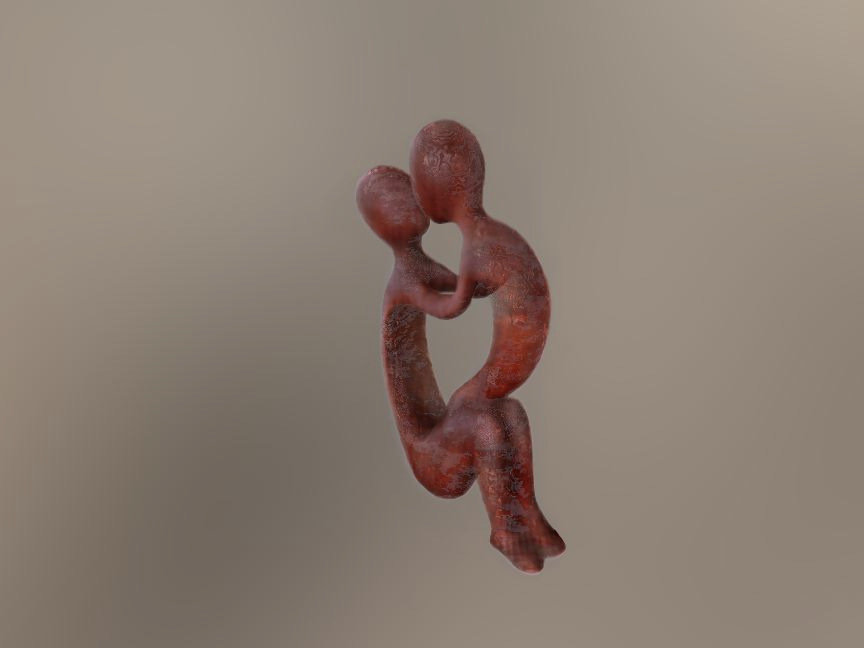}%
\\%
%
\end{tabular}%

    \titlecaption{Real World BRDF Decomposition and Relighting}{The decomposition produces plausible BRDFs, and re-rendered images are close to the ground truth input images. Note that the estimated parameters are hardly affected by the shadows visible in the input of the gnome scene. When relit with the estimated SG of a validation view, the appearance is well reproduced. Even in a different perspective or under completely novel, artificial illumination the recovered BRDF parameters result in convincing images.}
    \label{fig:rw_brdf}
\end{figure*}


The proposed method recovers shape, appearance, and illumination for relighting in unconstrained settings.
Our reconstruction and relighting performance on synthetic sets is measured against ground truth images and known BRDF parameters. 
For real-world examples, we present novel, relit views and compare the renderings with validation images excluded from training.
If the environment map for the validation image is known, we directly use this for relighting. Otherwise, we recover the unseen
illumination by optimizing the SGs through the frozen network in 1000 steps using stochastic gradient descent with a learning rate of $0.1$.
One-to-one comparisons with previous methods are challenging, as most methods use different capturing setups. 
We can, however, compare to the outcome of NeRF when trained on a similar scene. 
NeRF cannot relight the object under novel illumination, and even NeRF-w and our simplified NeRF-A baseline can only interpolate between seen illuminations.



We also perform an ablation study to study the influence of our novel training techniques.
We refer to the supplementary material for additional results and the extracted textured meshes for different scenes.

\inlinesection{Datasets.}
We use three synthetic scenes to showcase the quality of the estimated BRDF parameters.
We use three textured models
(Globe~\cite{Globe}, Wreck~\cite{carWreck}, Chair~\cite{Chair})
and render each model with a varying environment illumination per image. For a fixed illumination synthetic dataset, we use the Lego, Chair and Ship scenes from NeRF~\cite{mildenhall2020}.

We also evaluate using two real-world scenes from the British Museum's photogrammetry dataset: an Ethiopian Head~\cite{daniel_pett_ethiopian_head} and a Gold Cape~\cite{daniel_pett_gold_cape}. 
These scenes feature an object in a fixed environment with either a rotating object or a camera.
Additionally, we captured our own scenes under varying illumination at various times of day (Gnome, MotherChild). 

\begin{table}[!htb]
    \centering
    \footnotesize%

\extrarowheight=1pt
\aboverulesep=0pt
\belowrulesep=0pt
\setlength{\tabcolsep}{2pt}%
\begin{tabular}{lccc}%
    \toprule%
    \multicolumn{1}{c}{Method [PSNR\textuparrow]}  & \multicolumn{1}{c}{Diffuse} & \multicolumn{1}{c}{Specular} & \multicolumn{1}{c}{Roughness}  \\%
    \midrule%
    \scriptsize Li \etal & 1.06          & ---          & \thirdbest17.18   \\%
    \scriptsize Li \etal + NeRF & \thirdbest1.15          & ---          & \secondbest\textbf{17.28}          \\%
    \scriptsize Ours                 & \secondbest\textbf{18.24} & \secondbest\textbf{25.70}   & 15.00 \\\bottomrule%
\end{tabular}%

    \titlecaption{BRDF estimation}{Comparison with a recent state-of-the-art method in BRDF decomposition under environment illumination~\cite{Li2018a}. Li \etal: directly on test images, Li \etal + NeRF: NeRF trained on BRDFs from \cite{Li2018a}.}
    \label{tab:brdf_comparison}
\end{table}

\begin{table}[!htb]
    \centering
    \footnotesize%

\extrarowheight=1pt
\aboverulesep=0pt
\belowrulesep=0pt
\begin{tabular}{llcccc}%
    \toprule%
                                            & & \multicolumn{2}{c}{Fixed Illumination} & \multicolumn{2}{c}{Varying Illumination} \\
    \cmidrule(l){3-4} \cmidrule(l){5-6}
    & \multicolumn{1}{c}{Method} & \multicolumn{1}{c}{PSNR\textuparrow} & \multicolumn{1}{c}{SSIM\textuparrow} & \multicolumn{1}{c}{PSNR\textuparrow} & \multicolumn{1}{c}{SSIM\textuparrow}\\%
    \midrule%
    \parbox[t]{2mm}{\multirow{3}{*}{\rotatebox[origin=c]{90}{\scriptsize Syn.}}} & %
    \scriptsize NeRF %
    & \secondbest\textbf{34.24}   & \secondbest\textbf{0.97}   
    & 21.05                & 0.89\\
    & \scriptsize NeRF-A %
    & \thirdbest32.44       & \secondbest\textbf{0.97}   
    & \secondbest\textbf{28.53}                & \thirdbest0.92\\
    & \scriptsize Ours                        %
    & 30.07             & \thirdbest0.95      
    & \thirdbest27.96               & \secondbest\textbf{0.95} \\\midrule
    \parbox[t]{2mm}{\multirow{3}{*}{\rotatebox[origin=c]{90}{\scriptsize Real}}} & %
    \scriptsize NeRF %
    & \thirdbest23.34             & \thirdbest0.85   
    & 20.11                & 0.87 \\
    & \scriptsize NeRF-A %
    & 22.87             & 0.83   
    & \secondbest\textbf{26.36}                & \thirdbest{0.94} \\
    & \scriptsize Ours          %
    & \secondbest\textbf{23.86}             & \secondbest\textbf{0.88}      
    & \thirdbest25.81                & \secondbest\textbf{0.95}\\\bottomrule
\end{tabular}%

    \titlecaption{Novel view synthesis}{Comparison with NeRF and NeRF-A on novel view synthesis (with relighting in varying illumination). Notice NeRF and NeRF-A is not capable of relighing in any unseen illuminations, nor is an extraction of a textured mesh from the network easily possible.}
    \label{tab:nerf_novel_view_comparisons}
\end{table}

\inlinesection{BRDF decomposition results.} %
\fig{fig:syn_brdf} shows exemplary views and decomposition results of the synthetic Car Wreck and Chair scenes. In all cases, we observe the estimated re-renderings to be very similar
to GT.
The estimated BRDF parameters may not match perfectly in some places compared to the GT, but given the purely passive unknown illumination setup, they still reproduce the GT images.
Causes for deviations are the inherent ambiguity of the decomposition problem as well as the differences in shading based on SG \vs the high-resolution GT environment map.

Several sub-tasks of the unconstrained shape and BRDF decomposition problem have been addressed by earlier works. 
Unfortunately, trying to recover parameters separately or sequentially, \eg geometry, BRDF, or illumination, often fails in challenging scenes.  
We show that COLMAP fails to reconstruct a plausible geometry for some of our data sets in the supplementary material . If the following stages rely on accurate geometry, the pipeline cannot recover meaningful material properties from the inaccurate shape.
We also tried recovering the BRDF parameters (diffuse and roughness) for each image separately using the work of Li \etal~\cite{Li2018a} followed by NeRF to handle the view interpolation. 
To run on view independent BRDF parameters, we adapted NeRF accordingly.
However, NeRF fails to create a coherent geometry, as each image results in drastically different BRDF parameter maps. 

We, therefore, conclude that joint optimization of shape and SVBRDF is essential
for this extremely ambiguous problem. Quantitative comparisons with Li \etal are shown in \tbl{tab:brdf_comparison}.
These are average PSNR results over our synthetic datasets (Globe, Wreck, and Chair).
We decompose our basecolor into \emph{diffuse} and specular to enable comparison with Li \etal, which uses a \emph{diffuse} and roughness parameterization.
It is worth noting that Li \etal here is a weak baseline, but the closest available, as their method expects a flash light in conjunction to the environment illumination. However, as most scenes are captured with an outside environment illumination, the flash will be barely noticeable due to the strong sun light. 


\inlinesection{Relighting and novel view synthesis.}
In \fig{fig:rw_brdf}, novel views and plausible relighting in unseen environments are shown for our real-world data sets.
The relighted images are visually close to the held-out validation images.
Furthermore, a novel view can be relighted with the lighting from a different view.
Note that some fine details are missing in the reconstructions of the Gold Cape, which is caused by small inaccuracies in the camera registration. 
Also the MotherChild model is missing some highlights especially at grazing angles, which can be attributed to the limitations of the SG based rendering model.

While no ground truth BRDF exists, the estimated parameters for the Gnome (\fig{fig:rw_brdf}) seems plausible. 
The material is correctly classified as non-metallic (black metalness map), has a higher roughness, and the normal also aligns well with the shape. In the central valley, where dirt is collected, the BRDF parameters increase in roughness compared to the clean, smooth concrete pillow surface. The color is also captured well, and in re-rendering, the similarity to the ground-truth is evident.

\begin{figure}[!htb]
    \centering
    \normalsize%
\renewcommand{\arraystretch}{0.6}%
\setlength{\tabcolsep}{1pt}%
\begin{tabular}{@{}l>{\centering\arraybackslash}m{45pt}>{\centering\arraybackslash}m{45pt}>{\centering\arraybackslash}m{45pt}>{\centering\arraybackslash}m{45pt}@{}}%
&\footnotesize Frame 3&\footnotesize Frame 6&\footnotesize Frame 9 &\footnotesize Relighting\\%
\midrule%
\rotatebox[origin=c]{90}{\footnotesize NeRF}&\includegraphics[width=45pt]{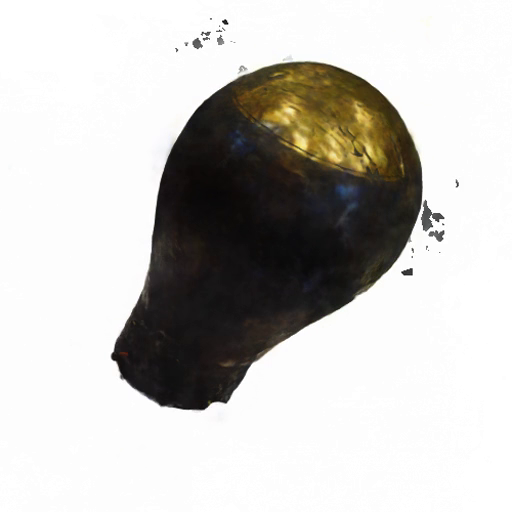}&\includegraphics[width=45pt]{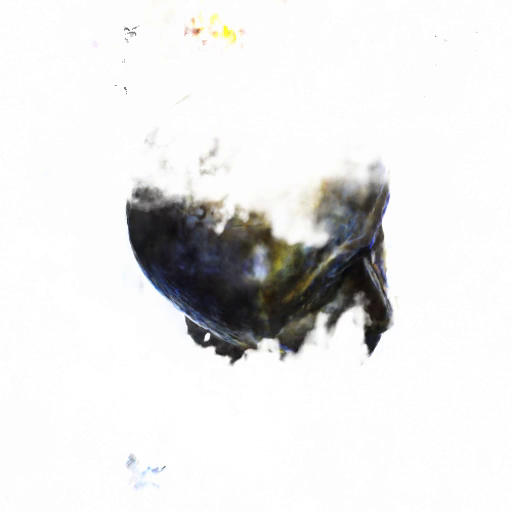}&\includegraphics[width=45pt]{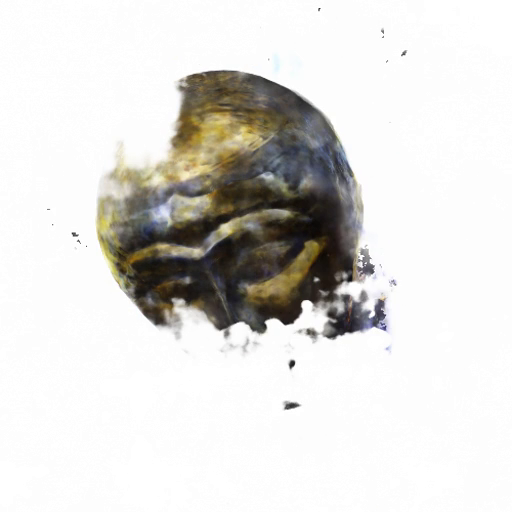}& Not Available\\%
\rotatebox[origin=c]{90}{\footnotesize NeRF-A}&\includegraphics[width=45pt]{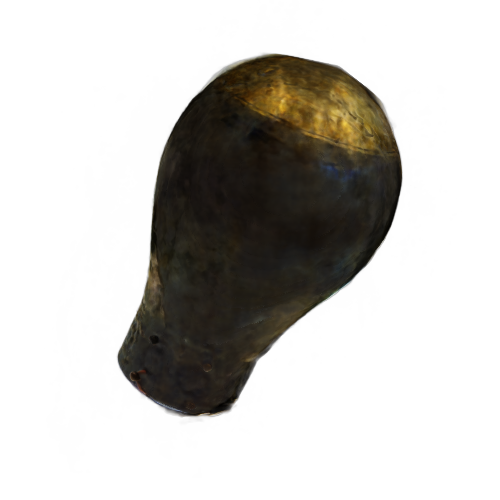}&\includegraphics[width=45pt]{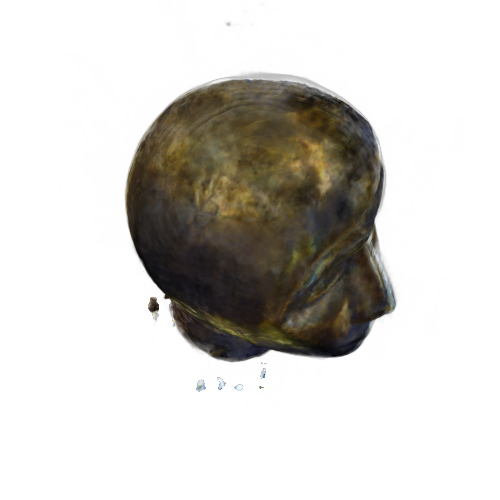}&\includegraphics[width=45pt]{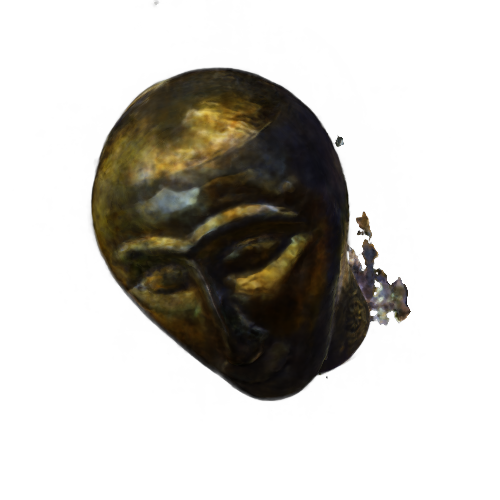}& Not Available\\%
\rotatebox[origin=c]{90}{\footnotesize Ours}&\includegraphics[width=45pt]{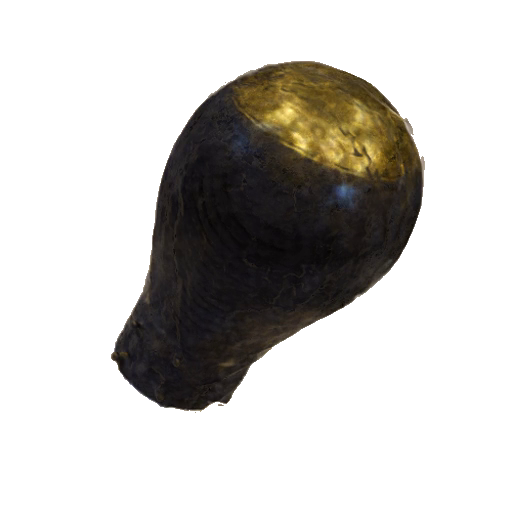}&\includegraphics[width=45pt]{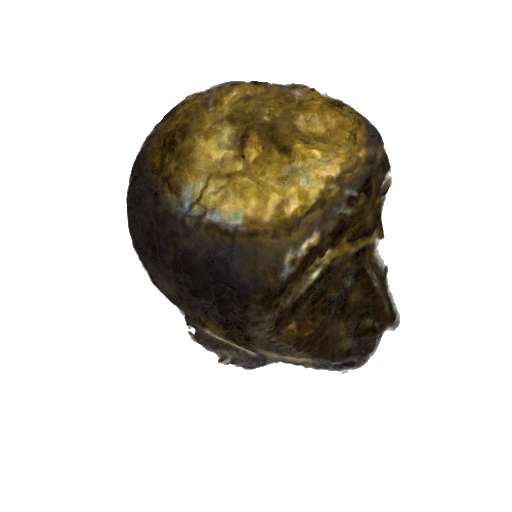}&\includegraphics[width=45pt]{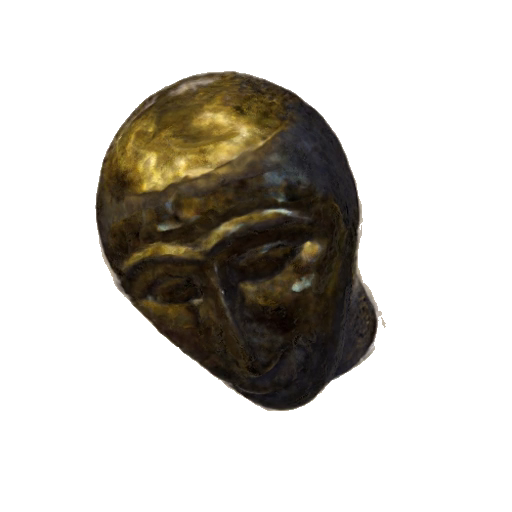}&\includegraphics[width=45pt]{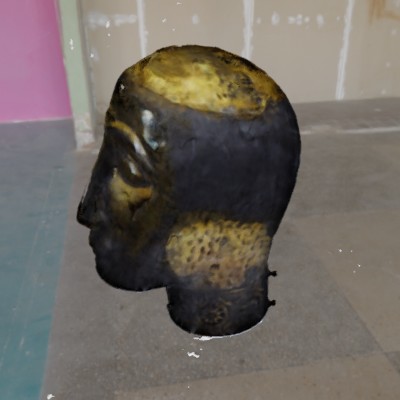}\\
\end{tabular}%

    \titlecaption{Novel View Comparison with NeRF and NeRF-A on real-world Ethiopian Head}{Notice the improved consistency in our method. NeRF introduces highlights as floaters in the radiance volume that inconsistently occlude the scene geometry in other views. Additionally, we showcase the quality in relighting the head with our method.}
    \label{fig:rw_nerf_novel_view_comparisons}
\end{figure}

Another evaluation focuses on the use of our method purely for novel view synthesis, with implicit relighting. 
In this setting, our method can be compared with NeRF~\cite{mildenhall2020} and also an extension to NeRF, called NeRF-A, which inspired from \cite{martinbrualla2020nerfw}. NeRF-A models the appearance change per image in a 48 dimensional latent vector. It is worth noting that NeRF-A is a strong baseline as the task is simpler compared to NeRD and it is only capable of relighting within known scenes. On fixed illumination scenes, NeRF-A is not capable of relighting.
\tbl{tab:nerf_novel_view_comparisons} shows the quantitative results over multiple datasets,
real-world (Real) and synthetic (Syn.), on the test views wherein the ``Fixed Illumination'' case only novel view synthesis is performed, and in the ``Varying Illumination'' case, novel view synthesis and relighting. Here, ``Varying Illumination'' also refers to case where
the object is rotating w.r.t the camera and therefore the relative illumination is varying.
The corresponding datasets for the fixed and synthetic case are from NeRF (Ship, Lego, Chair), and for varying, we use ours (Globe, Wreck, Chair). For the real-world comparison, Cape provide fixed illumination, and the Gnome, Head and MotherChild 
scenes are recorded in varying environments.
PSNR and SSIM results show that NeRF performs quite poorly in varying illumination cases. NeRF-A on the other hand is a strong baseline in the varying illumination case which we mostly match or surpass while solving a more challenging problem which allows for more flexible relighting use cases.

\fig{fig:rw_nerf_novel_view_comparisons} shows the novel view synthesis results of NeRF, NeRF-A and NeRD (Ours) on the Ethiopian Head real-world scene.
The object rotates in front of the camera. We, therefore, compose the Head on a white background in \fig{fig:rw_nerf_novel_view_comparisons} as NeRF cannot handle a static background with a fixed camera. During training, both models recreate the input quite closely. However, in the test views, NeRF added spurious geometry to mimic highlights for specific camera locations, which are not seen by other cameras in the training set. NeRF-A can expess the relative illumination change in the appearance embedding and can improve the reconstruction quality compared to NeRF. However, as only a single illumination type is seen NeRF-A is still not capable of relighting under arbitrary illumination.
Due to our physically motivated setup with the explicit decomposition of shape, reflectance, and illumination, these issues are almost completely removed.
Our method creates convincing object shapes and reflection properties, which, in addition, allow for relighting in novel settings.

Overall, it is evident that NeRF will not work with varying illuminations, clearly demonstrating the advantage of our more flexible decomposition. 
It is also worth noting that even if an appearance embedding as in NeRF-w~\cite{martinbrualla2020nerfw} our our simplified NeRF-A baseline is used, the method can only interpolate between seen illuminations. Our model is capable of relighting even if the scene was only captured in a single fixed illumination.

\begin{table}[!htb]
    \centering
    \footnotesize%

\setlength{\tabcolsep}{2pt}%
\begin{tabular}{@{}p{18mm}ccccc@{}}%
    \toprule%
    \scriptsize Method                & \scriptsize Base Color      & \scriptsize Metalness       & \scriptsize Roughness       & \scriptsize Normal           & \scriptsize Re{-}Render     \\%
    \midrule%
    \scriptsize w/o Grad. Normal         & 0.1264          & 0.1203          & 0.3192          & 0.1664                    & 0.0893          \\%
    \scriptsize w/o Com. BRDF       & 0.1828          & 0.2496          & 0.2827          & 0.0089                  & 0.0759          \\%
    \scriptsize w/o WB                    & 0.1059          & 0.0870          & 0.2754          & 0.0087            & 0.0655 \\ %
    \midrule
    \scriptsize Full Model                 &\textbf{0.0796} &\textbf{0.0784}   & \textbf{0.2724} &\textbf{0.0084}   &\textbf{0.0592}\\\bottomrule%
\end{tabular}%


    \titlecaption{Ablation Study}{The MSE loss on 10 test views with ablation of gradient (grad.) normals, compressed (Com.) BRDF and white balancing (WB) on the globe dataset.}
    \label{tab:ablation_study}
\end{table}

\inlinesection{Ablation study.}
In \tbl{tab:ablation_study}, we ablate the gradient-based normal estimation, the BRDF interpolation in a compressed space, and incorporating the white balancing in the optimization.
We perform this study on the Globe scene as it contains reflective, metallic, and diffuse materials and fine geometry.
One of the largest improvements stems from the addition of gradient-based normals.
The coupling of shape and normals improves the BRDF and illumination separation. Normals cannot be rotated freely to mimic specific reflections. 
The compressed BRDF space also improves the result, especially in the metalness parameter estimation.
This indicates that the joint optimization of the encoder/decoder network $\text{N}_{\phi_2}$ effectively optimizes similar materials across different surface samples. 
The white balancing fixes the absolute intensity and color of the SGs, which indirectly forces the BRDF parameters into the correct range.


\section{Conclusion}
\vspace{-2mm}

In this work, we tackle an extremely challenging problem of decomposing shape, illumination, and reflectance by augmenting coordinate-based radiance fields with explicit representations for the BRDF and the illumination. 
This decomposition renders our approach significantly more robust than simple appearance-based representations, or other multi-view stereo approaches \wrt changes in the illumination, cast shadows, or glossy reflections. 
Additionally, we propose a method to link the surface normal to the object's actual shape during optimization.
This link allows a photometric loss to alter the shape by backpropagation through differentiable rendering.
Our method enables realistic real-time rendering \emph{and} relighting under arbitrary unseen illuminations 
via explicit mesh extraction from the neural volume.

While the results from the method are convincing, there exists several limitations.
Currently, no explicit shadowing is modeled while the object is optimized.
Especially in scenes with a static environment illumination and deep crevices, a shadow will be baked into the diffuse albedo.
Additionally, the chosen SGs environment model helps in a stable and fast shading evaluation but is often limiting when high-frequency light effects are present in a scene. 
A different, maybe implicit environment representation might produce better results, but it would need to support efficient BRDF evaluation.

\ificcvfinal
\inlinesection{Acknowledgement}
This work was funded by the Deutsche Forschungsgemeinschaft (German Research Foundation) - Projektnummer 276693517 - SFB 1233 and by EXC number 2064/1 – Project number 390727645.
\fi


\ifthenelse{\boolean{addsupplements}}{
\part*{Supplementary}
\setcounter{table}{0}
\setcounter{figure}{0}
\renewcommand{\thetable}{A\arabic{table}}
\renewcommand{\thefigure}{A\arabic{figure}}

\inlinesection{Implementation details.}
The main network $\text{N}_{\theta_1}$/$\text{N}_{\phi_1}$ uses 8 MLP layers with a feature dimension of 256 and ReLU activation. The input coordinate $\vect{x}$ is transformed by the Fourier output $\gamma(\vect{x})$ with 10 bands to 63 features. 
For the \emph{sampling network}, the output from the main network is then transformed to the density $\sigma$ with a single MLP layer without any activation. The flattened 192 SGs parameters $\Gamma^j$ are compacted to 16 features using a fully connected layer ($\text{N}_{\theta_2}$) without any activation. As the value range can be large in the real-world illuminations, the amplitudes $\alpha$ are normalized to $[0,1]$. The main network output is then concatenated with the SGs embeddings and passed to the final prediction network. Here, an MLP with ReLU activation is first reducing the joined input to 128 features. The final color prediction $c^j$ is handled in the last layer without activation and an output dimension of 3.

The \emph{decomposition network} uses the output from $\text{N}_{\phi_1}$ and directly predicts the direct color $d$ and the density in a layer without activation and 4 output dimensions (RGB+$\sigma$). The main network output is then passed to several ReLU activated layers which handle the BRDF compression $\text{N}_{\theta_2}$. The feature outputs are as followed: 32, 16, 2 (no activation), 16, 16, 5 (no activation). The final five output dimensions correspond to the number of parameters of the BRDF model. The compressed embedding with two feature outputs is regularized with a $\mathcal{L}^2$ norm with a scale of $0.1$ and further clipped to a value range of $-40$ to $40$ to keep the value ranges in the beginning stable. 
Per batch, 1024 rays are cast into a single scene.

\inlinesection{Loss and learning rate schedule.}
For adjusting the losses and learning rate during the training, we use exponential decay: $p(i; v, r, s) = v r^\frac{i}{s}$.
The learning rate then uses $p(i; 0.000375, 0.1, 250000)$, the direct color $d$ loss is faded out using $p(i; 1, 0.75, 1500)$ and the alpha loss is faded in using $p(i; 1, 0.9, 5000)$. During the first 1000 steps, we also do not optimize the SGs parameters and first use the white balancing only to adjust the mean environment strength, as this step also sets the illumination strength per image based on the exposure values.

\inlinesection{Mesh extraction.}
The ability to extract a consistent textured mesh from NeRD after training is one of the key advantages of the decomposition approach and enables real-time rendering and relighting.
This is not possible with NeRF-based approaches where the view-dependent appearance is directly baked into the volume.

We use the following four general steps to extract textured meshes:
\begin{enumerate}
    \item A very dense point cloud representation of the surface is extracted. This step utilizes the same rendering functions used during training, which ensures that the resulting 3D coordinates are consistent with the training. To generate the rays for the rendering step, we sample the \emph{decomposition network} for $\sigma$ in a regular grid within the view volume determined by the view frustums of the cameras. We construct a discrete PDF from this grid, which is then sampled to generate about 10 million points where $\sigma$ is high. The rays are constructed by following the normals at those points to get the ray-origins. We use the slightly jittered inverted normals as ray directions. See \fig{fig:meshing}~$a$ to $c$ for visualizations of this step.
    \item For meshing, we use the Open3D~\cite{Zhou2018} implementation of the Poisson surface reconstruction algorithm \cite{kazhdan2006poisson} using the normals from NeRD. Before meshing, we perform two cleanup steps: First, we reject all points where the accumulated opacity along a ray is lower than $0.98$. Secondly, we perform statistical outlier removal from Open3D. Those steps are visualized in \fig{fig:meshing}~$d$ and $e$
    \item We use Blender's~\cite{blender} \emph{Smart UV Project} to get a simple UV-unwrapping for the mesh. Reducing the mesh resolution beforehand is an optional step that reduces the computational burden for using the mesh later. This is also done using Blender via \emph{Decimate Geometry} or the \emph{Voxel Remesher}. 
    \item We bake the surface coordinates and geometry normals into a floating-point texture of the desired resolution. The textures are generated by generating and rendering one ray per texel to look up the BRDF parameters and shading normals with NeRD. A result is show in \fig{fig:meshing}~$f$.
\end{enumerate}

\begin{figure}
    \centering
    \input{supplement_figures/mesh_creation/mesh_creation}
    \titlecaption{Mesh Generation}{a) Cameras and view frustum. b) Points sampled where $\sigma$ was high. c) Rendered point-cloud with basecolor. d) Outlier removal. e) Mesh with vertex-colors. f) Low res mesh with material textures.}
    \label{fig:meshing}
\end{figure}

\inlinesection{Dataset details.}
In \tbl{tab:dataset_details}, we list the trained resolution, the number of total images, and the test train split for each dataset. Exemplary images of the real-world datasets are shown in \fig{fig:datasets}.

\begin{table}
    \centering
    \footnotesize
    \begin{tabular}{lcccc}
        \toprule%
        Dataset & Resolution (W$\times$H) & \#Images & \#Train & \#Test   \\ \midrule%
        \scriptsize Globe & $400 \times 400$ & 210 & 200 & 10 \\ %
        \scriptsize Car Wreck & $400 \times 400$ & 210 & 200 & 10 \\ %
        \scriptsize Chair & $400 \times 400$ & 210 & 200 & 10 \\ %
        \scriptsize Ethiopian Head & $500 \times 500$ & 66 & 62 & 4 \\ %
        \scriptsize Gold Cape & $456 \times 456$ & 119 & 111 & 8 \\ %
        \scriptsize Gnome & $752 \times 502$ & 103 & 96 & 7 \\ %
        \scriptsize MotherChild & $864 \times 648$  & 104 & 97 & 7 \\ %
         \bottomrule%
    \end{tabular}
    \titlecaption{Dataset Overivew}{Overview about the resolution and number of images used for training.}
    \label{tab:dataset_details}
\end{table}

\begin{figure}
    \centering
    \begin{tabular}{@{}rr>{\centering\arraybackslash}m{0.85\linewidth}@{}}
    \rotatebox[origin=c]{90}{\footnotesize Gnome}&\rotatebox[origin=c]{90}{\scriptsize Varying Illumination}&\includegraphics[width=\linewidth]{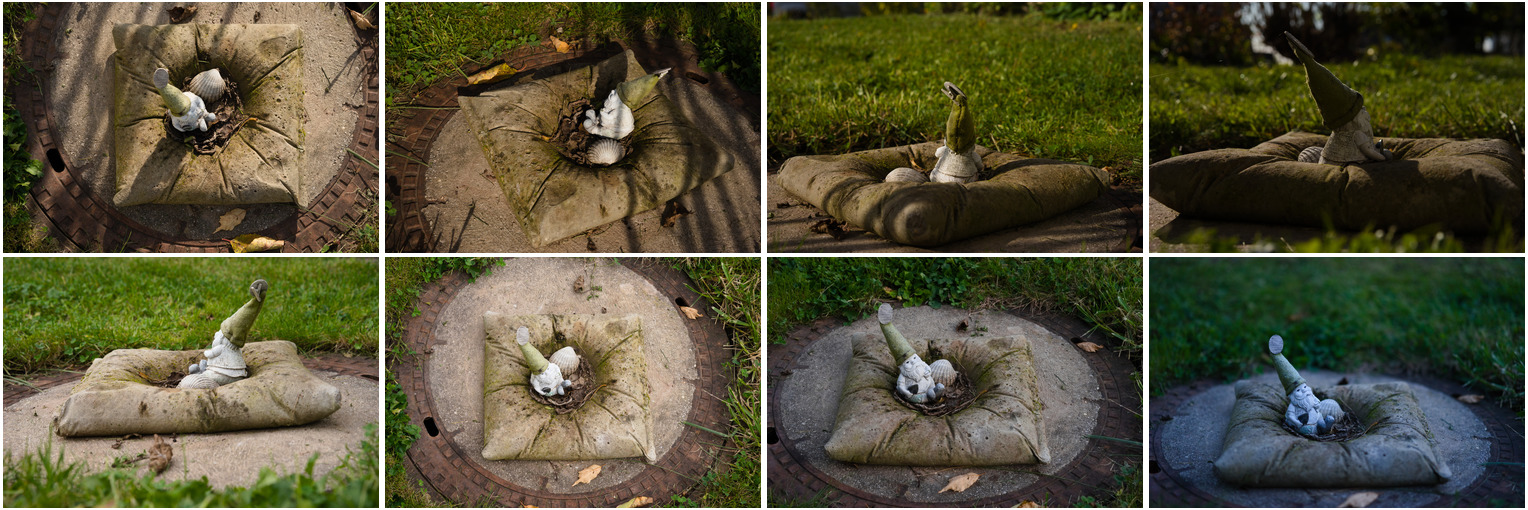}\\
    \rotatebox[origin=c]{90}{\footnotesize MotherChild}&\rotatebox[origin=c]{90}{\scriptsize Varying Illumination}&\includegraphics[width=\linewidth]{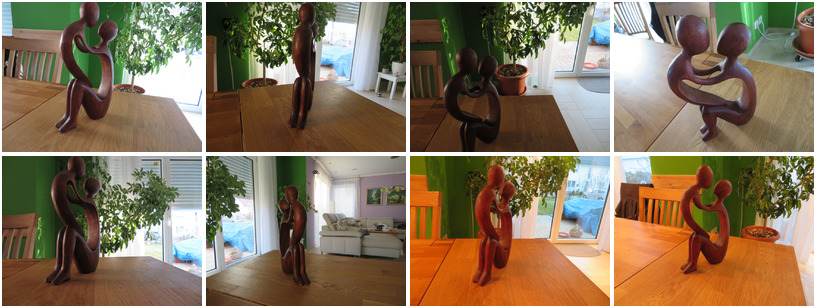}\\
    \rotatebox[origin=c]{90}{\footnotesize Cape}&\rotatebox[origin=c]{90}{\scriptsize Static Illumination}&\includegraphics[width=\linewidth]{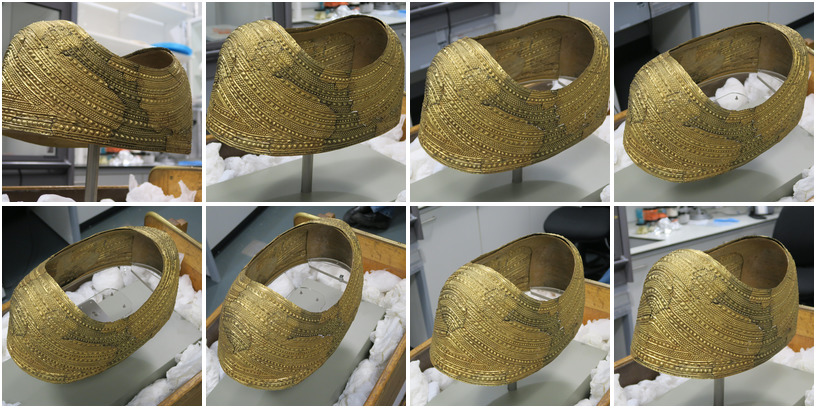}\\
    \rotatebox[origin=c]{90}{\footnotesize EthiopianHead}&\rotatebox[origin=c]{90}{\scriptsize Static Illumination}&\includegraphics[width=\linewidth]{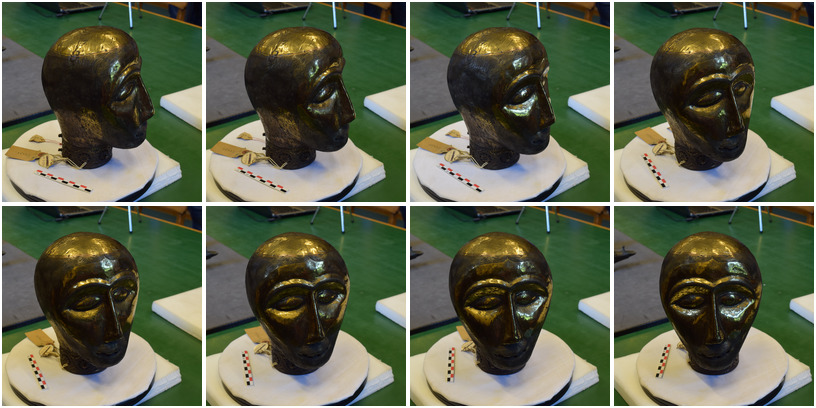}\\
    \end{tabular}
    \titlecaption{Datasets}{Exemplary images of our real-world datasets. Notice the challenging environment illumination in the varying illumination scenes. The gnome dataset even features shadows from the environment.}
    \label{fig:datasets}
\end{figure}

\begin{figure}
    \centering
    \normalsize%
\renewcommand{\arraystretch}{0.6}%
\setlength{\tabcolsep}{0pt}%
\begin{tabular}{@{}l>{\centering\arraybackslash}m{70pt}>{\centering\arraybackslash}m{70pt}@{}}%
\toprule%
& \footnotesize Gold Cape&\footnotesize Ethiopian Head \\%
\midrule%
\rotatebox[origin=c]{90}{\footnotesize GT} & \includegraphics[width=70pt]{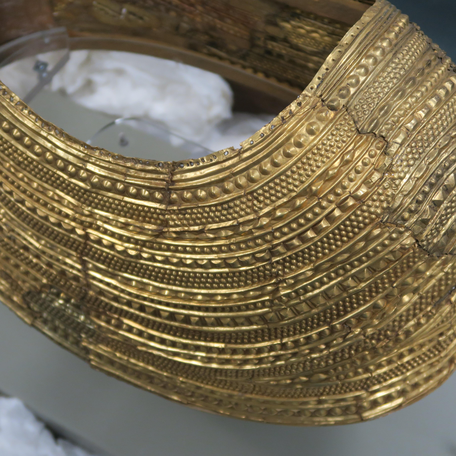} & \includegraphics[width=70pt]{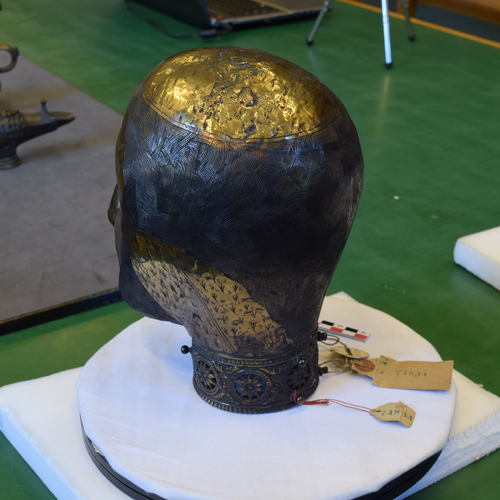} \\ %
\rotatebox[origin=c]{90}{\footnotesize Ours} & \includegraphics[width=70pt]{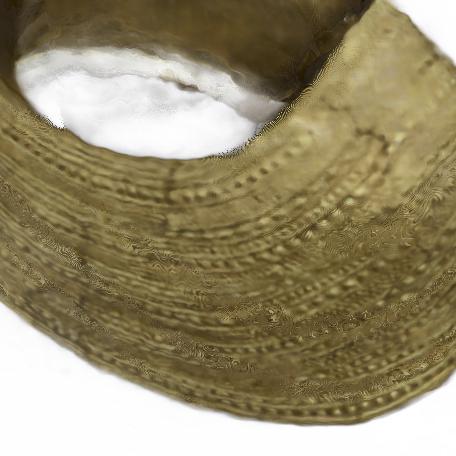} & \includegraphics[width=70pt]{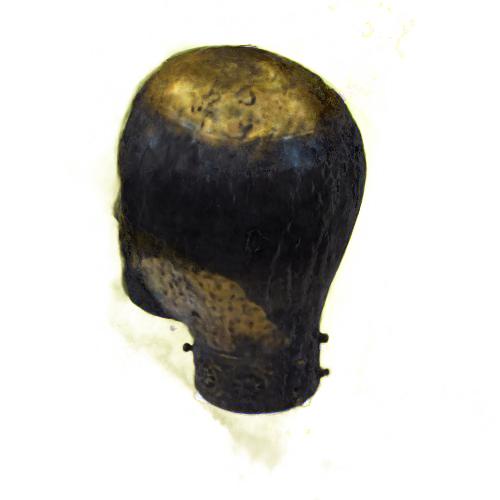} \\ %
%
\end{tabular}%

    \titlecaption{Real World Novel View Synthesis}{Comparison on real-world samples with novel view synthesis on test dataset views.}
    \label{fig:rw_novel_view}
\end{figure}

\begin{figure}
    \centering
    \normalsize%
\renewcommand{\arraystretch}{0.8}%
\setlength{\tabcolsep}{2pt}%
\begin{tabular}{@{}l>{\centering\arraybackslash}m{50pt}>{\centering\arraybackslash}m{50pt}>{\centering\arraybackslash}m{50pt}>{\centering\arraybackslash}m{50pt}@{}}%
\toprule%
& \footnotesize Chair & \footnotesize Gnome & \footnotesize Head & \footnotesize Cape\\%
\midrule%
\rotatebox[origin=c]{90}{\footnotesize Ours}&%
\includegraphics[width=50pt]{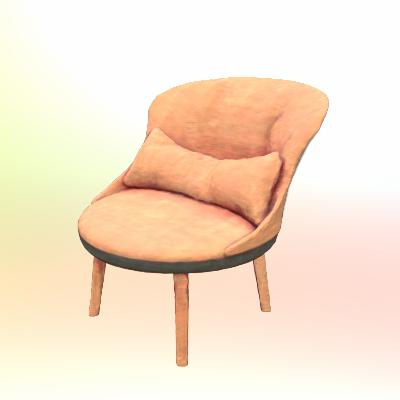}&%
\includegraphics[width=50pt]{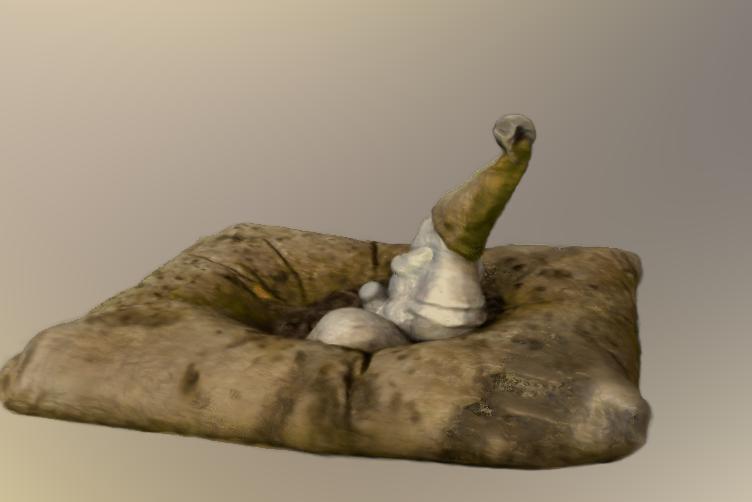}&%
\includegraphics[width=50pt]{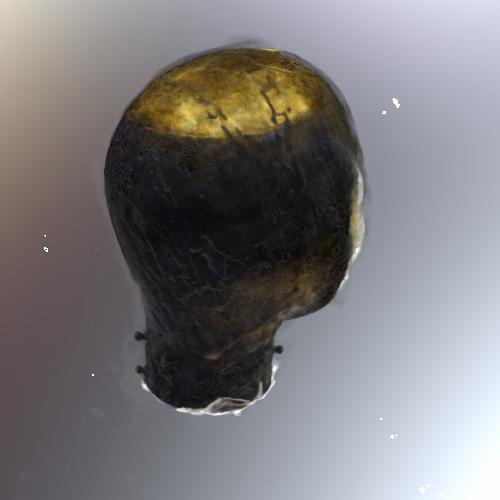}&%
\includegraphics[width=50pt]{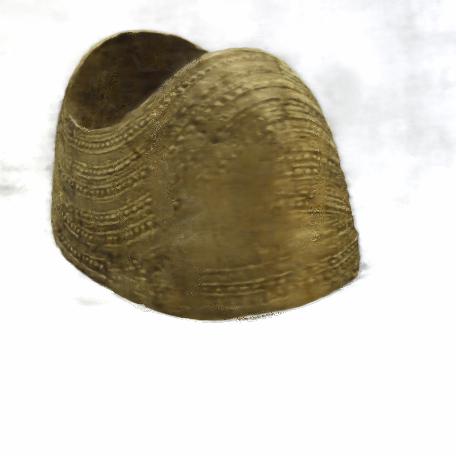}%
\\%
%
%
\rotatebox[origin=c]{90}{\footnotesize NeRF\cite{mildenhall2020}}%
&\includegraphics[width=50pt]{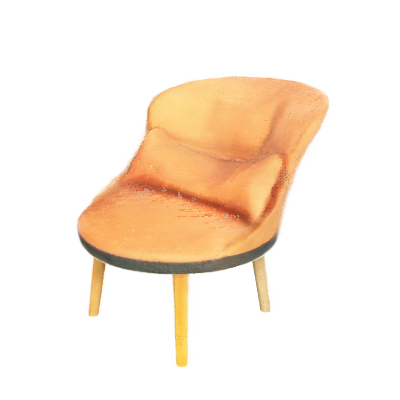}%
&\includegraphics[width=50pt]{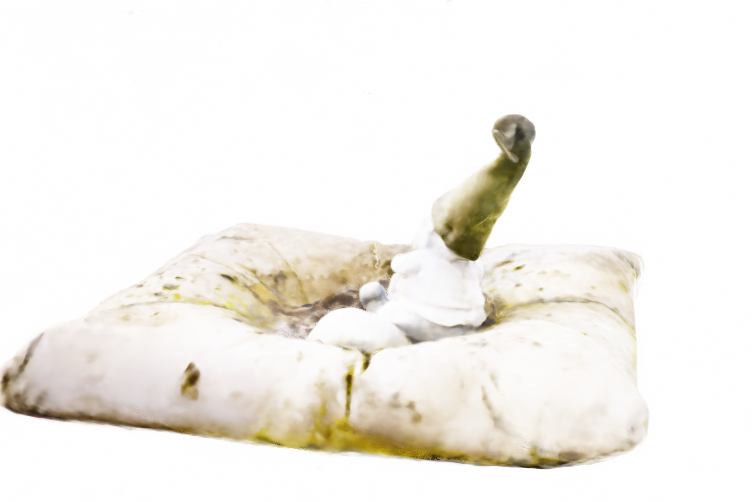}%
&\includegraphics[width=50pt]{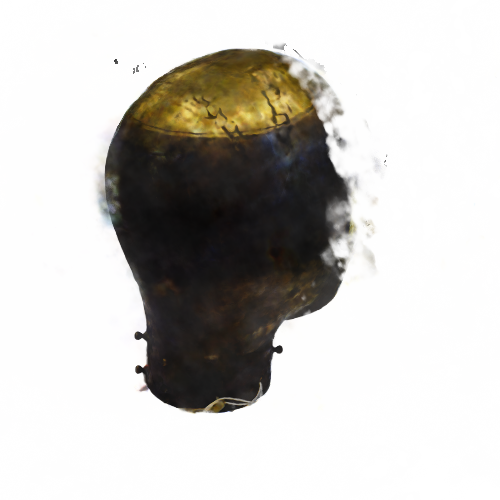}%
&\includegraphics[width=50pt]{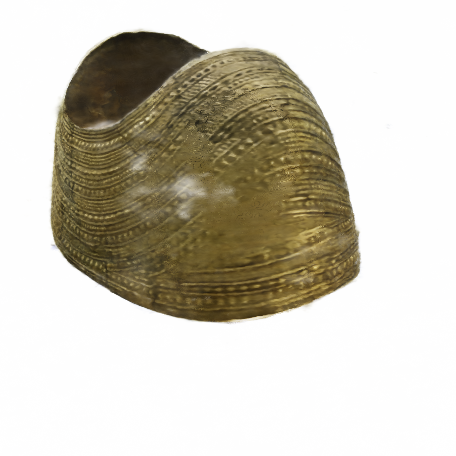}%
\\%
\rotatebox[origin=c]{90}{\footnotesize NeRF-A}%
&\includegraphics[width=50pt]{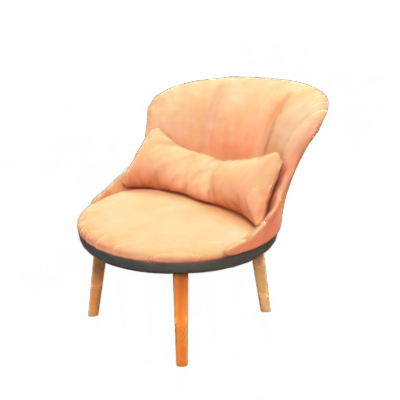}%
&\includegraphics[width=50pt]{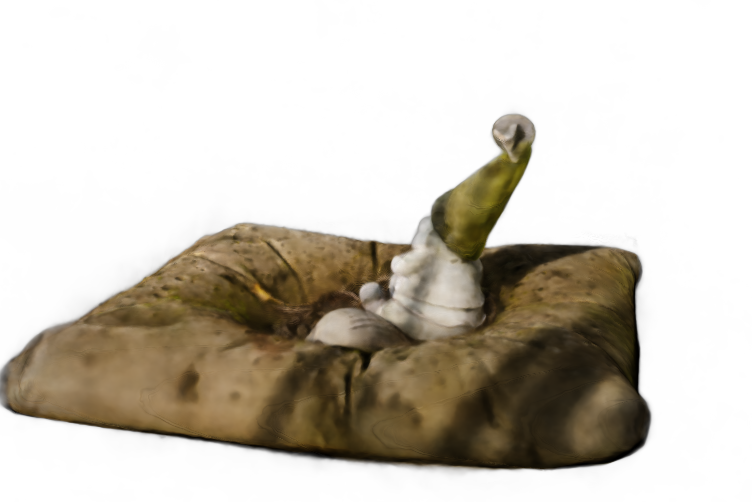}%
&\includegraphics[width=50pt]{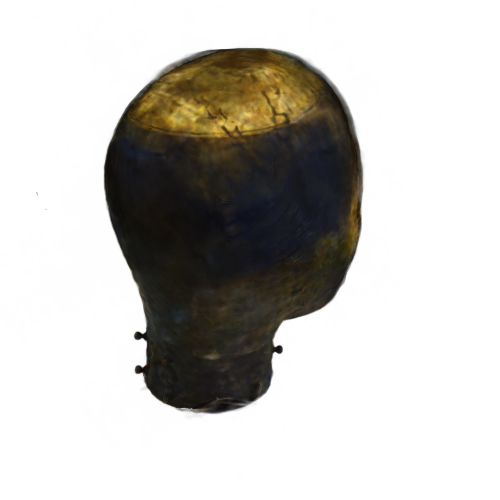}%
&\includegraphics[width=50pt]{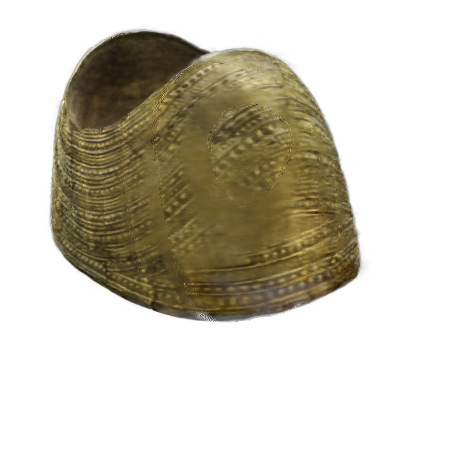}%
\\%
%
%
\rotatebox[origin=c]{90}{\footnotesize GT}&%
\includegraphics[width=50pt]{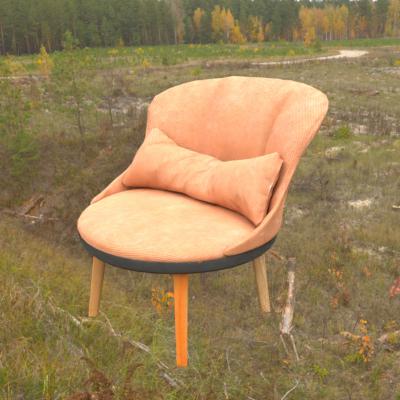}&%
\includegraphics[width=50pt]{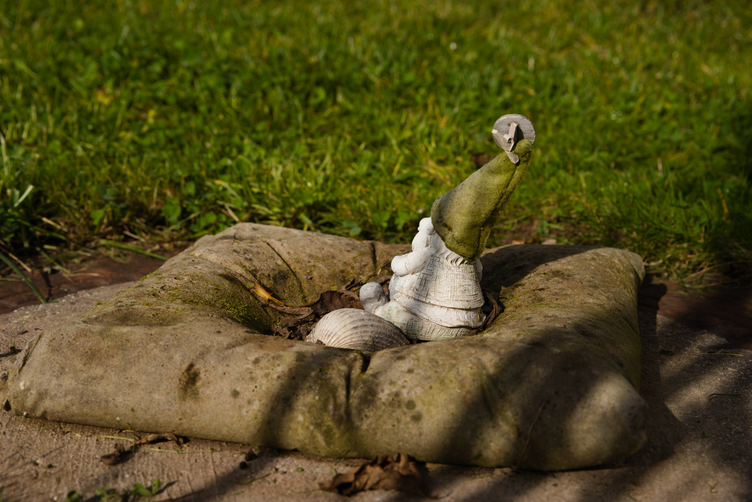}&%
\includegraphics[width=50pt]{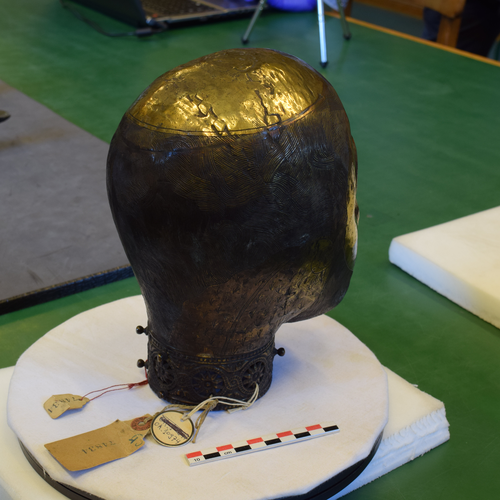}&%
\includegraphics[width=50pt]{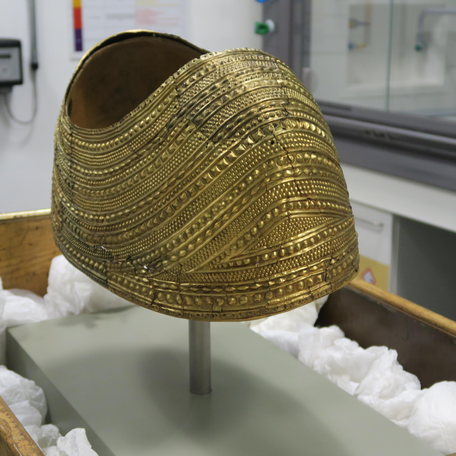}%
\\\bottomrule%
\end{tabular}

    \titlecaption{Comparison with NeRF and NeRF-A}{Comparison with NeRF and NeRF-A on various scenes. Here, it is evident that NeRF fails as expected on scenes with varying illumination (Chair, Gnome).}
    \label{fig:nerf_comparison}
\end{figure}

\begin{figure}
    \centering
    \begin{subfigure}[t]{0.45\linewidth}
        \centering
        \includegraphics[height=\linewidth, trim=550 0 100 0, clip]{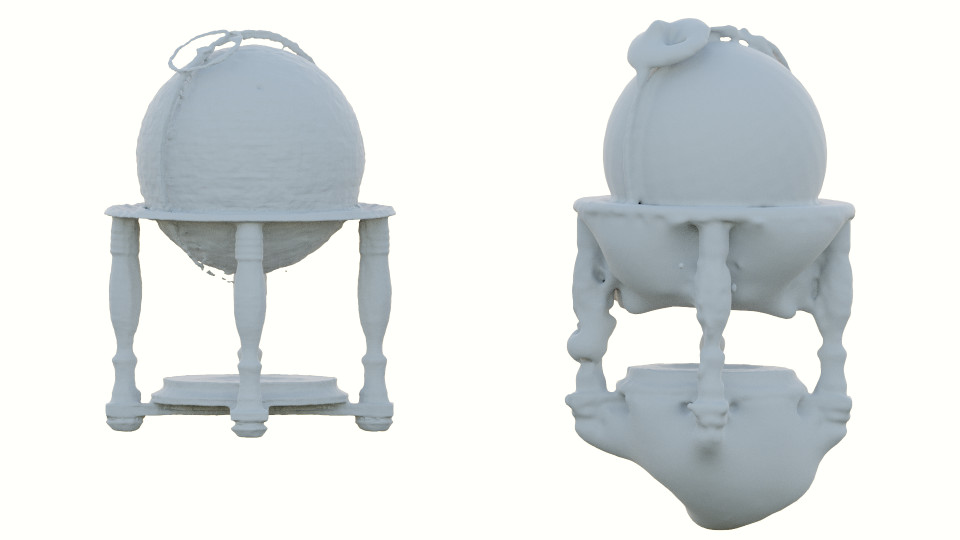}
        \titlecaption{COLMAP Reconstruction}{%
            COLMAP fails to recreate plausible geometry.
            }
        \label{fig:colmap_reconstruction}
    \end{subfigure}
    \hfill
    \begin{subfigure}[t]{0.45\linewidth}
        \centering
        \includegraphics[height=\linewidth, trim=100 0 550 0, clip]{figures/colmap_reconstruction/ColmapPaperSmall.jpg}
        \titlecaption{Ours}{%
            Our reconstruction can handle this complex scene.
        }
        \label{fig:our_reconstruction}
    \end{subfigure}
    \titlecaption{Geometry reconstruction}{%
            Comparison of a COLMAP reconstruction in our globe scene with varying illuminations.
        }
    \label{fig:geometry_reconstruction}
\end{figure}

\begin{figure*}
    \centering
    \normalsize%
\renewcommand{\arraystretch}{0.6}%
\setlength{\tabcolsep}{0pt}%
\begin{tabular}{@{}l>{\centering\arraybackslash}m{42pt}>{\centering\arraybackslash}m{42pt}>{\centering\arraybackslash}m{42pt}>{\centering\arraybackslash}m{42pt}>{\centering\arraybackslash}m{84pt}>{\centering\arraybackslash}m{42pt}>{\centering\arraybackslash}m{42pt}>{\centering\arraybackslash}m{42pt}>{\centering\arraybackslash}m{42pt}>{\centering\arraybackslash}m{42pt}@{}}%
&\scriptsize Base Color&\scriptsize Metalness&\scriptsize Roughness&\scriptsize Normal&\scriptsize Environment&\scriptsize Image&\scriptsize Relight 1& \scriptsize Relight 2 & \scriptsize Point Light 1 & \scriptsize Point Light 2\\\midrule%
\rotatebox[origin=c]{90}{\footnotesize Ours}&\includegraphics[width=42pt]{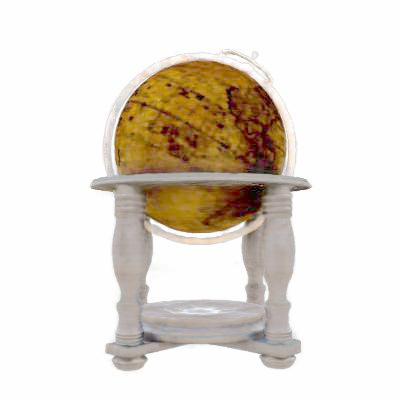}&\includegraphics[width=42pt]{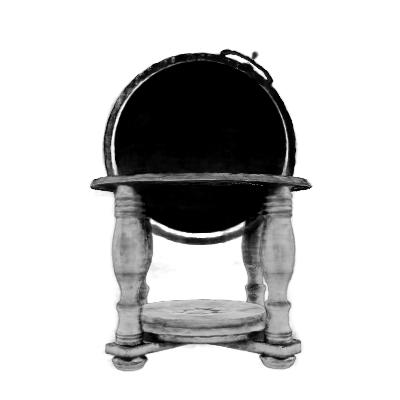}&\includegraphics[width=42pt]{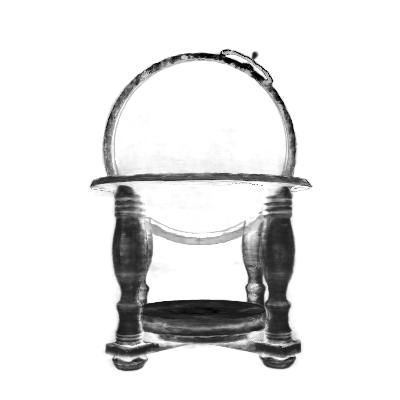}&\includegraphics[width=42pt]{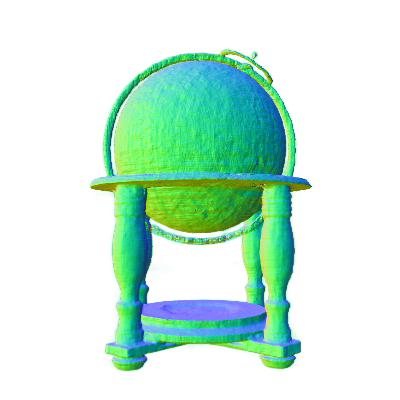}&\includegraphics[width=84pt]{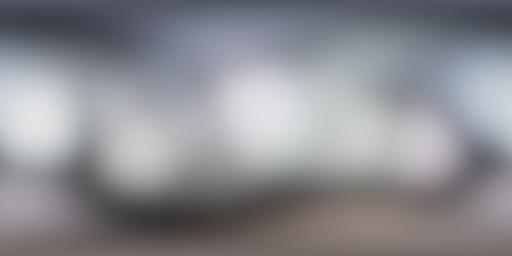}&\includegraphics[width=42pt]{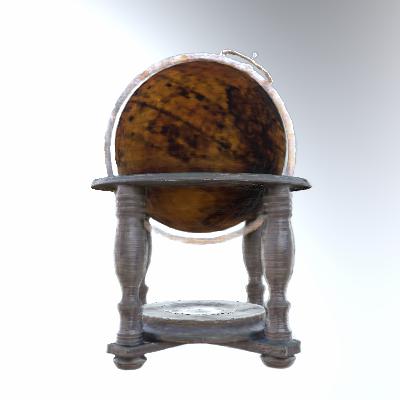}&\includegraphics[width=42pt]{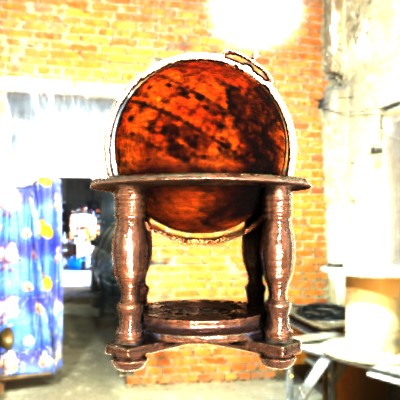}&\includegraphics[width=42pt]{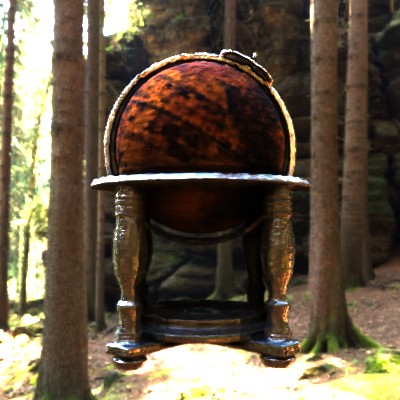}&\includegraphics[width=42pt]{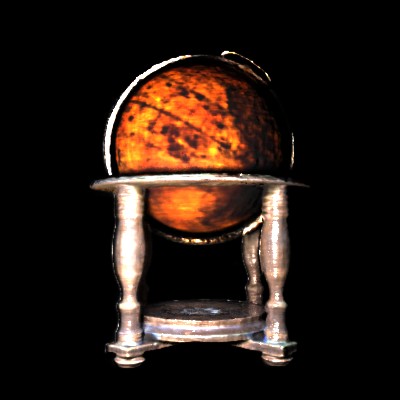}&\includegraphics[width=42pt]{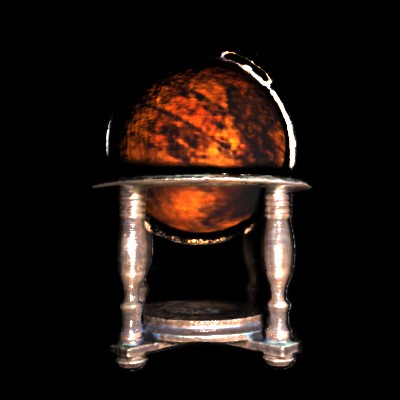}\\%
%

\rotatebox[origin=c]{90}{\footnotesize GT}&\includegraphics[width=42pt]{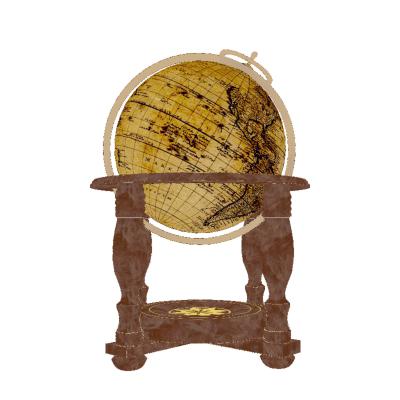}&\includegraphics[width=42pt]{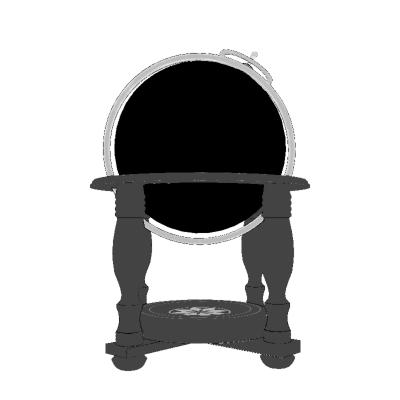}&\includegraphics[width=42pt]{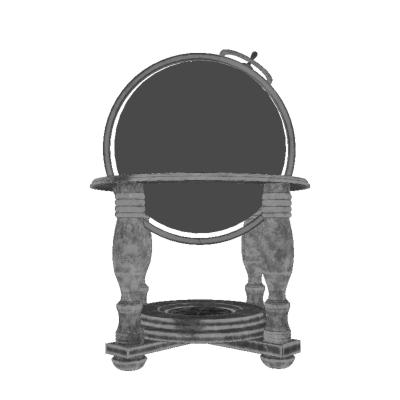}&\includegraphics[width=42pt]{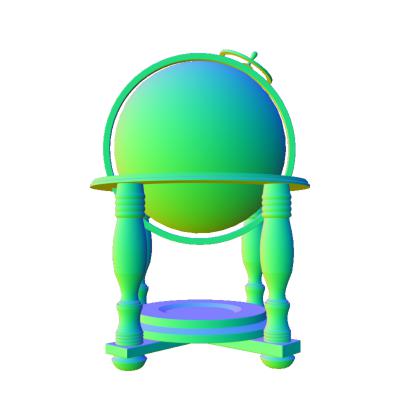}&\includegraphics[width=84pt]{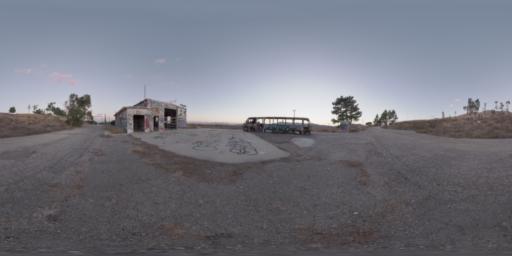}&\includegraphics[width=42pt]{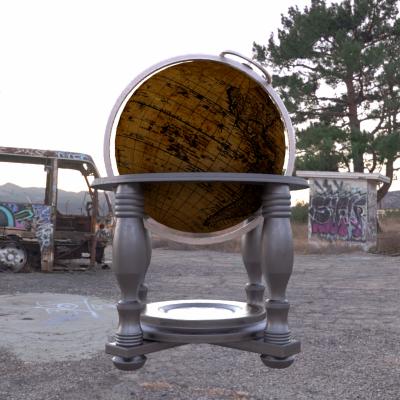}&\includegraphics[width=42pt]{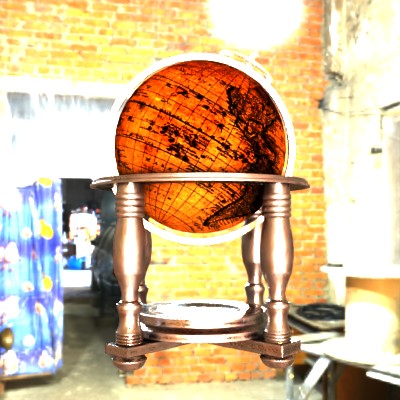}&\includegraphics[width=42pt]{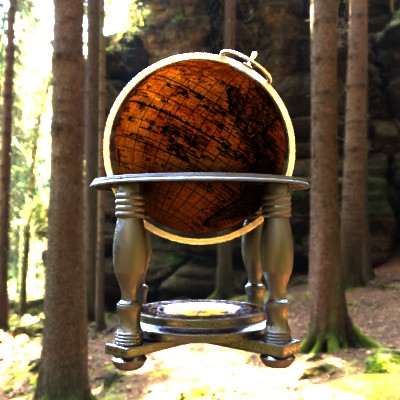}&\includegraphics[width=42pt]{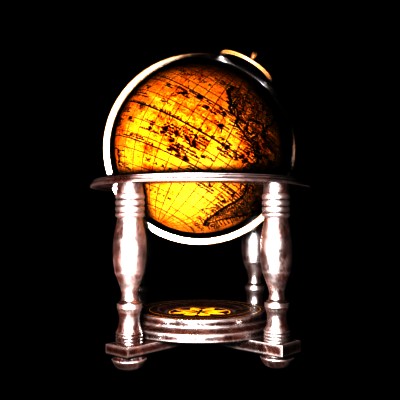}&\includegraphics[width=42pt]{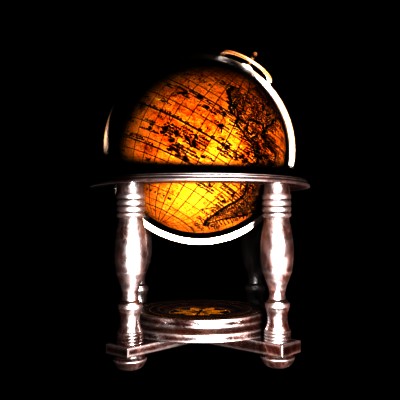}\\
\end{tabular}

    \titlecaption{Globe Decomposition}{Results on the decomposition of the synthetic globe scene. Even if the BRDF parameters do not capture the ground truth perfectly, the visual and quantitative error in re-rendering is extremely low. For most images, an alternative decomposition, which explains the input images, is found. As no other constraints exist, the solution is also plausible.}
    \label{fig:globe}
\end{figure*}

\begin{figure*}
    \centering
    \normalsize%
\setlength{\tabcolsep}{0pt}%
\begin{tabular}{@{}%
l%
p{10pt}%
>{\centering\arraybackslash}m{42pt}%
>{\centering\arraybackslash}m{42pt}%
>{\centering\arraybackslash}m{42pt}%
>{\centering\arraybackslash}m{42pt}%
>{\centering\arraybackslash}m{42pt}%
>{\centering\arraybackslash}m{42pt}%
>{\centering\arraybackslash}m{42pt}%
>{\centering\arraybackslash}m{42pt}%
>{\centering\arraybackslash}m{42pt}%
>{\centering\arraybackslash}m{42pt}%
l@{}}%
&&\scriptsize Img 1&\scriptsize Img 2&\scriptsize Img 3&\scriptsize Img 4&\scriptsize Img 5&\scriptsize Img 6&\scriptsize Img 7&\scriptsize Img 8&\scriptsize Img 9&\scriptsize Img 10&\tiny MSE\textdownarrow\\\midrule%
%
%
\multirow[b]{7}{4mm}{\rotatebox[origin=c]{90}{\scriptsize Diffuse}}&\rotatebox[origin=c]{90}{\tiny GT}&\includegraphics[width=42pt]{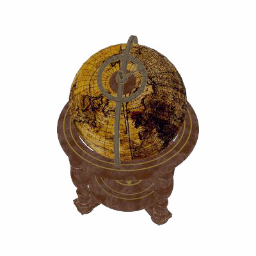}&\includegraphics[width=42pt]{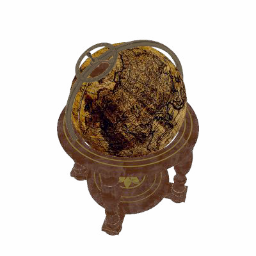}&\includegraphics[width=42pt]{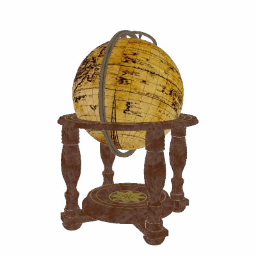}&\includegraphics[width=42pt]{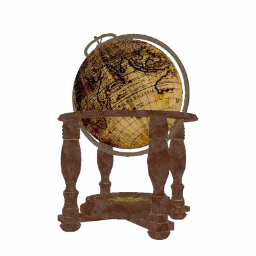}&\includegraphics[width=42pt]{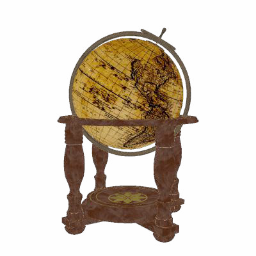}&\includegraphics[width=42pt]{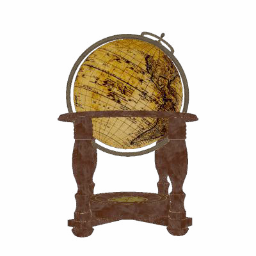}&\includegraphics[width=42pt]{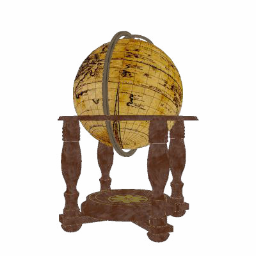}&\includegraphics[width=42pt]{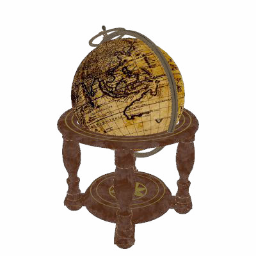}&\includegraphics[width=42pt]{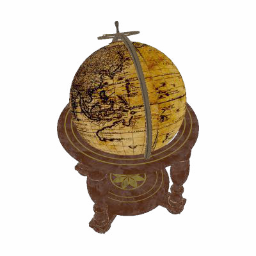}&\includegraphics[width=42pt]{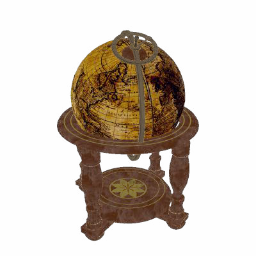}&\\ 
&\rotatebox[origin=c]{90}{\tiny Li \etal}&\includegraphics[width=42pt]{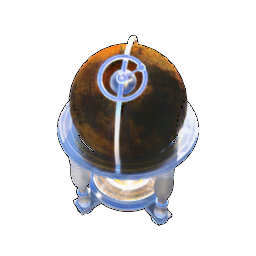}&\includegraphics[width=42pt]{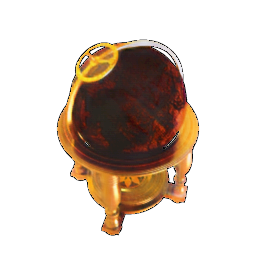}&\includegraphics[width=42pt]{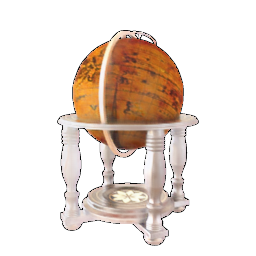}&\includegraphics[width=42pt]{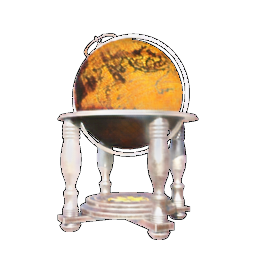}&\includegraphics[width=42pt]{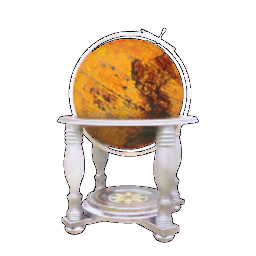}&\includegraphics[width=42pt]{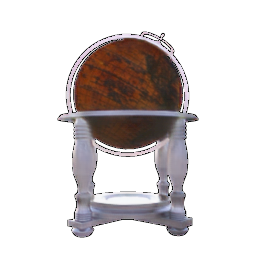}&\includegraphics[width=42pt]{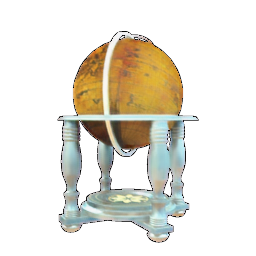}&\includegraphics[width=42pt]{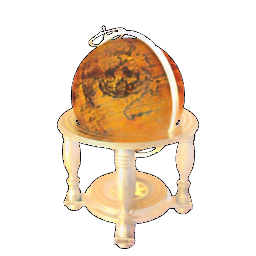}&\includegraphics[width=42pt]{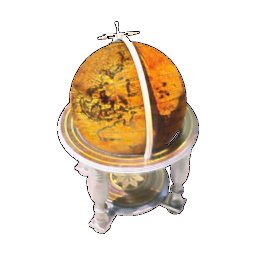}&\includegraphics[width=42pt]{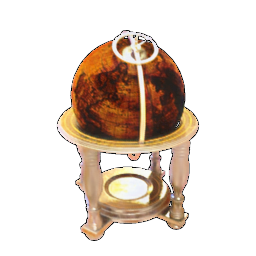}&\rotatebox[origin=c]{90}{\scriptsize 0.0361}\\
&\tiny MSE\textdownarrow&\tiny 0.0309&\tiny 0.0171&\tiny 0.0426&\tiny 0.0475&\tiny 0.0421&\tiny 0.0306&\tiny 0.0365&\tiny 0.0479&\tiny 0.0332&\tiny 0.0324&\\ 
&\rotatebox[origin=c]{90}{\tiny Li \etal + NeRF}&\includegraphics[width=42pt]{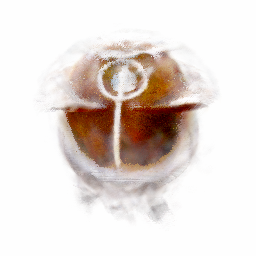}&\includegraphics[width=42pt]{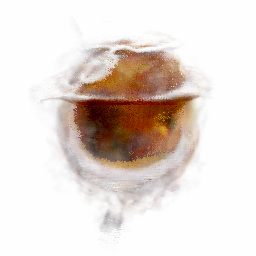}&\includegraphics[width=42pt]{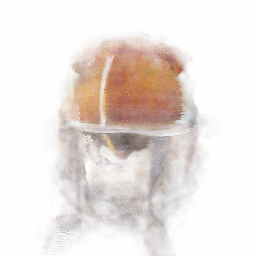}&\includegraphics[width=42pt]{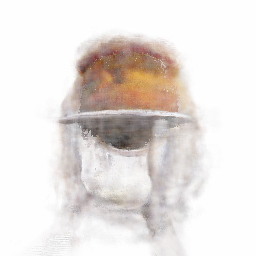}&\includegraphics[width=42pt]{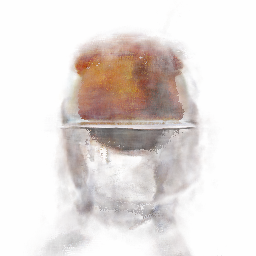}&\includegraphics[width=42pt]{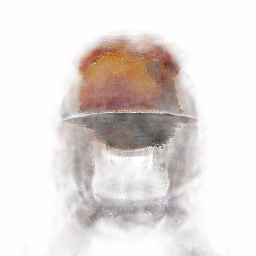}&\includegraphics[width=42pt]{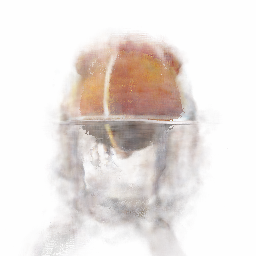}&\includegraphics[width=42pt]{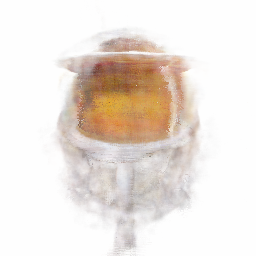}&\includegraphics[width=42pt]{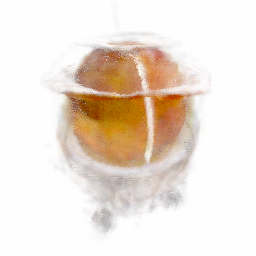}&\includegraphics[width=42pt]{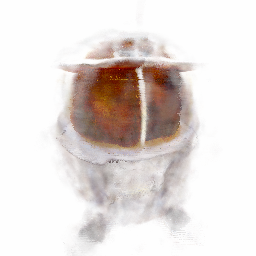}&\rotatebox[origin=c]{90}{\scriptsize 0.0508}\\
&\tiny MSE\textdownarrow&\tiny 0.0471&\tiny 0.0476&\tiny 0.0548&\tiny 0.0549&\tiny 0.0515&\tiny 0.0411&\tiny 0.0504&\tiny 0.0591&\tiny 0.0490&\tiny 0.0525&\\ 
&\rotatebox[origin=c]{90}{\tiny Ours}&\includegraphics[width=42pt]{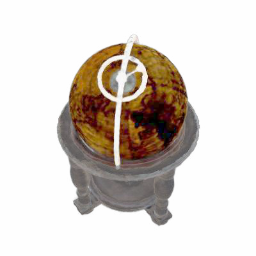}&\includegraphics[width=42pt]{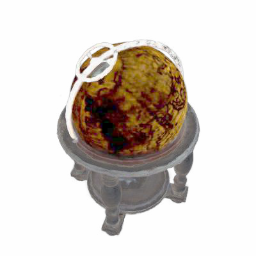}&\includegraphics[width=42pt]{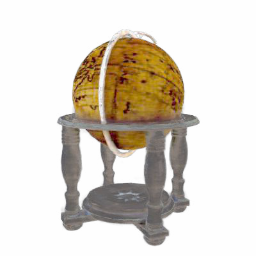}&\includegraphics[width=42pt]{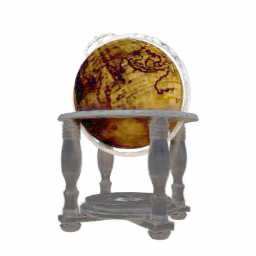}&\includegraphics[width=42pt]{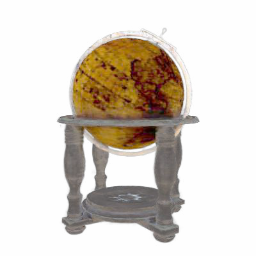}&\includegraphics[width=42pt]{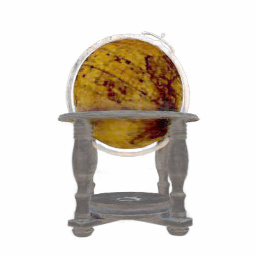}&\includegraphics[width=42pt]{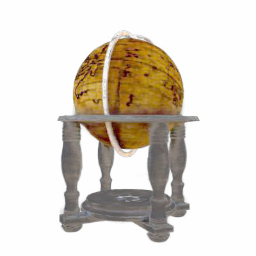}&\includegraphics[width=42pt]{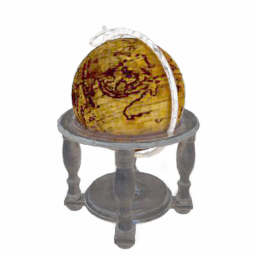}&\includegraphics[width=42pt]{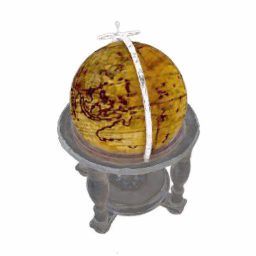}&\includegraphics[width=42pt]{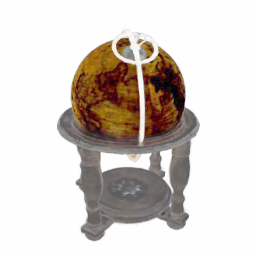}&\rotatebox[origin=c]{90}{\scriptsize \textbf{0.0128}}\\
&\tiny MSE\textdownarrow&\tiny 0.0142&\tiny 0.0151&\tiny 0.0115&\tiny 0.0124&\tiny 0.0133&\tiny 0.0131&\tiny 0.0113&\tiny 0.0125&\tiny 0.0114&\tiny 0.0136&\\\midrule 
%
%
\multirow[b]{7}{4mm}{\rotatebox[origin=c]{90}{\scriptsize Roughness}}&\rotatebox[origin=c]{90}{\tiny GT}&\includegraphics[width=42pt]{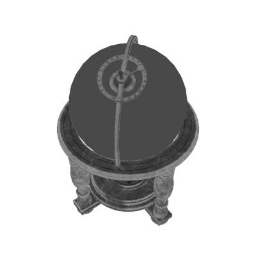}&\includegraphics[width=42pt]{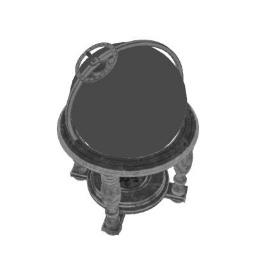}&\includegraphics[width=42pt]{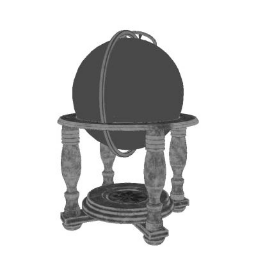}&\includegraphics[width=42pt]{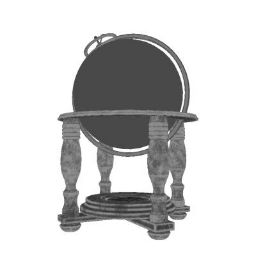}&\includegraphics[width=42pt]{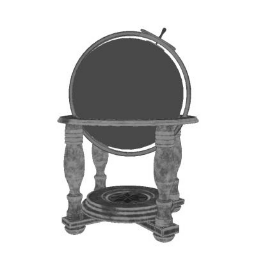}&\includegraphics[width=42pt]{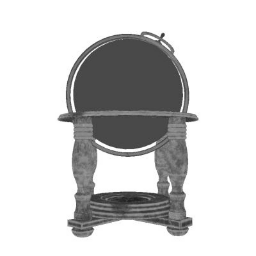}&\includegraphics[width=42pt]{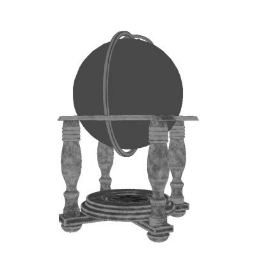}&\includegraphics[width=42pt]{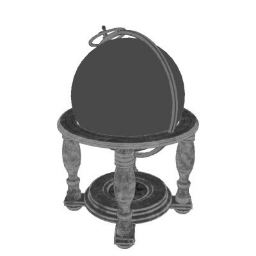}&\includegraphics[width=42pt]{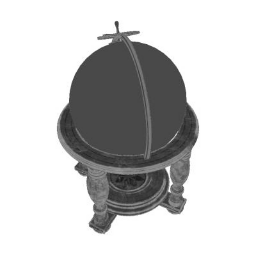}&\includegraphics[width=42pt]{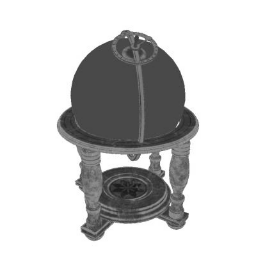}&\\
&\rotatebox[origin=c]{90}{\tiny Li \etal}&\includegraphics[width=42pt]{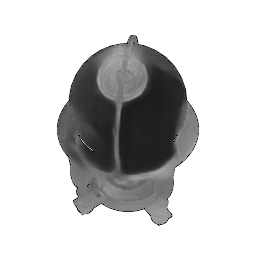}&\includegraphics[width=42pt]{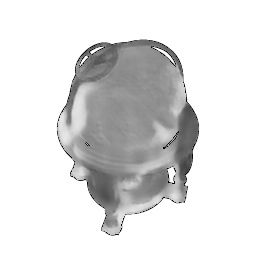}&\includegraphics[width=42pt]{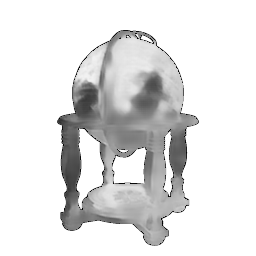}&\includegraphics[width=42pt]{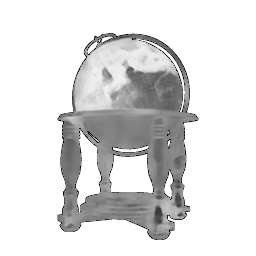}&\includegraphics[width=42pt]{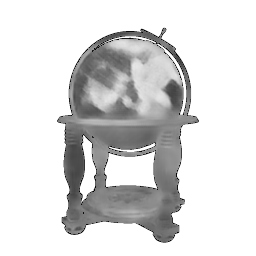}&\includegraphics[width=42pt]{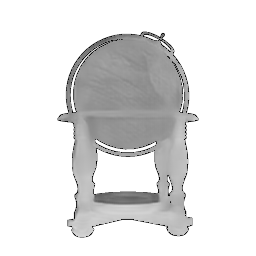}&\includegraphics[width=42pt]{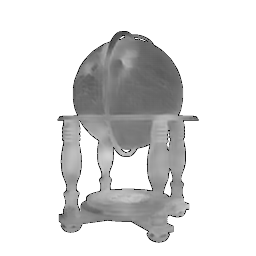}&\includegraphics[width=42pt]{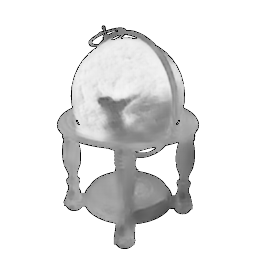}&\includegraphics[width=42pt]{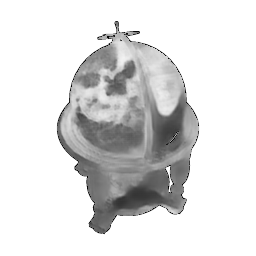}&\includegraphics[width=42pt]{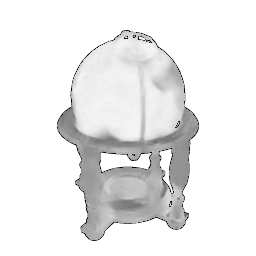}&\rotatebox[origin=c]{90}{\scriptsize \textbf{0.0310}}\\
&\tiny MSE\textdownarrow&\tiny 0.0071&\tiny 0.0280&\tiny 0.0374&\tiny 0.0284&\tiny 0.0229&\tiny 0.0240&\tiny 0.0128&\tiny 0.0484&\tiny 0.0322&\tiny 0.0687&\\
&\rotatebox[origin=c]{90}{\tiny Li \etal + NeRF}&\includegraphics[width=42pt]{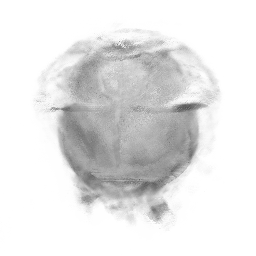}&\includegraphics[width=42pt]{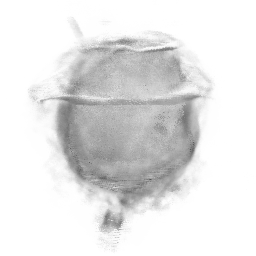}&\includegraphics[width=42pt]{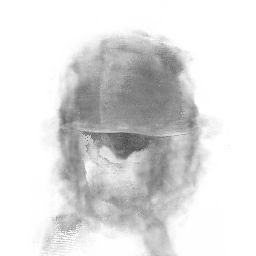}&\includegraphics[width=42pt]{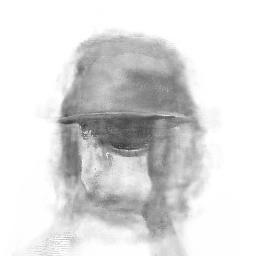}&\includegraphics[width=42pt]{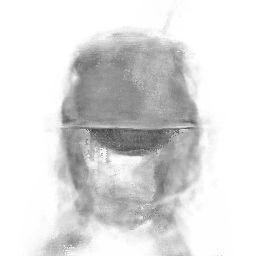}&\includegraphics[width=42pt]{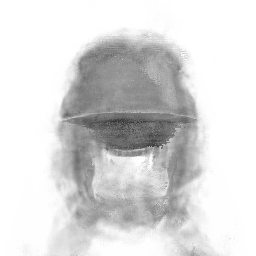}&\includegraphics[width=42pt]{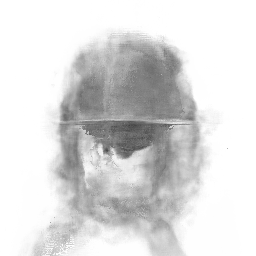}&\includegraphics[width=42pt]{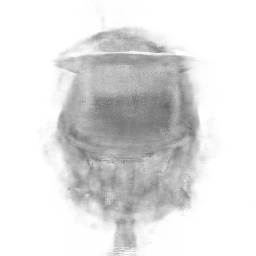}&\includegraphics[width=42pt]{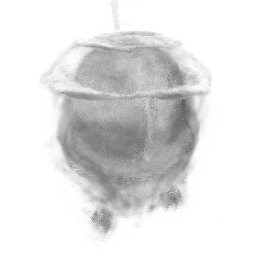}&\includegraphics[width=42pt]{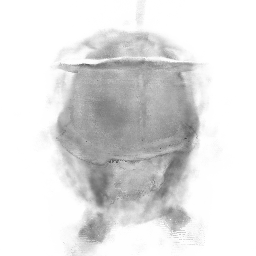}&\rotatebox[origin=c]{90}{\scriptsize 0.0377}\\
&\tiny MSE\textdownarrow&\tiny 0.0447&\tiny 0.0481&\tiny 0.0348&\tiny 0.0315&\tiny 0.0367&\tiny 0.0274&\tiny 0.0304&\tiny 0.0407&\tiny 0.0402&\tiny 0.0428&\\
&\rotatebox[origin=c]{90}{\footnotesize Ours}&\includegraphics[width=42pt]{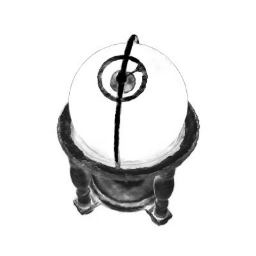}&\includegraphics[width=42pt]{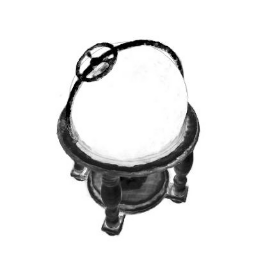}&\includegraphics[width=42pt]{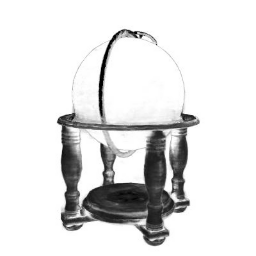}&\includegraphics[width=42pt]{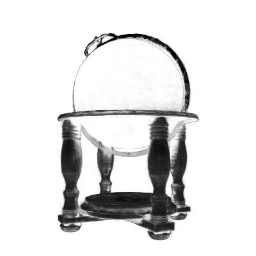}&\includegraphics[width=42pt]{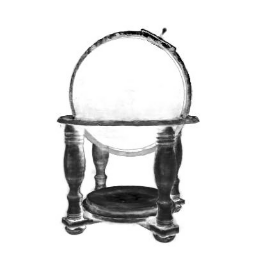}&\includegraphics[width=42pt]{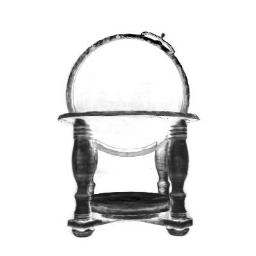}&\includegraphics[width=42pt]{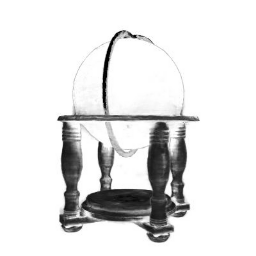}&\includegraphics[width=42pt]{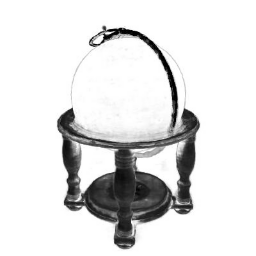}&\includegraphics[width=42pt]{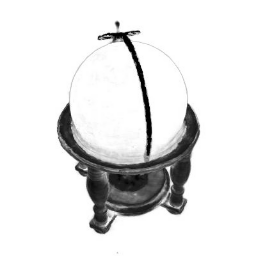}&\includegraphics[width=42pt]{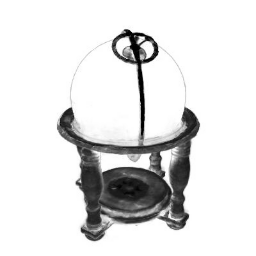}&\rotatebox[origin=c]{90}{\scriptsize 0.0777}\\
&\tiny MSE\textdownarrow&\tiny 0.0826&\tiny 0.0866&\tiny 0.0714&\tiny 0.0756&\tiny 0.0761&\tiny 0.0752&\tiny 0.0701&\tiny 0.0760&\tiny 0.0868&\tiny 0.0764&\\\midrule
%
%
\multirow[b]{3}{4mm}{\rotatebox[origin=c]{90}{\scriptsize Specular}}&\rotatebox[origin=c]{90}{\tiny GT}&\includegraphics[width=42pt]{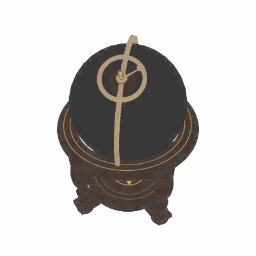}&\includegraphics[width=42pt]{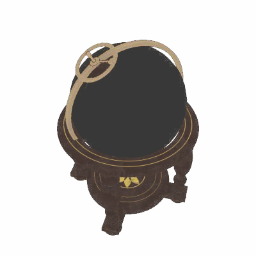}&\includegraphics[width=42pt]{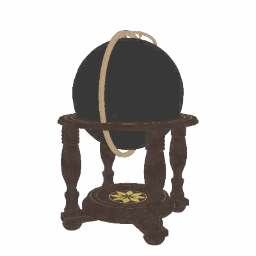}&\includegraphics[width=42pt]{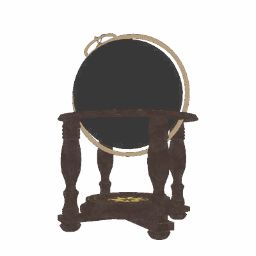}&\includegraphics[width=42pt]{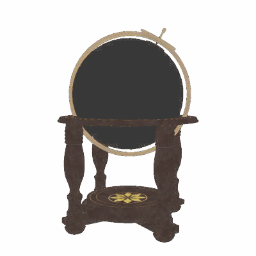}&\includegraphics[width=42pt]{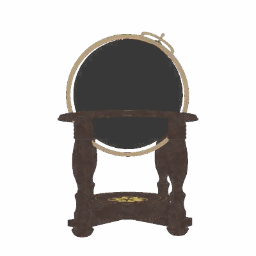}&\includegraphics[width=42pt]{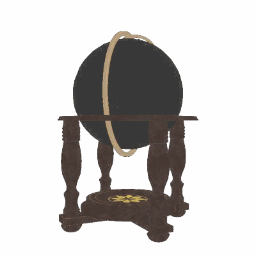}&\includegraphics[width=42pt]{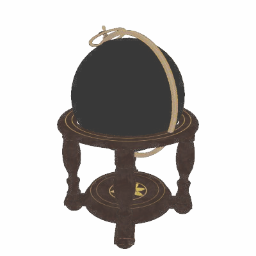}&\includegraphics[width=42pt]{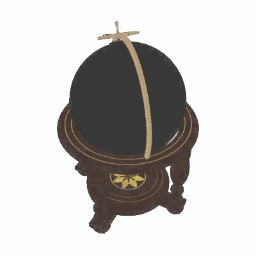}&\includegraphics[width=42pt]{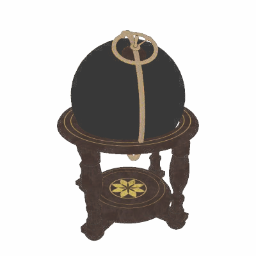}&\\
&\rotatebox[origin=c]{90}{\tiny Ours}&\includegraphics[width=42pt]{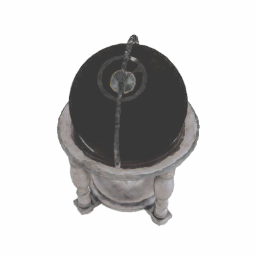}&\includegraphics[width=42pt]{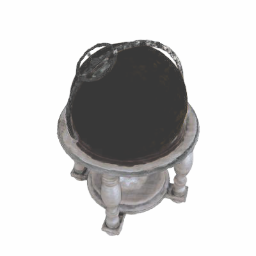}&\includegraphics[width=42pt]{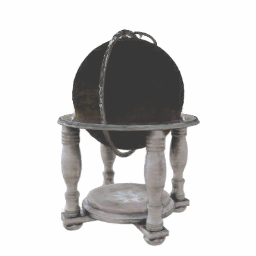}&\includegraphics[width=42pt]{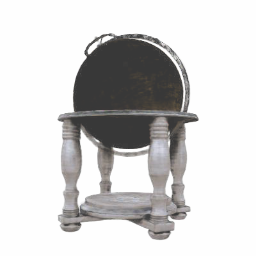}&\includegraphics[width=42pt]{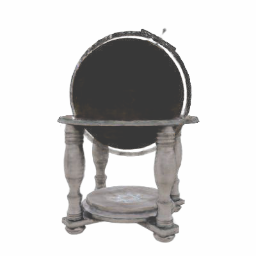}&\includegraphics[width=42pt]{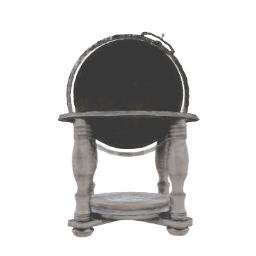}&\includegraphics[width=42pt]{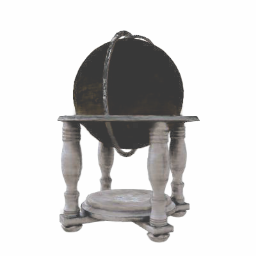}&\includegraphics[width=42pt]{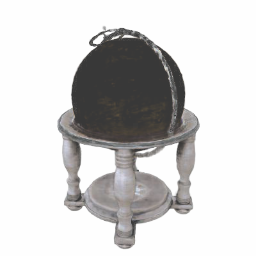}&\includegraphics[width=42pt]{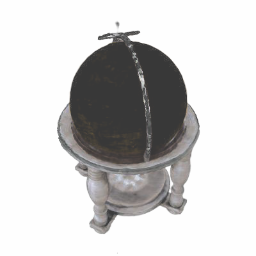}&\includegraphics[width=42pt]{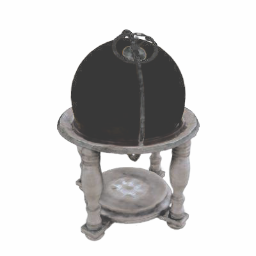}&\rotatebox[origin=c]{90}{\scriptsize 0.0132}\\
&\tiny MSE\textdownarrow&\tiny 0.0109&\tiny 0.0126&\tiny 0.0137&\tiny 0.0165&\tiny 0.0124&\tiny 0.0110&\tiny 0.0136&\tiny 0.0157&\tiny 0.0121&\tiny 0.0134&\\
\end{tabular}
    \titlecaption{Partial Estimation Globe Comparison}{Comparison with partial estimation methods. Here, we compare our method with Li \etal\cite{Li2018a} and Li \etal + NeRF. Li \etal is a method that decomposes a single image of an object illuminated by environment light into diffuse and roughness parameters. For the Li \etal + NeRF baseline, we first translated the training dataset and then trained a NeRF with disabled view conditioning on top. We then generate the novel test set views. For Li \etal, no view synthesis takes place, and the method is run on the test set directly. Notice how our method generates consistent results on all test views.}
    \label{fig:comparison_li}
\end{figure*}

\inlinesection{Additional Results.}
In this section, we show more visual and qualitative results for our training scenes. 
First, we show the performance on our other real-world datasets (Gold Cape and Ethiopian Head). Samples are shown in \fig{fig:rw_novel_view}. The details are preserved and apparent in our reconstructions. The reflective properties also match closely. We also want to highlight the prediction quality compared to NeRF~\cite{mildenhall2020} in \fig{fig:nerf_comparison}. Here, especially in the scenes with varying illumination (Chair and Gnome), NeRF fails as expected. Our method decomposes the information, and after rendering the view, synthesis is close to ground truth. In the scenes with fixed illumination (Head and Cape), the performance between NeRF and our method is on par in most parts. The main difference in MSE is due to the baked-in highlights of NeRF. Our physically grounded design using rendering reduces these artifacts drastically. We also want to point out that relighting a scene is not possible in NeRF.
Lastly, in \fig{fig:globe} an additional example for the accuracy of the BRDF prediction is shown. In the scene, the BRDFs do not reproduce the input perfectly. However, the re-rendering shows a small error and is visually also close. As the optimization is fully unconstrained, the decomposition found a solution that perfectly explains the input images. 

\inlinesection{Results from partial estimation techniques.}
In \fig{fig:geometry_reconstruction} we show a failure case of running COLMAP on data with varying illumination. An accurate surface normal is required for a correct BRDF estimation. If a pipeline is constructed where the first step is an independent geometry reconstruction, the method will fail. Another approach in a partial estimation is to decompose the BRDF for each image independently. Li \etal~\cite{Li2018a} is a method that decomposes objects illuminated by environment illumination only into diffuse and specular roughness. However, no novel views cannot be synthesized, and single image decomposition is highly ill-posed. In \fig{fig:comparison_li} results are shown. The diffuse parameter in our method is more consistent compared to Li \etal, and it is apparent that Li \etal failed to factor out the illumination from the diffuse. However, the roughness is slightly better for Li \etal but is not as consistent, and the roughness is highly correlated with the texture of the globe, which is not correct. Our method is biased towards the extreme roughness value range but is more consistent. It is also worth noting that the roughness parameter plays a minor role compared to the diffuse color during re-rendering. If the color of an object is off, it is more visible than slight alterations in how reflective it is. Additionally, our method can estimate the specular color, which is a challenging task and allows our method to render metals correctly.

As Li \etal does not allow for drastic novel view synthesis -- except slightly based on the estimated depth map -- one approach to solve this is to use NeRF on top. By running Li \etal on the train set and then constraining NeRF to not use view-dependent effects and extending the RGB space to RGB + roughness, we can try to join the distinct images in a volumetric model. In \fig{fig:comparison_li} it is clearly visible that this method fails, as each image is quite different from the other, and NeRF cannot place the varying information at the correct locations.

\inlinesection{Notations.}
All notations in this work are listed in \tbl{tab:notations}.

\begin{table*}
    \centering
    \footnotesize
    \begin{tabular}{lp{3.5cm}p{10.5cm}}
\toprule%
Symbol 			        	& $\in$ 										& Description \\ \midrule%
$q$ 				        & $\mathbb{N}$ 									& Number of images\\%
$s$					        & $\mathbb{N}$ 									& Number of pixel in each image\\%
$I_j$ 				        & $\mathbb{R}^{s \times 3}; j \in {1,...,q}$ 	& A specific image \\%
$\vect{x}$ 			        & $\mathbb{R}^{3}$ 								& The 3D coordinate in (x,y,z) \\%
$\vect{b}$			        & $\mathbb{R}^{5}$								& The BRDF parameters for the analytical cook torrance model \\%
$\vect{n}$			        & $\mathbb{R}^{3}$								& The surface normal \\%
$\sigma$			        & $\mathbb{R}$									& The density in the volume \\%
$\matt{\Gamma}$		        & $\mathbb{R}^{24 \times 7}$					& The parameters for the spherical Gaussians environment illumination \\%
$t$                         & $\mathbb{R}$                                  & Used to query a position in distance $t$ on a ray \\%
$\vect{o}$                  & $\mathbb{R}^{3}$                              & The ray origin \\%
$\vect{d}$                  & $\mathbb{R}^{3}$                              & The ray direction \\%
$\vect{p}$                  & $\mathbb{R}^{z}$                              & A placeholder for a output of an object. Can be either BRDF parameters $\vect{b}$ or color $\vect{c}$. The dimensions $z$ are dependent on the output type. \\%
$\vect{c}^j$                & $\mathbb{R}^{3}$                              & The potentially illumination dependent optimized color for the image $j$  \\%
$\vect{c}^j_{{\omega_o}_r}$ & $\mathbb{R}^{3}$                              & The illumination and view dependent optimized color for the image $j$ \\ %
$\vect{\hat{c}}^j$          & $\mathbb{R}^{3}$                              & The actual color for the image $j$ \\%
$t_n$                       & $\mathbb{R}$                                  & The near clipping distance for the view frustum \\%
$t_f$                       & $\mathbb{R}$                                  & The far clipping distance of the view frustum \\%
$\vect{\omega_i}$           & $\mathbb{R}^{3}$                              & The incoming light direction (Pointing away from the surface) \\%
$\vect{\omega_o}$           & $\mathbb{R}^{3}$                              & The outgoing reflected light direction (Pointing away from the surface) \\ 
$\Omega$    &   & Defines the hemisphere at a point in normal direction $\vect{n}$ \\ \bottomrule%
Function        & $\in$         & Description \\ \midrule %
$L_o(\vect{x}, \vect{\omega_o})$ & $f(\mathbb{R}^{3} \times \mathbb{R}^{3}) \mapsto \mathbb{R}^{3}$ & The amount of outgoing light in the specified direction \\%
$L_i(\vect{x}, \vect{\omega_i})$ &  $f(\mathbb{R}^{3} \times \mathbb{R}^{3}) \mapsto \mathbb{R}^{3}$ & The amount of incoming light from a specified direction \\%
$f_r(\vect{x}, \vect{\omega_i}, \vect{\omega_o})$ & $f(\mathbb{R}^{3} \times \mathbb{R}^{3} \times \mathbb{R}^{3}) \mapsto \mathbb{R}^{3}$ & The BRDF, which is dependent on the position on the surface and the incoming and outgoing light directions \\%
$\rho_d(\vect{\omega_o}, \vect{\Gamma}, \vect{n}, \vect{b})$ & $f((\mathbb{R}^{3} \times \mathbb{R}^{7} \times \mathbb{R}^{3} \times \mathbb{R}^{5}) \mapsto \mathbb{R}^{3}$ & The diffuse lobe evaluation using spherical Gaussian representations \\%
$\rho_s(\vect{\omega_o}, \vect{\Gamma}, \vect{n}, \vect{b})$ & $f((\mathbb{R}^{3} \times \mathbb{R}^{7} \times \mathbb{R}^{3} \times \mathbb{R}^{5}) \mapsto \mathbb{R}^{3}$ & The specular lobe evaluation using spherical Gaussian representations \\%
$\gamma(\vect{x})$            & $f(\mathbb{R}^{3}) \mapsto \mathbb{R}^{3z}$ & Maps the input point coordinate to a higher dimensional embedding using $z$ Fourier embedding  \\%
\bottomrule%
    \end{tabular}
    \titlecaption{Notations}{Overview of all notations used in this work.}
    \label{tab:notations}
\end{table*}

}{}

{\small
\bibliographystyle{ieee_fullname}
\bibliography{nerd}

\begin{thebibliography}{10}\itemsep=-1pt

\bibitem{ISO12232}
Technical Committee~ISO/TC 42.
\newblock {Photography — Digital still cameras — Determination of exposure
  index, ISO speed ratings, standard output sensitivity, and recommended
  exposure index}.
\newblock Standard, International Organization for Standardization, 2019.

\bibitem{Aittala2018}
Miika Aittala, Timo Aila, and Jaakko Lehtinen.
\newblock Reflectance modeling by neural texture synthesis.
\newblock In {\em ACM Transactions on Graphics (ToG)}, 2018.

\bibitem{Albert2018}
Rachel Albert, Dorian~Yao Chan, Dan~B. Goldman, and James~F. O'Brian.
\newblock Approximate {svBRDF} estimation from mobile phone video.
\newblock In {\em Eurographics Symposium on Rendering}, 2018.

\bibitem{Asselin2020}
Louis-Philippe Asselin, Denis Laurendeau, and Jean-François Lalonde.
\newblock Deep svbrdf estimation on real materials.
\newblock In {\em International Conference on 3D Vision (3DV)}, 2020.

\bibitem{Barron2015}
Jonathan~T. Barron and Jitendra Malik.
\newblock Shape, illumination, and reflectance from shading.
\newblock In {\em IEEE Transactions on Pattern Analysis and Machine
  Intelligence (PAMI)}, 2015.

\bibitem{bi2020b}
Sai Bi, Zexiang Xu, Pratul Srinivasan, Ben Mildenhall, Kalyan Sunkavalli,
  Miloš Hašan, Yannick Hold-Geoffroy, David Kriegman, and Ravi Ramamoorthi.
\newblock Neural reflectance fields for appearance acquisition.
\newblock {\em ArXiv e-prints}, 2020.

\bibitem{bi2020a}
Sai Bi, Zexiang Xu, Kalyan Sunkavalli, Miloš Hašan, Yannick Hold-Geoffroy,
  David Kriegman, and Ravi Ramamoorthi.
\newblock Deep reflectance volumes: Relightable reconstructions from multi-view
  photometric images.
\newblock In {\em European Conference on Computer Vision (ECCV)}, 2020.

\bibitem{bi2020c}
Sai Bi, Zexiang Xu, Kalyan Sunkavalli, David Kriegman, and Ravi Ramamoorthi.
\newblock Deep 3d capture: Geometry and reflectance from sparse multi-view
  images.
\newblock In {\em IEEE Conference on Computer Vision and Pattern Recognition
  (CVPR)}, 2020.

\bibitem{Boss2018}
Mark Boss, Fabian Groh, Sebastian Herholz, and Hendrik P.~A. Lensch.
\newblock {Deep Dual Loss BRDF Parameter Estimation}.
\newblock In {\em Workshop on Material Appearance Modeling}, 2018.

\bibitem{Boss2020}
Mark Boss, Varun Jampani, Kihwan Kim, Hendrik~P.A. Lensch, and Jan Kautz.
\newblock Two-shot spatially-varying brdf and shape estimation.
\newblock In {\em IEEE Conference on Computer Vision and Pattern Recognition
  (CVPR)}, 2020.

\bibitem{Burley2012}
Brent Burley.
\newblock Physically based shading at disney.
\newblock In {\em ACM Transactions on Graphics (SIGGRAPH)}, 2012.

\bibitem{carWreck}
{c}gtrader.
\newblock {C}arwreck.
\newblock
  \url{https://www.cgtrader.com/free-3d-models/vehicle/other/car-wreck-pbr-game-asset}.

\bibitem{Chair}
{c}gtrader.
\newblock {C}hair.
\newblock
  \url{https://www.cgtrader.com/free-3d-models/furniture/chair/freifrau-easy-chair-pbr}.

\bibitem{blender}
Blender~Online Community.
\newblock {\em Blender - a 3D modelling and rendering package}.
\newblock Blender Foundation, Stichting Blender Foundation, Amsterdam, 2018.

\bibitem{Cook1982}
Robert~L. Cook and Kenneth~E. Torrance.
\newblock A reflectance model for computer graphics.
\newblock {\em ACM Transactions on Graphics (ToG)}, 1982.

\bibitem{Deschaintre2018}
Valentin Deschaintre, Miika Aitalla, Fredo Durand, George Drettakis, and Adrien
  Bousseau.
\newblock Single-image {SVBRDF} capture with a rendering-aware deep network.
\newblock In {\em ACM Transactions on Graphics (ToG)}, 2018.

\bibitem{Deschaintre2019}
Valentin Deschaintre, Miika Aitalla, Fredo Durand, George Drettakis, and Adrien
  Bousseau.
\newblock Flexible {SVBRDF} capture with a multi-image deep network.
\newblock In {\em Eurographics Symposium on Rendering}, 2019.

\bibitem{Deschaintre2020}
Valentin Deschaintre, George Drettakis, and Adrien Bousseau.
\newblock Guided fine-tuning for large-scale material transfer.
\newblock In {\em Eurographics Symposium on Rendering}, 2020.

\bibitem{dong2014}
Yue Dong, Guojun Chen, Pieter Peers, Jianwen Zhang, and Xin Tong.
\newblock Appearance-from-motion: Recovering spatially varying surface
  reflectance under unknown lighting.
\newblock {\em ACM Transactions on Graphics (SIGGRAPH ASIA)}, 2014.

\bibitem{Gao2019}
Duan Gao, Xiao Li, Yue Dong, Pieter Peers, and Xin Tong.
\newblock Deep inverse rendering for high-resolution svbrdf estimation from an
  arbitrary number of images.
\newblock In {\em ACM Transactions on Graphics (SIGGRAPH)}, 2019.

\bibitem{Gardner2017}
Marc-Andr\'{e} Gardner, Kalyan Sunkavalli, Ersin Yumer, Xiaohui Shen, Emiliano
  Gambaretto, Christian Gagn\'{e}, and Jean-Fran\c{c}ois Lalonde.
\newblock Learning to predict indoor illumination from a single image.
\newblock {\em ACM Transactions on Graphics (ToG)}, 2017.

\bibitem{Goldman2009}
Dan~B. Goldman, Brian Curless, Aaron Hertzmann, and Steven~M. Seitz.
\newblock Shape and spatially-varying {BRDF}s from photometric stereo.
\newblock {\em IEEE Transactions on Pattern Analysis and Machine Intelligence
  (PAMI)}, 2009.

\bibitem{haber2009}
Tom Haber, Christian Fuchs, Phillipe Bekaer, Hans-Peter Seidel, Michael
  Goesele, and Hendrik P.~A. Lensch.
\newblock Relighting objects from image collections.
\newblock In {\em IEEE Conference on Computer Vision and Pattern Recognition
  (CVPR)}, 2009.

\bibitem{Kajiya1986}
James~T. Kajiya.
\newblock The rendering equation.
\newblock In {\em ACM Transactions on Graphics (SIGGRAPH)}, 1986.

\bibitem{kazhdan2006poisson}
Michael Kazhdan, Matthew Bolitho, and Hugues Hoppe.
\newblock Poisson surface reconstruction.
\newblock In {\em Proceedings of the fourth Eurographics symposium on Geometry
  processing}, volume~7, 2006.

\bibitem{Kimiccv17}
Kihwan Kim, Jinwei Gu, Stephen Tyree, Pavlo Molchanov, Matthias Nie{\ss}ner,
  and Jan Kautz.
\newblock A lightweight approach for on-the-fly reflectance estimation.
\newblock In {\em IEEE International Conference on Computer Vision (ICCV)},
  2017.

\bibitem{kingma2014adam}
Diederik~P Kingma and Jimmy Ba.
\newblock Adam: A method for stochastic optimization.
\newblock {\em ArXiv e-prints}, 2014.

\bibitem{lawrence2004}
Jason Lawrence, Szymon Rusinkiewicz, and Ravi Ramamoorthi.
\newblock Efficient brdf importance sampling using a factored representation.
\newblock {\em ACM Transactions on Graphics (ToG)}, 2004.

\bibitem{Lensch2001}
Hendrik Lensch, Jan Kautz, Michael Gosele, and Hans-Peter Seidel.
\newblock Image-based reconstruction of spatially varying materials.
\newblock In {\em Eurographics Conference on Rendering}, 2001.

\bibitem{Lensch2003}
Hendrik~P.A. Lensch, Jochen Lang, M.~Sa Asla, and Hans‐Peter Seidel.
\newblock Planned sampling of spatially varying brdfs.
\newblock In {\em Computer Graphics Forum}, 2003.

\bibitem{Li2017}
Xiao Li, Yue Dong, Pieter Peers, and Xin Tong.
\newblock Modeling surface appearance from a single photograph using
  self-augmented convolutional neural networks.
\newblock In {\em ACM Transactions on Graphics (ToG)}, 2017.

\bibitem{li2020inverse}
Zhengqin Li, Mohammad Shafiei, Ravi Ramamoorthi, Kalyan Sunkavalli, and
  Manmohan Chandraker.
\newblock Inverse rendering for complex indoor scenes: Shape, spatially-varying
  lighting and svbrdf from a single image.
\newblock In {\em IEEE Conference on Computer Vision and Pattern Recognition
  (CVPR)}, 2020.

\bibitem{Li2018}
Zhengqin Li, Kalyan Sunkavalli, and Manmohan Chandraker.
\newblock Materials for masses: {SVBRDF} acquisition with a single mobile phone
  image.
\newblock In {\em European Conference on Computer Vision (ECCV)}, 2018.

\bibitem{Li2018a}
Zhengqin Li, Zexiang Xu, Ravi Ramamoorthi, Kalyan Sunkavalli, and Manmohan
  Chandraker.
\newblock Learning to reconstruct shape and spatially-varying reflectance from
  a single image.
\newblock In {\em ACM Transactions on Graphics (SIGGRAPH ASIA)}, 2018.

\bibitem{liu2020}
Lingjie Liu, iatao Gu, Kyaw Zaw~Lin, Tat-Seng Chua, and Christian Theobalt.
\newblock Neural sparse voxel fields.
\newblock In {\em Advances in Neural Information Processing Systems (NeurIPS)},
  2020.

\bibitem{martinbrualla2020nerfw}
Ricardo Martin-Brualla, Noha Radwan, Mehdi S.~M. Sajjadi, Jonathan~T. Barron,
  Alexey Dosovitskiy, and Daniel Duckworth.
\newblock {NeRF in the Wild: Neural Radiance Fields for Unconstrained Photo
  Collections}.
\newblock In {\em ArXiv e-prints}, 2020.

\bibitem{mescheder2019occupancy}
Lars Mescheder, Michael Oechsle, Michael Niemeyer, Sebastian Nowozin, and
  Andreas Geiger.
\newblock Occupancy networks: Learning 3d reconstruction in function space.
\newblock In {\em IEEE Conference on Computer Vision and Pattern Recognition
  (CVPR)}, 2019.

\bibitem{mildenhall2020}
Ben Mildenhall, Pratul Srinivasan, Matthew Tancik, Jonathan~T. Barron, Ravi
  Ramamoorthi, and Ren Ng.
\newblock Nerf: Representing scenes as neural radiance fields for view
  synthesis.
\newblock In {\em European Conference on Computer Vision (ECCV)}, 2020.

\bibitem{Nam2018}
Giljoo Nam, Diego Gutierrez, and Min~H. Kim.
\newblock Practical {SVBRDF} acquisition of 3d objects with unstructured flash
  photography.
\newblock In {\em ACM Transactions on Graphics (SIGGRAPH ASIA)}, 2018.

\bibitem{Nicodemus1965}
Fred~E. Nicodemus.
\newblock Directional reflectance and emissivity of an opaque surface.
\newblock {\em Applied Optics}, 1965.

\bibitem{daniel_pett_ethiopian_head}
Daniel Pett.
\newblock {Ethiopian Head}, 2016.

\bibitem{daniel_pett_gold_cape}
Daniel Pett.
\newblock {BritishMuseumDH/moldGoldCape: First release of the Cape in 3D}, Mar.
  2017.

\bibitem{Ramamoorthi2001}
Ravi Ramamoorthi and Pat Hanrahan.
\newblock A signal-processing framework for inverse rendering.
\newblock In {\em ACM Transactions on Graphics (SIGGRAPH)}, 2001.

\bibitem{sang2020single}
Shen Sang and Manmohan Chandraker.
\newblock Single-shot neural relighting and svbrdf estimation.
\newblock In {\em European Conference on Computer Vision (ECCV)}, 2020.

\bibitem{schoenberger2016sfm}
Johannes~Lutz Sch\"{o}nberger and Jan-Michael Frahm.
\newblock Structure-from-motion revisited.
\newblock In {\em Conference on Computer Vision and Pattern Recognition
  (CVPR)}, 2016.

\bibitem{schoenberger2016mvs}
Johannes~Lutz Sch\"{o}nberger, Enliang Zheng, Marc Pollefeys, and Jan-Michael
  Frahm.
\newblock Pixelwise view selection for unstructured multi-view stereo.
\newblock In {\em European Conference on Computer Vision (ECCV)}, 2016.

\bibitem{Sengupta2019}
Soumyadip Sengupta, Jinwei Gu, Kihwan Kim, Guilin Liu, David~W. Jacobs, and Jan
  Kautz.
\newblock Neural inverse rendering of an indoor scene from a single image.
\newblock In {\em IEEE International Conference on Computer Vision (ICCV)},
  2019.

\bibitem{sitzmann2020siren}
Vincent Sitzmann, Julien~N.P. Martel, Alexander~W. Bergman, David~B. Lindell,
  and Gordon Wetzstein.
\newblock Implicit neural representations with periodic activation functions.
\newblock In {\em Advances in Neural Information Processing Systems (NeurIPS)},
  2020.

\bibitem{Sitzmann2019DeepVoxelsLP}
Vincent Sitzmann, Justus Thies, Felix Heide, M. Nie{\ss}ner, G. Wetzstein, and
  M. Zollh{\"o}fer.
\newblock Deepvoxels: Learning persistent 3d feature embeddings.
\newblock {\em IEEE Conference on Computer Vision and Pattern Recognition
  (CVPR)}, 2019.

\bibitem{srinivasan2020}
Pratul~P. Srinivasan, Boyang Deng, Xiuming Zhang, Matthew Tancik, Ben
  Mildenhall, and Jonathan~T. Barron.
\newblock Nerv: Neural reflectance and visibility fields for relighting and
  view synthesis.
\newblock In {\em ArXiv e-prints}, 2020.

\bibitem{tancik2020fourfeat}
Matthew Tancik, Pratul~P. Srinivasan, Ben Mildenhall, Sara Fridovich-Keil,
  Nithin Raghavan, Utkarsh Singhal, Ravi Ramamoorthi, Jonathan~T. Barron, and
  Ren Ng.
\newblock Fourier features let networks learn high frequency functions in low
  dimensional domains.
\newblock {\em Advances in Neural Information Processing Systems (NeurIPS)},
  2020.

\bibitem{Globe}
{T}urbo{S}quid.
\newblock {S}tanding {G}lobe.
\newblock \url{https://www.turbosquid.com/3d-models/3d-standing-globe-1421971}.

\bibitem{Wang2009}
Jiaping Wang, Peiran Ren, Minmin Gong, John Snyder, and Baining Guo.
\newblock All-frequency rendering of dynamic, spatially-varying reflectance.
\newblock In {\em ACM Transactions on Graphics (SIGGRAPH ASIA)}, 2009.

\bibitem{Xia2016}
Rui Xia, Yue Dong, Pieter Peers, and Xin Tong.
\newblock Recovering shape and spatially-varying surface reflectance under
  unknown illumination.
\newblock In {\em ACM Transactions on Graphics (SIGGRAPH ASIA)}, 2016.

\bibitem{qiangeng2019disn}
Qiangeng Xu, Weiyue Wang, Duygu Ceylan, Radomir Mech, and Ulrich Neumann.
\newblock Disn: Deep implicit surface network for high-quality single-view 3d
  reconstruction.
\newblock In {\em Advances in Neural Information Processing Systems (NeurIPS)},
  2019.

\bibitem{xu2019}
Zexiang Xu, Sai Bi, Kalyan Sunkavalli, Sunil Hadap, Hao Su, and Ravi
  Ramamoorthi.
\newblock Deep view synthesis from sparse photometric images.
\newblock {\em ACM Transactions on Graphics (ToG)}, 2019.

\bibitem{xu2018}
Zexiang Xu et~al.
\newblock Deep image-based relighting from optimal sparse samples.
\newblock {\em ACM Transactions on Graphics (ToG)}, 2018.

\bibitem{yariv2020multiview}
Lior Yariv, Yoni Kasten, Dror Moran, Meirav Galun, Matan Atzmon, Basri Ronen,
  and Yaron Lipman.
\newblock Multiview neural surface reconstruction by disentangling geometry and
  appearance.
\newblock {\em Advances in Neural Information Processing Systems (NeurIPS)},
  2020.

\bibitem{Ye2018}
Wenjie Ye, Xiao Li, Yue Dong, Pieter Peers, and Xin Tong.
\newblock Single image surface appearance modeling with self-augmented cnns and
  inexact supervision.
\newblock {\em Computer Graphics Forum}, 2018.

\bibitem{Zhang2020InverseRendering}
Jianzhao Zhang, Guojun Chen, Yue Dong, Jian Shi, Bob Zhang, and Enhua Wu.
\newblock Deep inverse rendering for practical object appearance scan with
  uncalibrated illumination.
\newblock In {\em Advances in Computer Graphics}, 2020.

\bibitem{physg2020}
Kai Zhang, Fujun Luan, Qianqian Wang, Kavita Bala, and Noah Snavely.
\newblock Physg: Inverse rendering with spherical gaussians for physics-based
  material editing and relighting.
\newblock In {\em IEEE Conference on Computer Vision and Pattern Recognition
  (CVPR)}, 2021.

\bibitem{zhang2020}
Kai Zhang, Gernot Riegler, Noah Snavely, and Vladlen Koltun.
\newblock Nerf++: Analyzing and improving neural radiance fields.
\newblock {\em ArXiv e-prints}, 2020.

\bibitem{Zhou2018}
Qian-Yi Zhou, Jaesik Park, and Vladlen Koltun.
\newblock {Open3D}: {A} modern library for {3D} data processing.
\newblock {\em ArXiv e-prints}, 2018.

\end{thebibliography}
}

\end{document}